\documentclass{article}
\usepackage[english]{babel}
\usepackage[utf8]{inputenc}

\usepackage{natbib}

\usepackage{color,xcolor,ucs}

\usepackage{subfig}
\usepackage{tabularx}
\usepackage{float}
\usepackage{amsfonts}
\usepackage{helvet}         % selects Helvetica as sans-serif font
\usepackage{courier}        % selects Courier as typewriter font
\usepackage{type1cm}        % activate if the above 3 fonts are
\usepackage{amsmath}
\usepackage{amssymb}
\usepackage{makeidx}         % allows index generation
\usepackage{comment}         % allows index generation
\usepackage{graphicx}        % standard LaTeX graphics tool
\usepackage[space]{grffile}
\usepackage{multicol}        % used for the two-column index
\usepackage[bottom]{footmisc}% places footnotes at page bottom
\usepackage{bm}
\usepackage{epstopdf}
\usepackage{multirow}
\usepackage{txfonts}
\usepackage{blkarray}

\usepackage{url}
\usepackage{mathtools}
\DeclarePairedDelimiterX{\norm}[1]{\lVert}{\rVert}{#1}

\usepackage[unicode=true, bookmarks=true,colorlinks=true]{hyperref}
\usepackage{xr-hyper}

\usepackage{geometry}
\usepackage{titlesec}

\titleclass{\subsubsubsection}{straight}[\subsection]

\newcounter{subsubsubsection}[subsubsection]
\renewcommand\thesubsubsubsection{\thesubsubsection.\arabic{subsubsubsection}}
\titleformat{\subsubsubsection}
  {\normalfont\normalsize\bfseries}{\thesubsubsubsection}{1em}{}
\titlespacing*{\subsubsubsection}
{0pt}{3.25ex plus 1ex minus .2ex}{1.5ex plus .2ex}

\makeatletter
\renewcommand\paragraph{\@startsection{paragraph}{5}{\z@}%
  {3.25ex \@plus1ex \@minus.2ex}%
  {-1em}%
  {\normalfont\normalsize\bfseries}}
\renewcommand\subparagraph{\@startsection{subparagraph}{6}{\parindent}%
  {3.25ex \@plus1ex \@minus .2ex}%
  {-1em}%
  {\normalfont\normalsize\bfseries}}
\def\toclevel@subsubsubsection{4}
\def\toclevel@paragraph{5}
\def\toclevel@paragraph{6}
\def\l@subsubsubsection{\@dottedtocline{4}{7em}{4em}}
\def\l@paragraph{\@dottedtocline{5}{10em}{5em}}
\def\l@subparagraph{\@dottedtocline{6}{14em}{6em}}
\makeatother

\usepackage{xspace}

\usepackage[font=small,labelfont=bf]{caption}

\DeclareMathOperator*{\argmin}{arg\,min}
\DeclareMathOperator*{\argmax}{arg\,max}

\newcommand{\probb}[1]{\ensuremath{\mathbb{P}({#1})}}
\newcommand{\half}[0]{\frac{1}{2}}
\newcommand{\comb}[4]{{}^{#2} \! #1_{#4}^{#3}}

\newcommand{\belief}[0]{\emph{belief}\xspace}

\newcommand{\unfocused}[0]{\texttt{\emph{unfocused}}\xspace}
\newcommand{\focused}[0]{\texttt{\emph{focused}}\xspace}

\newcommand{\Focused}[0]{\texttt{\emph{Focused}}\xspace}

\newcommand{\involved}[0]{\emph{involved}\xspace}
\newcommand{\notinvolved}[0]{\emph{not involved}\xspace}
\newcommand{\increment}[0]{\emph{increment}\xspace}
\newcommand{\increments}[0]{\emph{increments}\xspace}

\newcommand{\actnaug}[0]{\emph{not-augmented}\xspace}
\newcommand{\actsqr}[0]{\emph{squared}\xspace}
\newcommand{\actrect}[0]{\emph{rectangular}\xspace}

\newcommand{\Actnaug}[0]{\emph{Not-augmented}\xspace}
\newcommand{\Actsqr}[0]{\emph{Squared}\xspace}
\newcommand{\Actrect}[0]{\emph{Rectangular}\xspace}

\newcommand{\ramdl}[0]{\emph{rAMDL}\xspace}

\newcommand{\ramdltree}[0]{\emph{rAMDL-Tree}\xspace}

\newcommand{\recursive}[0]{\emph{Recursive}\xspace}
\newcommand{\backsubstitution}[0]{\emph{Backsubstitution}\xspace}

\newcommand{\twostageapr}[0]{\emph{2-stage}\xspace}
\newcommand{\rectapr}[0]{\emph{Rectangular}\xspace}

\newtheorem{lemma}{Lemma}

\makeatletter
\def\BState{\State\hskip-\ALG@thistlm}
\makeatother

\makeindex             % used for the subject index
\usepackage[linesnumbered, inoutnumbered]{algorithm2e}

\usepackage{rotating}

\providecommand{\keywords}[1]{\textbf{\textit{Keywords}} \\#1}

\title{
General Purpose Incremental Covariance Update
and\\
Efficient Belief Space Planning via\\
Factor-Graph
Propagation Action Tree
}

\author{\renewcommand\footnotemark{}Dmitry Kopitkov and Vadim Indelman
	\thanks{D. Kopitkov is with the Technion Autonomous Systems Program (TASP), Technion - Israel Institute of Technology, Haifa 32000, Israel, {\tt dimkak@technion.ac.il
}. V. Indelman is with the Department of Aerospace Engineering, Technion - Israel Institute of Technology, Haifa 32000, Israel, {\tt vadim.indelman@technion.ac.il}. This work was  supported by the Israel Science Foundation. %This work was partially supported by the Technion Autonomous Systems Program (TASP) and by the Ministry of Science \& Technology, Israel \& the Russian Foundation for Basic Research, the Russian Federation?.
		}	
}

\date{}

\begin{document}

\maketitle

% ======================
\begin{abstract}

Fast covariance calculation is required both for SLAM (e.g.~in order to solve data association) and for evaluating the information-theoretic term for different candidate actions in belief space planning (BSP). In this paper we make two primary contributions. First, we develop a novel general-purpose incremental covariance update technique, which efficiently recovers specific covariance entries after any change in the inference problem, such as introduction of new observations/variables or re-linearization of the state vector. Our approach is shown to recover them faster than other state-of-the-art methods.
Second, we present a computationally efficient approach for BSP in high-dimensional state spaces, leveraging our incremental covariance update method. State of the art BSP approaches perform belief propagation for each candidate action and then evaluate an objective function that typically includes an information-theoretic term, such as entropy or information gain. Yet, candidate actions often have similar parts (e.g.~common trajectory parts), which are however evaluated separately for each candidate. Moreover, calculating the information-theoretic term involves a costly determinant computation of the entire information (covariance) matrix which is $O(n^3)$ with $n$ being dimension of the state or costly Schur complement operations if only marginal posterior covariance of certain variables is of interest. 
Our approach, \ramdltree, extends our previous BSP method \ramdl \citep{Kopitkov17ijrr}, by exploiting incremental covariance calculation and performing calculation re-use between common parts of non-myopic candidate actions, such that these parts are evaluated only once, in contrast to  existing approaches. %Additionally, our approach has only a one-time calculation that depends on $n$, while evaluating action impact does not depend on $n$. 				
%Our approach, \ramdltree, re-uses calculations between common parts of candidate actions, that are evaluated only once, independently of the number of candidate actions that are sharing them, and has only a one-time calculation with dependence on $n$.
To that end, we represent all candidate actions together in a single unified graphical model, which we introduce and call a \emph{factor-graph propagation} (FGP) action tree.		
%By representing all candidates together in a single unified graphical model, which we introduce and call a \emph{factor-graph propagation} (FGP) action tree, we handle the actions' mutual parts only once, in contrast to the state of the art.
%\DK{[Last two sentences have the same message. Can we combine them into one?]} 
Each arrow (edge) of the FGP action tree represents a sub-action of one (or more) candidate action sequence and in order to evaluate its information impact we require specific covariance entries of an intermediate belief represented by tree's vertex 
from which the edge is coming out (e.g.~tail of the arrow).
%from which the edge is coming out. 
%		 \VI{[mention (in one sentence) the two passes in the tree to motivate why/what covariances are needed?]}
%\VI{To do so, we combine the presented herein incremental covariance update method with our previous BSP approach, \ramdl \citep{Kopitkov17ijrr} [Suggest to omit this sentence and instead  to mention \ramdl in the sentence above, where you already specify cov calculations and calculations re-use between common parts. Best to briefly distinguish btw this re-use of calculation than what is performed by \ramdl]}.
Overall, our approach has only a one-time calculation that depends on $n$, while evaluating action impact does not depend on $n$. We perform a careful examination of our approaches in simulation, considering the problem of autonomous navigation in unknown environments, where \ramdltree shows superior performance compared to \ramdl, while determining the same best actions.		
%\newline		
%Belief Space Planning (BSP) is computationally challenging in setting of high-dimensional state space. Many candidate actions need to be evaluated through objective function that has various terms, such as information-theoretic entropy and information gain, and may require determinant calculation of entire information (covariance) matrix which is $O(n^3)$ with $n$ being dimension of state. Moreover, in many cases the candidates themselves may have similar parts (e.g. common trajectory parts) which are typically evaluated separately for each candidate.
%\newline		
%Here we represents the planning problem via factor-graph formulation and provide BSP approach, \ramdltree, that has only one-time calculation with dependence on $n$, independently of candidates number. 		
%Moreover, by representing all candidates together in one unified graphical model, which we call \emph{factor-graph propagation} (FGP) action tree, we handle the actions' mutual parts only once, further reducing the computation requirements. 		
%		We present incremental covariance update which efficiently recovers specific covariance entries within the FGP action tree, and combine it with our previous BSP approach, \ramdl, in order to introduce the \ramdltree algorithm. Eventually, we perform a careful examination of our approach in simulation of autonomous navigation in unknown environment where \ramdltree shows superior performance.		

\end{abstract}

\keywords{Covariance recovery, belief space planning, active SLAM, informative planning, active inference, autonomous navigation}

	\section{Introduction}\label{Sec_Intro}

Autonomous operation in unknown or uncertain environments is a fundamental
problem in robotics and is an essential part in numerous applications such as autonomous
navigation in unknown environments, target tracking, search-and-rescue scenarios
and autonomous manufacturing. It requires both computationally efficient inference and planning approaches, where the former is responsible for tracking the posterior probability distribution function given available data thus far, and the latter is dealing with finding the optimal action given that distribution and a task-specific objective. 
%Planning involves decision making while accounting for different sources of uncertainty. 
Since the state is unknown and
only partially observable,  planning is performed in the belief space, where each
instance is a distribution over the original state, while accounting for different sources of uncertainty.
 Such a problem can be naturally viewed as a partially observable Markov decision process (POMDP), which
was shown to be computationally intractable and typically is solved by approximate approaches.
The planning and decision making problems are challenging both theoretically and computationally.
First, we need to accurately model future state belief as a function of future
action while considering probabilistic aspects of state sensing. Second, we need
to be able to efficiently evaluate utility of this future belief and to find an optimal
action, and to do so on-line. %real-time performance. 

The utility function in belief space planning (BSP) typically involves an information-theoretic term, which quantifies the posterior uncertainty of a future belief, and therefore requires access to the covariance (information) matrix of appropriate variables \citep{Kopitkov17ijrr}. Similarly, covariance of specific variables is also required in the inference phase, for example, in the context of data association \citep{Kaess09ras}. 
%Covariance of specific variables is often required by both inference and planning phases. For example, evaluation of future action, such as its impact on uncertainty of the state estimation, requires covariances of variables that are involved in the action \citep{Kopitkov17ijrr}, and data association problem requires covariances of nearby landmarks \citep{Kaess09ras}. 
However, the recovery of specific covariances is computationally expensive in high-dimensional state spaces:
while the belief is typically represented in the (square-root) information form to admit computationally efficient updates \citep{Kaess12ijrr}, retrieving the covariance entries requires an inverse of the corresponding (potentially) high-dimensional information matrix. Although sophisticated methods exist to efficiently perform such inverse by exploiting sparsity of square-root information matrix and by reordering state variables for better such sparsity \citep{Kaess12ijrr}, the overall complexity still is at least quadratic w.r.t. state dimension \citep{Ila15icra}. Moreover, in case of planning such computation needs to be performed for each candidate action.

The computational efficiency of the covariance recovery and the planning process is the main point of this paper. We develop a novel method to incrementally update covariance entries after any change of the inference problem, as defined next. Moreover, we present
a planning algorithm which leverages the key ability of incremental covariance updates and by exploiting action similarity 
is much faster and yet exact w.r.t. alternative
state-of-the-art techniques.

The inference problem can be represented by a set of currently available observations and state variables whose value we are to infer. For example, in a typical SLAM (simultaneous localization and mapping) scenario these variables are the robot poses along a trajectory and landmarks of the environment, while the observations are motion odometry and projection/range measurements. Covariances of the state variables can change as a result of any change in the inference problem, such as introduction of new observations or augmentation of state (e.g.~introduction of a new robot pose). Moreover, covariances also depend on current linearization point of the state vector, which in turn can also change after introduction of new observations. In this paper we scrupulously analyze each such possible change in the inference problem and show how covariance entries can be appropriately incrementally updated. Such capability to incrementally update covariance entries is important not only for the inference phase but also for efficiently addressing information-theoretic belief space planning, as we describe next.

BSP is typically solved by determining the best action, given an objective function,
from a set of candidate actions while accounting for different sources of uncertainty. Such an approach requires to evaluate the
utility of each action from a given set of candidate actions. 
%create the candidate set (by sampling, discretization, etc.) and to evaluate
%utility of each candidate action. 
This evaluation is usually done \emph{separately} for each candidate action and
typically consists of two stages. First, posterior belief for candidate action
is propagated and explicit inference is performed. Second, an
application-specific objective function is evaluated given candidate action and
the corresponding posterior belief. Yet, inference over the posterior belief and
evaluation of the objective function can be computationally expensive,
especially when the original state is high-dimensional since both parts'
complexity depends on its dimension.
% \VI{[relation to curse of dimensionality and curse of history?]}

%Typically, both belief propagation and objective calculation are highly
%expensive from time-consumption point of view, especially when the original
%state is high-dimensional since both parts' complexity depends on its
%dimension. 

\begin{figure}[t]
	\centering
	
	\subfloat{\includegraphics[width=0.9\textwidth]{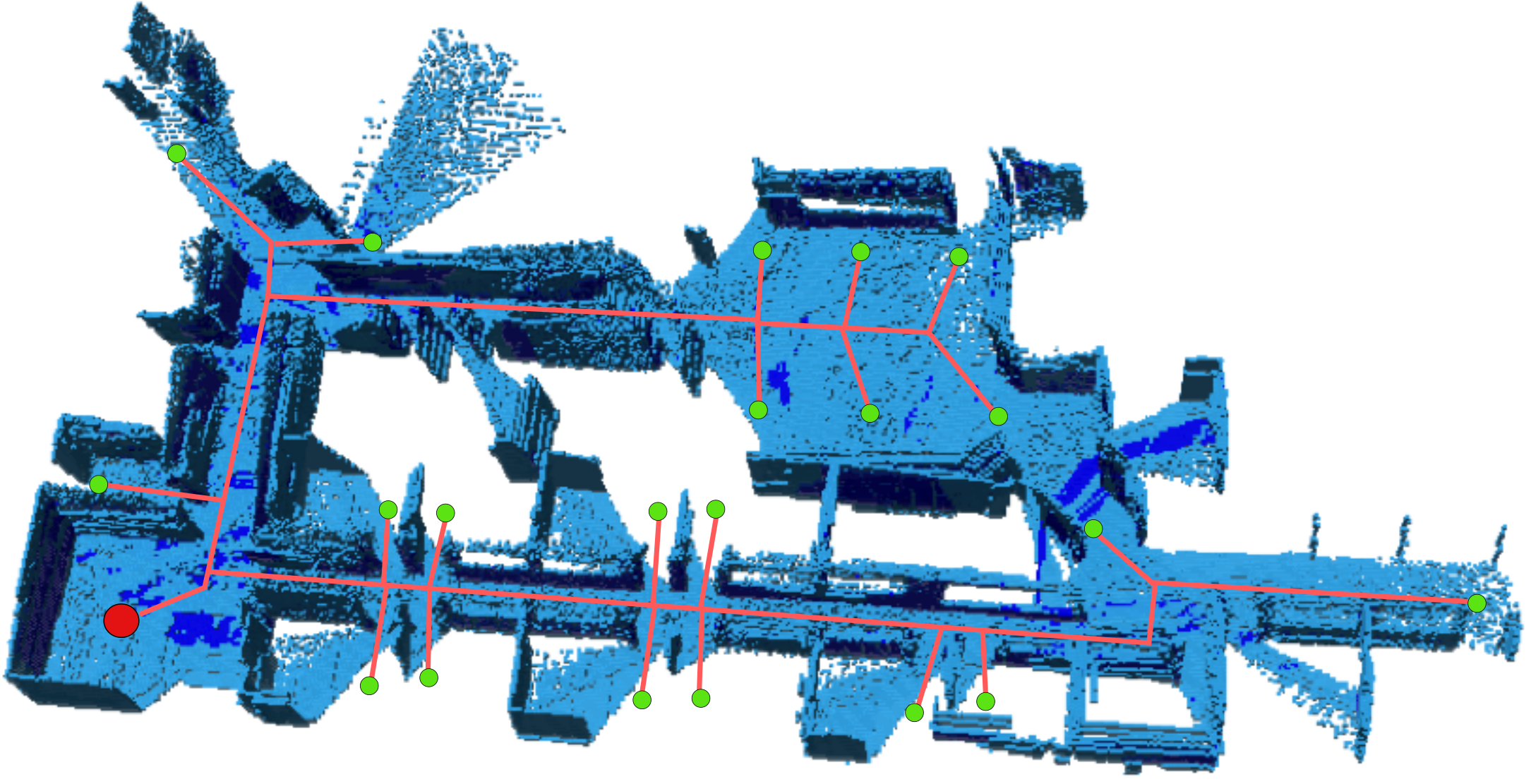}}
	\protect
	\caption{Illustration of possible candidate actions during exploration of an unknown environment by an autonomous robot. Robot's current position is marked by red circle; red lines and green points represent trajectory and final position of each candidate action respectively. As can be seen, actions share many mutual parts.
	}
	\label{fig:CandidatesSetSFig}
\end{figure}

However, in many BSP applications candidate (non-myopic) actions are partially overlapping, i.e.~have similar parts. For instance, in a building exploration scenario, candidate actions are trajectories to different locations in a building (see Figure \ref{fig:CandidatesSetSFig}) that were provided e.g.~by sampling-based motion planning approaches;
%predefined goal that were provided by trajectory sampling process,
some of these sampled trajectories will have mutual parts. Typically, these common parts will be evaluated a number of times, as part of evaluation of each action that shares it. Given that we know what are the similar parts between the different candidate actions, it can significantly reduce runtime complexity if we could handle these similar parts only once.

In this paper we present a technique for computation re-use between the candidate actions and exploitation of actions' similarity, while leveraging the above-mentioned method for incremental covariance updates. We show that such a technique greatly reduces the total decision making runtime. Moreover, we argue that for most cases, explicit inference over the
posterior belief is not required and that computation of the objective
function can be done efficiently with complexity that is independent of state
dimension. In general, the objective function of BSP can contain multiple terms,  such as
control cost, distance to  goal and an information-theoretic term (e.g.~entropy,
information gain or mutual information). Arguably, in typical settings the
control cost and distance to goal can be calculated without explicit inference
over the posterior belief, since these terms depend only on linearization point
of the state vector. In this paper we  show that also the information term does
not require an explicit inference over the posterior belief and that action similarity can be efficiently exploited, concluding that
BSP problem can be solved without performing time-consuming explicit inference
over the posterior belief at all. %\VI{[how much of this is attributed to this paper and not to rAMDL?]} \DK{Some. But we describe the new method and this paper in stand-alone way. Readers are not familiar with our old work.}

To that end, we present a new paradigm that represents all  candidate (sequence of) actions
in a single unified data structure that allows to exploit the similarities
between candidate actions while evaluating the impact of each such action. We
refer to this structure as \emph{factor-graph propagation} (FGP) action tree,
and show that the developed herein incremental covariance calculation method allows us to compute information gain of the tree's various parts. This, in turn, can be used to efficiently evaluate the information term of
different candidate actions while re-using calculations when possible. Combining our recently-developed \ramdl
approach \citep{Kopitkov17ijrr} with \emph{factor-graph
	propagation} (FGP) action tree and incremental covariance update, yields an
approach that calculates  action impact without explicitly performing inference
over the posterior belief, while re-using calculations among different candidate
actions.

%We will show how it can help us to evaluate the information term in a very
%efficient way. Combining our previous BSP approach \ramdl \citep{Kopitkov17ijrr} with new technique of representing candidates in
%\emph{factor-graph propagation} (FGP) action tree, we are succeeding to
%evaluate all the candidates fast by re-using the calculations and exploiting
%the fact that many candidates are sharing the same newly introduced
%measurements.

To summarize, our main contributions in this paper are as follows: 
(a) we develop an incremental covariance update method to calculate specific
covariance entries after any change in inference problem;
(b) we present \emph{factor-graph propagation} (FGP) action tree,
%with vertexes being prior, intermediate and posterior beliefs, 
that represents all candidate actions in single hierarchical model and allows
to formulate mutual parts of the actions as a single sub-actions; 
(c) we apply incremental covariance update method to calculate covariance entries from intermediate and posterior beliefs within the FGP action
tree, with complexity independent of state dimension;
and (d) we combine the FGP action tree graphical model, the incremental covariance
update method and \ramdl approach \citep{Kopitkov17ijrr} to yield a
new algorithm \ramdltree that efficiently solves an information-theoretic BSP
problem while handling candidates' mutual parts only once.

This paper is organized as following. In Section \ref{Sec_Rel_Work} we describe the relevant work done in this field. Section \ref{Sec_Notations} contains preliminaries and problem definition. In Section \ref{Sec_Approach}, we describe our approaches for incremental covariance recovery (Section \ref{sec:IncCovUpdate}) and information-theoretic BSP problem (Section \ref{sec:BSPApproachIntro}). Further, in Section \ref{sec:Results} we provide our simulation results that emphasize advantages of the presented herein approaches. Finally, in Section \ref{sec:Conclusions} we conclude the discussion about the introduced methods and point out several directions for future research. Additionally, at the end of this paper there is an appendix where we put proofs of several lemmas to improve readability.
%not obscure this paper's primary conceptual contribution.

% =============
\section{Related Work}\label{Sec_Rel_Work}
 
In this section we discuss the most relevant work to our approach, starting with computationally efficient covariance calculation and then proceeding to state of the art belief space planning approaches.

\subsubsection*{Computationally Efficient Covariance Recovery in High-Dimensional State Spaces}

Fast covariance recovery, under the Gaussian inference setting, is an active research area that has been addressed by several works in the recent years. Na\"{i}vely calculating an inverse of a high-dimensional information matrix is prohibitively expensive. However, these calculations can be avoided by exploiting sparsity of the square root information matrix, yielding a recursive method to calculate the required entries \citep{Golub80laa}, which has been recently also proposed by Kaess and Dellaert \citep{Kaess09ras} within their incremental smoothing and mapping solver.
%In \citep{Kaess09ras}, the authors present a recursive approach to calculate covariances from the square-root information matrix by exploiting its sparse structure. 
Although such method is faster than a simple inverse of square-root information matrix, the covariances are still calculated from scratch and the complexity depends on state dimension $n$. Moreover, in order to calculate a specific block of covariance matrix, the recursive approach may still need to calculate the entire covariance matrix (with dimensions $n \times n$) which is very undesirable for high-dimensional state spaces. 

More recently, Ila et al.~\citep{Ila15icra} introduced an approach to incrementally update covariances after the inference problem was changed. Given specific prior covariance entries that were calculated in previous timestep, their approach efficiently calculates covariance deltas to these entries, which comes out to be much faster than the recursive approach from \citep{Kaess09ras}. Although this approach is similar in spirit to our method of incremental covariance update, it is more limited in the following sense. Its theoretical part deals only with the specific scenario where new observations were introduced to the inference problem, without adding new variables. On the other hand, the mathematical formulation of their approach does not handle the common scenario where the state vector is augmented with new variables, 
%The scenario where state vector is augmented with new variables is not handled, 
although the simulation part of \citep{Ila15icra} suggests that their approach can also be applicable in this case in practice. We emphasize that this approach is not mathematically sound in the state augmentation case, since such a case involves singular matrices that are assumed to be invertible according to the derivation of \citep{Ila15icra}. Moreover, in case of state relinearization, the authors use a recursive method as fallback  and calculate covariances from scratch. In contrast, we present a general approach that is mathematically sound and is capable of dealing with any change in the inference problem, including state augmentation and relinearization. Moreover, even though a limited version of incremental covariance update has been developed \citep{Ila15icra}, it was not considered within a BSP problem, which is one of our main contributions in this work.

\subsubsection*{Belief Space Planning}

As was mentioned above, BSP is an instantiation of a POMDP problem. Optimal
solution of POMDP is known to be intractable \citep{Kaelbling98ai} in
high-dimensional domains due to curse of dimensionality. Therefore, most of the
modern research is focused on approximation methods that solve the planning
problem in sub-optimal form with tractable runtime complexity. These approximation
methods can be categorized into those that discretize the
state/action/measurement space domains and those that act in continuous spaces.	Approaches that perform discretization 
%	The "discretize" approaches 
include sampling \citep{Prentice09ijrr,
	AghaMohammadi14ijrr}, simulation \citep{Stachniss05rss} and point-based value iteration \citep{Pineau06jair} methods. Planning approaches that operate in continuous spaces, often also termed as direct trajectory optimization methods,
calculate a locally optimal solution given an initial nominal solution using different optimization techniques such as dynamic programming and gradient descent
%by typically	using information from the objective's gradient or sequential quadratic programming 
%	\citep{Indelman15ijrr,VanDenBerg12ijrr,Patil14wafr, Walls15iros, Platt10rss}.
\citep{Indelman15ijrr,VanDenBerg12ijrr,Patil14wafr, Platt10rss}.

Additionally, BSP methods can be separated into those that solve myopic and non-myopic decision making. While myopic approaches, also known as next best view (NBV) approaches in computer vision community (e.g.~\citep{Wenhardt07cvpr, Dunn09bmvc}), reason about actions taking the system only one step into the future, non-myopic planning %(e.g.~\citep{Platt10rss,He11jair,VanDenBerg12ijrr,Kim14ijrr,Chaves14iros,Indelman15ijrr}) 
(e.g.~\citep{Platt10rss,He11jair,VanDenBerg12ijrr,Kim14ijrr,Indelman15ijrr}) 
deals with sequences of actions taking the system multiple steps into the future. Clearly, for more complex tasks non-myopic methods will perform better as the time period before receiving the reward can be long. Yet, such methods are typically more computationally expensive as more effort is required to consider different probabilistic outcomes along the long planning horizon. In this paper we consider a non-myopic setting and formulate the problem through factor graphs. 

An information-theoretic BSP problem seeks for an optimal action that maximally
reduces estimation uncertainty of the state vector. Such a problem can be
separated into two main cases - \unfocused BSP tries to reduce uncertainty of
all variables inside the state vector, whereas \focused BSP is only interested to
reduce uncertainty of a predefined subset (termed as \focused variables) of these variables. Typically, the two
cases have different best
actions, with optimal action from \unfocused BSP potentially providing little information about \focused variables of \focused BSP (see e.g.~\citep{Levine13nips}). In both cases, the objective function usually calculates posterior entropy or information gain (of all variables from the state vector or of only \focused variables) and may have high computation complexity that depends on state dimension $n$. For instance, the calculation of \unfocused posterior entropy usually requires determinant computation of information (covariance) matrix which is in general $O(n^3)$, and is smaller for sparse matrices as in SLAM problems \citep{Bai96jcam}. Calculation of \focused posterior entropy is even more expensive and requires additional Schur complement computation.

Recently, we presented a novel approach, \ramdl \citep{Kopitkov17ijrr}, to efficiently calculate entropy and information gain for both \focused and \unfocused cases. This method requires only one-time calculation that depends on dimension $n$ - computation of specific prior marginal (or conditional) covariances. Given these prior covariances, \ramdl evaluates information impact of each candidate action independently of state dimension $n$. Such a technique was shown to significantly  reduce runtime (by orders of magnitude) compared to standard approaches.

Yet, in most  BSP approaches, including our own  \ramdl approach, the similarity between candidate actions is not exploited and each candidate is evaluated from scratch. To the best of our knowledge,  only the work by Chaves et al. \citep{Chaves16iros} was done in this direction. Their approach performs fast explicit inference over the posterior belief, by constraining variable ordering of the Bayes tree data structure to have candidates' common variables eliminated first. Still, this approach has its limitations. It explicitly calculates the posterior belief for each action, and though it is done fast, it still requires additional memory to store such posterior beliefs. Further, it does not deal with information-theoretic objective functions whose runtime complexity is usually very expensive, as mentioned above. Moreover, it can only be applied when the SLAM algorithm is implemented using Bayes tree \citep{Kaess12ijrr}, and it was shown to work only for the case where actions are trajectories constrained to have only a single common part.

In contrast, in this paper we develop a BSP technique that re-uses calculations in a general way, by exploiting potentially any number of mutual parts between the candidate actions. It is expressed in terms of factor graphs and can be applied not just for trajectory planning, but for any decision making problem expressed via factor graphs. Moreover, our technique can be implemented independently of a chosen SLAM factor graph optimization infrastructure. We combine several algorithmic concepts together - a unified graphical model FGP action tree, incremental covariance update and \ramdl
approach \citep{Kopitkov17ijrr}, and present a BSP solution that does not require explicit inference over the posterior belief while carefully evaluating information impact of each action in an exact way.

% ===============
\section{Notations and Problem Formulation}
\label{Sec_Notations}

\begin{table}
	\centering
	
	\begin{tabular}{|c|c|}
		\hline
		& \\[-8pt]
		\textbf{Notation} & \textbf{Description} \tabularnewline
		\hline
		& \\[-8pt]
		%		\hline
		$X_{-}$ & \textbf{Problem 1}: state vector before a change in inference problem;\\
		& \textbf{Problem 2}: state vector at planning time \tabularnewline
		%		\hline		
		& \\[-8pt]
		$X_{+}$ & \textbf{Problem 1}: state vector after a change in inference problem;\\
		&  \textbf{Problem 2}: future state vector after applying a specific candidate action \tabularnewline	
		%		\hline		
		& \\[-8pt]
		$X_{new}$ & \textbf{Problem 1}: new state variables introduced after a change in inference problem;\\
		&  \textbf{Problem 2}: new state variables introduced after  applying a specific candidate action \tabularnewline	
		%		\hline		
		& \\[-8pt]
		$F_{new}$ & \textbf{Problem 1}: new factors set introduced after a change in inference problem;\\
		&  \textbf{Problem 2}: new factors set introduced after  applying a specific candidate action \tabularnewline	
		%		\hline		
		& \\[-8pt]
		$\Lambda_{-}$ and $\Lambda_{+}$ & prior and posterior information matrices  \tabularnewline
		%		\hline
		& \\[-8pt]
		$\Lambda_{+}^{Aug}$ & prior information matrix $\Lambda_{-}$ augmented with zero rows and
		columns\\
		& that represent new state variables $X_{new}$ (see Figure
		\ref{fig:AugmentInfoMAtBSFig})  \tabularnewline
		%		\hline
		& \\[-8pt]
		$\Sigma_{-}$ and $\Sigma_{+}$ & prior and posterior covariance matrices  \tabularnewline
		%		\hline
		& \\[-8pt]
		$R_{-}$ and $R_{+}$ & prior and posterior square-root information upper-triangular matrices  \tabularnewline
		%		\hline
		& \\[-8pt]
		$b[X]$ & belief of state vector $X$ %(its estimation)
		 \tabularnewline		
		%		\hline
		& \\[-8pt]
		$\mathcal{H}(b[X])$ & differential entropy of  belief $b[X]$ \tabularnewline		
		%		\hline
		& \\[-8pt]
		$\Sigma^{M,Y}$ & marginal covariance of some state subset $Y$\\
		 & (partition of covariance matrix $\Sigma$ with columns$\backslash$rows belonging to $Y$) \tabularnewline	
		%		\hline		
		& \\[-8pt]
		$I(a)$ & \increment of candidate action $a$, represents new factors and new state variables\\
		& introduced into inference problem after $a$ is executed \tabularnewline	
		%		\hline		
		& \\[-8pt]
		$\mathcal{A}$ & set of candidate actions considered in BSP \tabularnewline	
		%		\hline		
		& \\[-8pt]
		$A$ & noise-weighted Jacobian of newly introduced factors
		w.r.t. all state variables \\[-10pt]
		& \\
		\hline				
	\end{tabular}\protect\caption{\label{tab:PFNotations} Main notations used within problem definition (Section \ref{Sec_Notations}).}
\end{table}

Consider a high-dimensional problem-specific state vector
$X_{-}\in\mathbb{R}^{n}$ at the current time, where we use the notation "-" to
represent the (a priori) state vector before applying the next action. Depending on the application, $X_{-}$ can represent robot configuration and poses (optionally
also past and current poses), environment-related variables or any other
variables of interest. Additionally, consider factors $F =
\{f^{1}(X^{1}),\ldots,f^{n_f}(X^{n_f}) \}$ that were added to the inference problem
till (and including) current time, where each factor $f^{j}(X^{j})$  represents
a specific measurement model, motion model or prior, and as such involves
appropriate state variables $X^{j} \subseteq X_{-}$.

The joint pdf (probability density function) can be then written as
\begin{equation}
\probb{X_{-}|history} \propto
\prod_{j=1}^{n_f}
f^{j}(X^{j}), 
\label{eq:FactorPDF}
\end{equation}
where $history$ contains all the information gathered by the current time
(measurements, controls, etc.).

As common in many inference problems, we will assume that all factors have a
Gaussian form:
\begin{equation}
f^{j}(X^{j}) \propto 
\exp (- \half \Vert h^{j}(X^{j}) - r^{j} \Vert_{\Sigma^{j}}^{2}),
\label{eq:FactorModel}
\end{equation}
with appropriate model
\begin{equation}
r^{j} = h^{j}(X^{j}) + \upsilon^{j}, \quad
\upsilon^{j} \sim \mathcal{N}(0, \Sigma^{j}),
\label{eq:FactorModel2}
\end{equation}
where $h^j$ is a known nonlinear function, $\upsilon^{j}$ is zero-mean Gaussian
noise and $r^j$ is the expected value of $h^j$, i.e.~$r^j =
\mathbb{E}[h^{j}(X^{j})]$. Such a factor representation is a general way to
express information about the state. In particular, it can represent a
measurement model, in which case, $h^j$ is the observation model, and  $r^j$ and
$\upsilon^j$ are, respectively, the actual measurement $z$ and measurement
noise. Similarly, it can also represent a motion model. A maximum a posteriori
(MAP) inference is the optimization solution of maximizing Eq.~(\ref{eq:FactorPDF}) w.r.t. state $X_{-}$. It can be efficiently calculated (see e.g.~\cite{Kaess12ijrr}) such
that
\begin{equation}
\probb{X_{-}|history} = \mathcal{N}(X_{-}^{\star}, \Sigma_{-})
= \mathcal{N}^{-1}(\eta_{-}^{\star}, \Lambda_{-})
\end{equation}
where $X^{\star}_{-}$, $\Sigma_{-}$, $\eta^{\star}_{-}$ and $\Lambda_{-}$ are respectively the current mean
vector, covariance matrix, information vector and information matrix (inverse of covariance matrix).

We shall refer to the belief $\probb{X_{-}|history}$ of state $X_{-}$ at the
current time as the prior \belief and write
\begin{equation}
b[X_{-}]\doteq \probb{X_{-}|history}.
\end{equation}
\noindent We now introduce the two problems this paper addresses, along with appropriate notations:  general purpose incremental covariance update, and computationally efficient belief space planning. As will be seen in Section \ref{Sec_Approach},  our approach to address the latter problem builds upon the solution to the first problem.

\begin{figure}[!t]
	\centering
	\begin{tabular}{ccccccc}
		
		\subfloat[\label{fig:SLAMExFig-a}]{\includegraphics[width=0.42\textwidth]{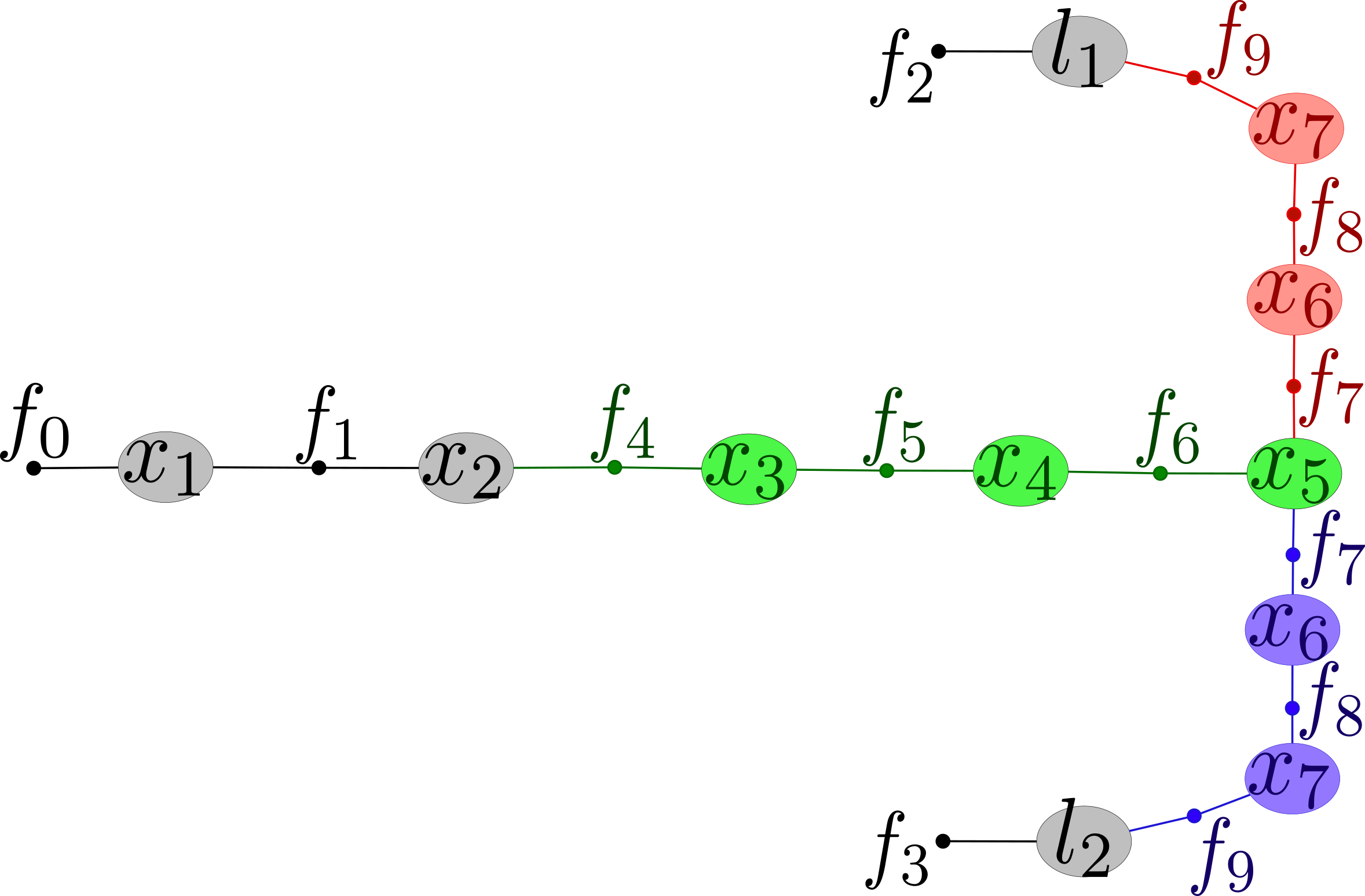}}
		&
		\hspace{15pt} &
		
		\subfloat[\label{fig:SLAMExFig-b}]{\includegraphics[width=0.19\textwidth]{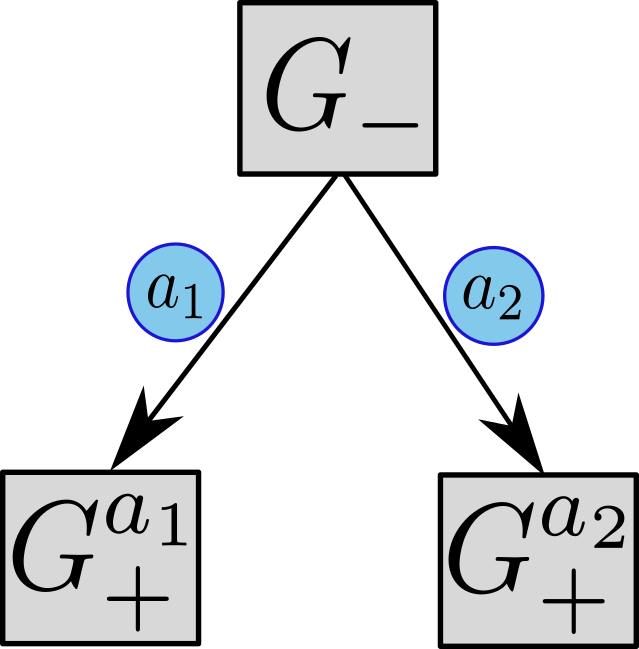}}
		&
		\hspace{15pt} &
		
		\subfloat[\label{fig:SLAMExFig-c}]{\includegraphics[width=0.19\textwidth]{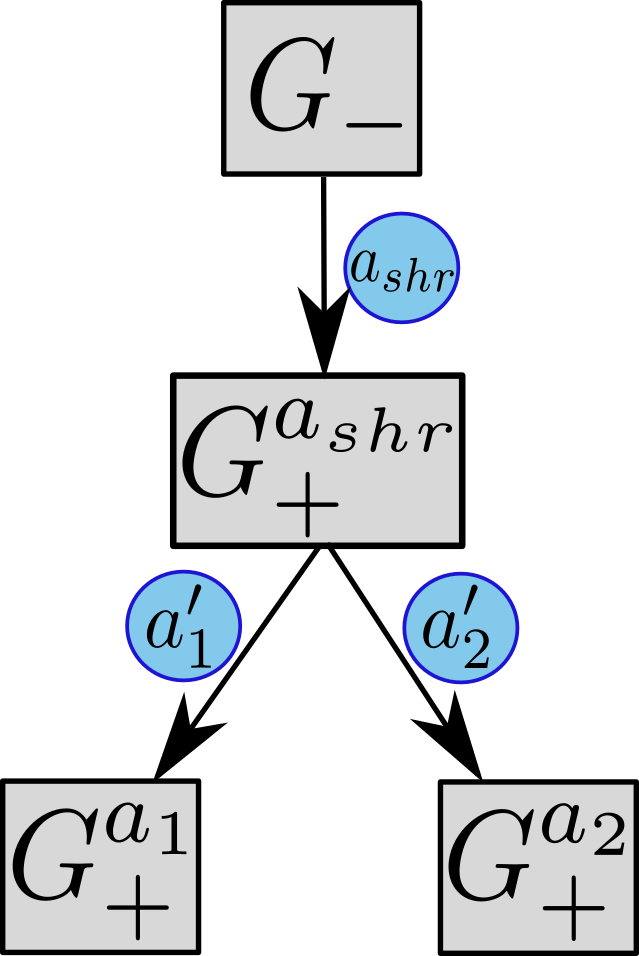}}
		\\
	\end{tabular}
	
	\protect
	\caption{Illustration of belief propagation in factor graph representation,
		taken from SLAM application.
		(a) Prior factor graph $G_{-}$ (colored in black) contains two robot poses $x_1$
		and $x_2$, and two landmarks $l_1$ and $l_2$, as also the prior and motion model
		factors $\{ f_0, f_1, f_2, f_3 \}$. Two different actions (trajectories) are
		considered. The first will take the robot to observe landmark $l_1$ and will
		augment the $G_{-}$ with new factors and state variables colored in green and
		red. The second will take robot to visit $l_2$ instead, and will augment the
		$G_{-}$ with new factors and state variables colored in green and purple. As can
		be seen, both candidate actions are sharing some of their new factors/state
		variables (colored in green).
		(b)-(c) Candidates from (a) represented as FGP action tree. In FGP action tree
		each vertex represents a specific factor graph (or the belief that is associated
		with it) and each edge represents a specific action - augmentation of the  factor graph
		with an \increment of the action, see Eq.~(\ref{eq:IncrementDef}). 
		(b): posterior factor graphs $G_{+}^{a_1}$ and $G_{+}^{a_2}$ are propagated
		separately for each action.
		(c): first the prior factor $G_{-}$ is augmented by a mutual \increment
		represented by $a_{shr}$ (colored green in (a)) and the posterior
		$G_{+}^{a_{shr}}$ is received. Next,  $G_{+}^{a_{shr}}$ is augmented
		separately by not-mutual \increments (colored red and purple in (a)) of each
		action, providing posterior factor graphs $G_{+}^{a_1}$ and $G_{+}^{a_2}$.
	}
	\label{fig:SLAMExFig}
\end{figure}
%

%%
%\begin{figure}[!t]
%	\centering
%
%\subfloat{\includegraphics[width=0.96\textwidth]{4.png}}
%	\protect
%	\caption{Illustration of belief propagation in factor graph representation.
%Two actions $a_i$ and $a_j$ are considered, introducing their own factor graphs
%$G(a_i)$ and $G(a_j)$ (colored in pink) that are \emph{connected} to prior
%factor graph $G_{-}$ through factor sets $\mathcal{F}^{conn}(a_i)$ and
%$\mathcal{F}^{conn}(a_j)$ (colored in green) respectively.}
%	\label{fig:AugCasePropFig}
%\end{figure}
%%
%
%

\subsubsection*{Problem 1: Covariance Recovery}

As  mentioned above, in many applications it is mandatory to recover covariance entries of belief $b[X_{-}]$. However, typically this belief is represented through its information form ($\eta_{-}^{\star}$ and $\Lambda_{-}$), or the square-root information upper-triangular matrix $R_{-}$, with $\Lambda_{-} = R_{-}^{T} \cdot R_{-}$.

%\VI{[Suggest to specify additional details here that fully describe the problem of incremental covariance update/recovery: given a change in inference system (needs to be defined), and previously calculated covariances, you are seeking to calculate efficiently covariances of the updated system. ]}

Considering a square-root representation, the covariance matrix is $\Sigma_{-} = R_{-}^{-1} \cdot R_{-}^{-T}$ and its specific covariance entries $\Sigma_{-} = (\sigma_{ij})$ can be calculated from entries $R_{-} = (r_{ij})$ as \citep{Golub80laa}
\begin{equation}
\sigma_{ii} = \frac{1}{r_{ii}}
\Bigg(
\frac{1}{r_{ii}} - 
\sum_{j = i + 1, r_{ij} \neq 0}^{n}
r_{ij} \sigma_{ji}
\Bigg),
\label{eq:RecursCovRec1}
\end{equation}
\begin{equation}
\sigma_{ij} = \frac{1}{r_{ii}}
\Bigg(
- \sum_{k = i + 1, r_{ik} \neq 0}^{j}
r_{ik} \sigma_{kj}
- \sum_{k = j + 1, r_{ik} \neq 0}^{n}
r_{ik} \sigma_{kj}
\Bigg).
\label{eq:RecursCovRec2}
\end{equation}
Note that in order to calculate the upper left covariance entry ($\sigma_{11}$), all other covariance entries are required. Therefore, worst case computation (and memory) complexity of this recursive approach is still quadratic in state dimension $n$.

In contrast, an incremental covariance update approach can be applied in order to recover the required covariance entries  more efficiently. At each timestep, solving the inference problem for the current belief $b[X_{-}]$ from Eq.~(\ref{eq:FactorPDF}) provides MAP estimate and the corresponding covariance or (square root) information matrix. However, at the next step the inference problem changes. 
 %At each timestep, the inference problem, which is represented by Eq.~(\ref{eq:FactorPDF}) and provides a MAP estimation $X^{\star}_{-}$ in form of current belief $b[X_{-}]$ and its covariance matrix, is changed.
 To see that, consider the belief at the next timestep $b[X_{+}]$ which was obtained by introducing new state variables $X_{new}$ (e.g. new robot poses in SLAM smoothing formulation), with $X_{+} = X_{-} \cup X_{new}$, and by adding new factors (e.g. new measurements, odometry, etc.) $F_{new} =
\{f_{new}^{1}(X_{+}^{1}),\ldots,f_{new}^{n_{new}}(X_{+}^{n_{new}}) \}$ where
$X_{+}^{j} \subseteq X_{+}$. Additionally, consider the set of variables $Y \subset X_{+}$ whose marginal covariance $\Sigma_{+}^{M,Y}$ from $b[X_{+}]$ we are interested in calculating. Note that these variables of interest can contain both old $Y_{old} \subset X_{-}$ and new $Y_{new} \subset X_{new}$ variables, with $Y = \{ Y_{old}, Y_{new} \}$.

Given that we already calculated the required covariance entries $\Sigma_{-}^{M,Y_{old}}$ from the current belief $b[X_{-}]$, in incremental covariance update approach we would like to update these entries after the change in the inference problem (from $b[X_{-}]$ to $b[X_{+}]$) as:
\begin{equation}
\Sigma_{+}^{M,Y_{old}} = \Sigma_{-}^{M,Y_{old}} + \Delta^{Y_{old}},
\label{eq:incrCovUpdAdditive}
\end{equation}
where $\Delta^{Y_{old}}$ represents the difference between old and new covariance entries. Additionally, in the general  case we might be interested in calculating the posterior covariance of new variables of interest $Y_{new} \subset Y$, i.e.~ $\Sigma_{+}^{M,Y_{new}}$, as well as also the cross-covariances between $Y_{old}$ and $Y_{new}$.

Likewise, also the conditional covariances, from the conditional pdf of one state subset conditioned on another, are required for information-theoretic BSP as was shown in \citep{Kopitkov17ijrr}. Hence, we would also like to develop a similar approach for incremental conditional covariance update.

A limited technique to perform an incremental update of marginal covariances was presented in \citep{Ila15icra}. The authors show how to update the covariance entries by downdating the posterior information matrix. Their derivation can be applied for the case where the state vector was not augmented during the change in the inference problem ($X_{new}$ is empty). However, that derivation is not valid for the case of state augmentation, which involves zero-padding of prior matrices (described below); such padding yields singular matrices and requires a more delicate handling. Even though their approach is not mathematically sound for the augmentation case, in the simulation part of \citep{Ila15icra} it is insinuated that the approach can also be applied here in practice. Still, the authors clearly declare that their approach does not handle relinearization of the state vector, which can often happen during the change in the inference problem. Further, \citep{Ila15icra} does not consider recovery of conditional covariances. In contrast, we develop a general purpose method that handles incremental (marginal and conditional) covariance updates in all of the above cases in a mathematically sound way.

In Section \ref{Sec_Approach} we categorize the above general change in the inference problem into different sub-cases. Further, we present an approach that
carefully handles each such sub-case
and incrementally updates covariances that were already calculated before the change in the inference problem, and also computes covariance of newly introduced state variables. As will be shown, the computational complexity of such a method, when applied to a problem where only the marginal covariances need to be recovered (i.e.~block diagonal of $\Sigma_{-}$), is linear in $n$ in the worst case. Furthermore, we will show how our incremental covariance update approach can be also applied  to incrementally update conditional covariance entries. Later, this capability will be an essential part in the derivation of our BSP method, \ramdltree.

\subsubsection*{Problem 2: Belief Space Planning}

Typically in BSP and decision making problems we have a set of candidate
actions $\mathcal{A}=\{a_1,a_2,...\}$ from which we need to pick the best action
according to a given objective function.
As shown in our previous work \citep{Kopitkov17ijrr}, the posterior belief for each action
can be viewed as a specific augmentation of the prior factor graph that represents
the prior \belief $b[X_{-}]$ (see Figure \ref{fig:SLAMExFig-a}). In this paper
we shall denote this factor graph by $G_{-}$. Each candidate action $a$ can add new
information about the state variables in form of new factors. Additionally, in
specific applications, action $a$ can also introduce new state variables into
the factor graph (e.g. new robot poses). Thus, similarly to change in inference problem described above for each action $a$ we can model
the newly introduced state variables denoted by $X_{new}$, defining the
posterior state vector (after applying the action) as $X_{+} = X_{-} \cup
X_{new}$. In a like manner, we denote the newly introduced factors by $F_{new} =
\{f_{new}^{1}(X_{+}^{1}),\ldots,f_{new}^{n_{new}}(X_{+}^{n_{new}}) \}$ where
$X_{+}^{j} \subseteq X_{+}$.

Therefore, similar to Eq.~(\ref{eq:FactorPDF}), after applying  candidate action
$a$, the posterior \belief $b[X_{+}]$ can be explicitly written as 
\begin{equation}
b[X_{+}] \propto
b[X_{-}]
\prod_{j=1}^{n_{new}}
f_{new}^{j}(X_{+}^{j}).
\label{eq:FactorBeliefPropogation}
\end{equation}
Such a formulation is  general and supports non-myopic action $a$ with any
planning horizon, that introduces into the factor graph multiple new state
variables and multiple factors with any measurement model. Still, in this paper
we assume factors to have a Gaussian form (Eq.~(\ref{eq:FactorModel})).

\begin{figure}[t]
	\centering
	
	\subfloat{\includegraphics[width=0.6\textwidth]{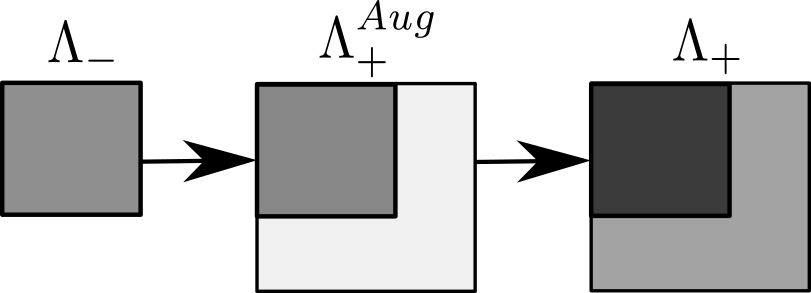}}
	\protect
	\caption{Illustration of $\Lambda_{+}$'s construction for a given candidate
		action in case new state variables $X_{new}$ were introduced into the state vector. First, $\Lambda_{+}^{Aug}$ is created by adding
		zero rows and columns representing the new state variables. Then, the new
		information of belief is computed through $\Lambda_{+} = \Lambda_{+}^{Aug} + A^T
		A$.
	}
	\label{fig:AugmentInfoMAtBSFig}
\end{figure}

For the sake of conciseness, in this paper the newly introduced factors and state
variables that are added when considering action $a$ will be called action $a$'s
\increment and denoted as 
\begin{equation}
I(a) \doteq \{ F_{new}, X_{new} \}
\label{eq:IncrementDef}
\end{equation}
The posterior information matrix, i.e.~the second moment of the belief
$b[X_{+}]$, can be written as
\begin{equation}
\Lambda_{+} = 
\Lambda_{-} + A^T \cdot A
\quad , \quad
\Lambda_{+} = 
\Lambda_{+}^{Aug} + A^T \cdot A
\label{eq:FactorPosteriorInfoMatrixBSDM}
\end{equation}
where we took the maximum likelihood assumption which considers that the above, a single optimization iteration (e.g.~Gauss Newton), sufficiently
captures action impact on the belief. Such an assumption is typical in
BSP literature (see, e.g.~\citep{Platt10rss, VanDenBerg12ijrr, Kim14ijrr,
	Indelman15ijrr}).
%meaning that the linearization point of state variables does not change after
%applying the action, it is not hard to show that the information matrix of
%$b[X_{+}]$ can be calculated as:
The left identity in Eq.~(\ref{eq:FactorPosteriorInfoMatrixBSDM}) is true when
$X_{new}$ is empty, while the right identity is valid for non-empty $X_{new}$.
The matrix $A$ is a noise-weighted Jacobian of newly introduced factors
$F_{new}$ w.r.t. state variables $X_{+}$;  $\Lambda_{+}^{Aug}$ is constructed by
first augmenting the prior information matrix $\Lambda_{-}$ with zero rows and
columns representing the new state variables $X_{new}$, as illustrated in Figure
\ref{fig:AugmentInfoMAtBSFig} (see e.g.~\citep{Kopitkov17ijrr}).

After modeling the posterior information matrix $\Lambda_{+}$ for action $a$,
the \unfocused information gain (uncertainty reduction of the entire state
vector $X_{+}$) can be computed as:
\begin{equation}
J_{IG}(a) =
\mathcal{H}(b[X_{-}]) - \mathcal{H}(b[X_{+}])
=
dim.const +
\frac{1}{2} \ln \frac{ \begin{vmatrix} \Lambda_{+} \end{vmatrix} }
{\begin{vmatrix} \Lambda_{-} \end{vmatrix}}
\label{eq:EntropyPriorDM}
\end{equation}
where $\mathcal{H}(\cdot)$ is differential entropy function that measures the uncertainty of
input belief, and $dim.const$ is a constant that only depends on the dimension
of $X_{+}$ and thus is ignored in this paper. Note that the above \unfocused information gain is typically used in applications where the set of new variables, $X_{new}$, is empty and so both $X_{-}$ and $X_{+}$ have the same dimension. In cases where $X_{new}$ is not empty (e.g. SLAM smoothing formulation), usually \focused information gain is used (see below).

The optimal action $a^*$ is then given by
 $a^* = \argmax_{a \in \mathcal{A}}
J_{IG}(a)$.

For \focused BSP problem we would like to reduce uncertainty of only a  subset
of state variables $X^F \subseteq X_{+}$. When $X^F$ consists of old variables
$X_{-}$, $X^F \subseteq X_{-}$, we can compute its information gain (IG). Such
IG is a reduction of $X^F$'s entropy after applying action $a$,
$\mathcal{H}(b[X_{-}^{F}]) - \mathcal{H}(b[X_{+}^{F}])$ where $b[X_{-}^{F}]$ and
$b[X_{+}^{F}]$ are prior and posterior beliefs of \focused variables $X^F$. In 
case $X^F$ consists of newly introduced variables $X_{new}$, $X^F \subseteq
X_{new}$, the IG function has no meaning as the prior belief $b[X_{-}^{F}]$ does
not exist. Instead, we can directly calculate $X^F$'s posterior entropy
$\mathcal{H}(b[X_{+}^{F}])$. The IG and entropy functions can be calculated
through respectively:
\begin{equation}
J_{IG}^F(a) = 
\frac{1}{2} \ln \frac{ \begin{vmatrix} \Sigma_{-}^{M,F} \end{vmatrix} }
{\begin{vmatrix} \Sigma_{+}^{M,F} \end{vmatrix}}
, \quad
J_{\mathcal{H}}^F(a) =
dim.const + 
\frac{1}{2} \ln \begin{vmatrix} \Sigma_{+}^{M,F} \end{vmatrix},
\label{eq:ObjFuncPosteriorDeterminantFocused}
\end{equation}
where $\Sigma_{-}^{M,F}$ and $\Sigma_{+}^{M,F}$ are prior and posterior marginal
covariance matrices of $X^F$, respectively. Note that in \focused BSP the
optimal action will be found through $a^* = \argmax_{a \in \mathcal{A}}
J_{IG}^{F}(a)$ or $a^* = \argmin_{a \in \mathcal{A}} J_{\mathcal{H}}^F(a)$.

To summarize, in order to solve an information-theoretic BSP problem, we are required to
calculate IG or entropy (Eqs.~(\ref{eq:EntropyPriorDM}) and
(\ref{eq:ObjFuncPosteriorDeterminantFocused})) for each candidate action $a$,
and then choose a candidate action with the maximal gain.

% ===============
\section{Approach}\label{Sec_Approach}

In this section we present our approaches that efficiently solve the incremental covariance recovery (Section \ref{sec:IncCovUpdate}) and information-theoretic BSP (Section \ref{sec:BSPApproachIntro}).

% ==================
\subsection{Incremental Covariance Update}\label{sec:IncCovUpdate}

In this section we present our technique for efficient update of covariance entries (see \textbf{Problem 1} in Section \ref{Sec_Notations}). In Section \ref{sec:FFuncPresent} we  show how to update marginal covariances of specified variables $Y \subset X_{+}$ after new state variables were introduced into the state vector and new factors were added, yet no state relinearization happened during the change in the inference problem. We will show that the information matrix of the entire state belief is propagated through \emph{quadratic} update form, similarly to Eq.~(\ref{eq:FactorPosteriorInfoMatrixBSDM}). Assuming such \emph{quadratic} update, we will derive a method to efficiently calculate the change in old covariance entries, to compute the new covariance entries and the cross-covariances between old and new state variables. Further, in Section \ref{sec:ReleanFFuncPresent} we will show that also in the relinearization case the information matrix update has an identical \emph{quadratic} update form and conclude that our method, derived in Section \ref{sec:FFuncPresent}, can also be applied when some of the state variables were relinearized. Finally, in Section \ref{sec:CondFFuncPresent} we will show that also the information matrix of a conditional pdf is updated through \emph{quadratic} update form and that the same technique from Section \ref{sec:FFuncPresent} can be applied in order to incrementally update conditional covariance entries. We will show that our approach's complexity, given the specific prior covariances, does not depend on state dimension $n$.

\subsubsection{Update of Marginal Covariance Entries}
\label{sec:FFuncPresent}

\begin{table}
	\centering
	
	\begin{tabular}{|c|c|}
		\hline
		& \\[-8pt]
		\textbf{Notation} & \textbf{Description} \tabularnewline
		\hline
		& \\[-8pt]
		%		\hline
		$Y$ & subset of state variables whose marginal covariance we are interested to update$\backslash$compute  \tabularnewline
		%		\hline		
		& \\[-8pt]
		$Y_{old}$ & old variables inside $Y$ \tabularnewline	
		%		\hline		
		& \\[-8pt]
		$Y_{new}$ & new variables inside $Y$ (that were introduced during the change in the inference problem) \tabularnewline	
		%		\hline		
		& \\[-8pt]
		$\comb{X}{I}{}{}$ & set of old \involved variables in the newly introduced factors $F_{new}$ \tabularnewline	
		%		\hline		
		& \\[-8pt]
		$W$ & variable union of sets $Y_{old}$ and $\comb{X}{I}{}{}$ \tabularnewline	
		%		\hline		
		& \\[-8pt]
		$n$ & dimension of a prior state vector $X_{-}$  \tabularnewline
		%		\hline
		& \\[-8pt]
		$N$ & dimension of a posterior state vector $X_{+}$  \tabularnewline
		%		\hline
		& \\[-8pt]
		$m$ & overall dimension of the newly introduced factors $F_{new}$  \tabularnewline
		%		\hline
		& \\[-8pt]
		$\comb{A}{I}{}{}$ & $m \times |\comb{X}{I}{}{}|$ matrix that consists of $A$'s columns belonging to variables in $\comb{X}{I}{}{}$ \\[-10pt]
		& \\
		\hline				
	\end{tabular}\protect\caption{\label{tab:IncrCovNotations} Main notations used through derivation of incremental covariance recovery approach.}
\end{table}

Consider \textbf{Problem 1} from Section \ref{Sec_Notations}. Consider the belief was propagated from $b[X_{-}]$ to $b[X_{+}]$ as described. Yet, let us assume for now that no state relinearization happened (we will specifically handle it in the next section). In this section we show that the posterior covariances of interest $\Sigma_{+}^{M,Y}$ can be efficiently calculated as
\begin{equation}
\Sigma_{+}^{M,Y} = f(\Sigma_{-}^{M,W}),
\end{equation}
where $\Sigma_{-}^{M,W}$ is the prior marginal covariance of set $W \doteq \{ Y_{old}, \comb{X}{I}{}{}\}$ and $f(\cdot)$ is a transformation function, with calculation complexity that does not depend on state dimension $n$. We derive this function in detail below. The set $Y_{old}$ contains old state variables inside $Y$ ($Y_{old} \subseteq X_{-}$) and  $\comb{X}{I}{}{} \subseteq X_{-}$ is the set of \involved variables in the newly introduced factors $F_{new}$ - variables that appear in $F_{new}$'s models (Eq.~(\ref{eq:FactorModel2})). Note that the update of old covariances (Eq.~(\ref{eq:incrCovUpdAdditive})) is only one part of this $f(\cdot)$, as also the computation of covariances for new state variables $Y_{new}$ and cross-covariances between $Y_{old}$ and $Y_{new}$.

%First, let us redefine the problem. Given a priori belief $b[X_{-}]$ with prior information matrix $\Lambda_{-} \in \mathbb{R}^{n \times n}$, the candidate action $a$ with an \increment $I(a) \doteq \{ F_{new}, X_{new} \}$ is applied and the posterior belief $b[X_{+}]$ is obtained. Consider the set of variables $Y \subset X_{+}$ whose marginal covariance $\Sigma_{+}^{M,Y}$ from $b[X_{+}]$ we would like to calculate. We are looking for function $\Sigma_{+}^{M,Y} = f(\Sigma_{-}^{M,W})$ where $\Sigma_{-}^{M,W}$ is prior marginal covariance of set $W \doteq \{ Y_{old}, \comb{X}{I}{}{}\}$. The variables set $Y_{old}$ is intersection $X_{-} \cap Y$, or in other words the old variables inside $Y$. The $\comb{X}{I}{}{} \subseteq X_{-}$ is the set of \involved variables in action's newly introduced factors $F_{new}$. Moreover, we are looking for function $f()$ whose complexity does not depend on state dimension $n$.

Next, let us separate all possible changes in the inference problem into different cases according to augmented state variables $X_{new}$ and the newly introduced factors $F_{new}$.

If $X_{new}$ is empty, we will call such a case as \actnaug. This case does not change the state vector ($X_{-} \equiv X_{+}$) and only introduces new information through new factors. The information matrix in this case can be updated through $\Lambda_{+} = 
\Lambda_{-} + A^T \cdot A$, where matrix $A \in \mathbb{R}^{m \times n}$ is a noise-weighted Jacobian of newly introduced factors $F_{new}$ w.r.t. state variables $X_{+}$, and $A$'s height $m$ is dimension of all new factors within $F_{new}$ (see Section \ref{Sec_Notations}).

Given $X_{new}$ is not empty, we will call such a case as \actrect. This case augments the state vector to be $X_{+} = \{ X_{-}, X_{new} \}$ and also introduces new information through the new factors. Here the information matrix can be updated through $\Lambda_{+} = 
\Lambda_{+}^{Aug} + A^T \cdot A
$ where $\Lambda_{+}^{Aug} \in \mathbb{R}^{N \times N}$ is a singular matrix that is constructed by first augmenting the prior information matrix $\Lambda_{-}$ with zero rows and columns representing the new state variables $X_{new}$, as illustrated in Figure \ref{fig:AugmentInfoMAtBSFig}; $N = |X_{+}| = n + |X_{new}|$ is the posterior state dimension; $A$ here will be an $m \times N$ matrix.

Finally, for the case when $X_{new}$ is not empty and total dimension of new factors $m$ is equal to the number of newly introduced variables $|X_{new}|$, we will call such a case as \actsqr. Clearly, the \actsqr case is a specific case of the \actrect case, which for instance can represent the new robot poses of candidate trajectory and the new motion model factors. The reason for this specific case to be dealt with in special way is due to the fact that its $f(\cdot)$ function is much simpler than $f(\cdot)$ function of the  more general \actrect case, as we will show below. Thus, when $m = |X_{new}|$ it would be advisable to use function $f(\cdot)$ of the \actsqr case.

The summery of the above cases can be found in Table \ref{tab:UpdateCasesSum}.

\begin{table}
	\centering
	
	\begin{tabular}{|c|c|c|c|c|}
		\hline
		& & & & \\[-8pt]
		\textbf{Case} & $\mathbf{X_{new}}$ & \textbf{Information} &
		\textbf{Posterior State} & $\mathbf{A}$\textbf{'s Dimension} \tabularnewline
		& & & & \\[-8pt]
		&  & \textbf{Update} &
        \textbf{Dimension} &  \tabularnewline
		\hline
		& & & & \\[-8pt]
		%		\hline
		\Actnaug & empty & $\Lambda_{+} = \Lambda_{-} + A^T \cdot A$ & $n$ & $m \times n$ \tabularnewline
		%		\hline		
		& & & & \\[-8pt]
		\Actrect & not empty & $\Lambda_{+} = \Lambda_{+}^{Aug} + A^T \cdot A$ &
		$N = n + |X_{new}|$ & $m \times N$ \tabularnewline	
		%		\hline		
		& & & & \\[-8pt]
		\Actsqr (subcase of \Actrect) & not empty & $\Lambda_{+} = \Lambda_{+}^{Aug} + A^T \cdot A$ &
		$N = n + |X_{new}|$ & $m \times N$, $m  = |X_{new}|$  \\[-10pt]
		& & & & \\
		\hline				
	\end{tabular}\protect\caption{\label{tab:UpdateCasesSum} Summery of all different variations of change in inference problem. We use $n$ to denote prior state dimension; $N$ - posterior state dimension; $m$ - dimension of all new factors within $F_{new}$.}
\end{table}

Next, 
%in Section \ref{sec:FinalF}
below we  present the function $f(\cdot)$ separately for each one of the \actnaug, \actrect and \actsqr cases. Although the function $f(\cdot)$ has an intricate form (especially in the \actrect case), all matrix terms involved in it have dimensions $m$, $|X_{new}|$ or $|\comb{X}{I}{}{}|$; hence, overall calculation of posterior $\Sigma_{+}^{M,Y}$ does not depend on state dimension $n$.

% In Section \ref{sec:DeriveF} we will show how this function was derived in each one of the cases.

\begin{lemma}\label{lemma:NAugFFuncLemma}For the not-augmented case, the posterior marginal covariance $\Sigma_{+}^{M,Y}$ can be calculated as:
	\begin{equation}
	\Sigma_{+}^{M,Y}
	=
	\Sigma_{-}^{Y} -
	B
	\cdot
	C^{-1}
	\cdot
	B^T
	, \quad
	B
	\triangleq
	\Sigma_{-}^{C}
	\cdot
	(\comb{A}{I}{}{})^T
	, \quad
	C
	\triangleq
	I_m + 
	\comb{A}{I}{}{} \cdot \Sigma_{-}^{I} \cdot (\comb{A}{I}{}{})^T
	\end{equation}
	where $\Sigma_{-}^I$, $\Sigma_{-}^{Y}$ and $\Sigma_{-}^{C}$ are parts of prior marginal covariance $\Sigma_{-}^{M,W}$ partitioned through $W = \{ Y, \comb{X}{I}{}{}\}$:
	\begin{equation}
	\Sigma_{-}^{M,W} =
	\begin{pmatrix} 
	\Sigma_{-}^{Y} & \Sigma_{-}^C\\
	(\Sigma_{-}^C)^T & \Sigma_{-}^I\\
	\end{pmatrix}
	\end{equation}
	and where $\comb{A}{I}{}{}$ consists of $A$'s columns belonging to \involved old variables $\comb{X}{I}{}{}$.
\end{lemma}

The proof of Lemma \ref{lemma:NAugFFuncLemma} is given in Appendix \ref{sec:NAugFFuncLemmaProof}. Note that sets  $Y$ and $\comb{X}{I}{}{}$ are not always disjoint. In case these sets have mutual variables, the cross-covariance matrix $\Sigma_{-}^C$ can be seen just as $\Sigma_{-}^{(Y,\comb{X}{I}{}{})}$ - partition of prior covariance matrix $\Sigma_{-}$ with rows belonging to $Y$ and columns belonging to $\comb{X}{I}{}{}$.

\begin{lemma}\label{lemma:RectFFuncLemma}
	For the rectangular case the prior marginal covariance $\Sigma_{-}^{M,W}$ and the posterior marginal covariance $\Sigma_{+}^{M,Y}$ have the forms:
	\begin{equation}
	\Sigma_{-}^{M,W} =
	\begin{pmatrix} 
	\Sigma_{-}^{Y_{old}} & \Sigma_{-}^C\\
	(\Sigma_{-}^C)^T & \Sigma_{-}^I\\
	\end{pmatrix}
	\end{equation}
	\begin{equation}
	\Sigma_{+}^{M, Y} =
	\begin{pmatrix} 
	\Sigma_{+}^{M, Y_{old}} & \Sigma_{+}^{(Y_{old}, Y_{new})}\\
	(\Sigma_{+}^{(Y_{old}, Y_{new})})^T & \Sigma_{+}^{M, Y_{new}}\\
	\end{pmatrix}
	\label{eq:PostStrRRRR}
	\end{equation}
	where we partition $Y$ variables into two subsets $Y_{old} \doteq X_{-} \cap Y$ and $Y_{new} \doteq X_{new} \cap Y$, and where $W = \{ Y_{old}, \comb{X}{I}{}{}\}$.
	
	Using parts of $\Sigma_{-}^{M,W}$ we can calculate parts of $\Sigma_{+}^{M,Y}$ as:
	\begin{equation}
	\Sigma_{+}^{M, Y_{old}}
	=
	\Sigma_{-}^{Y_{old}} -
	B
	\cdot
	G^{-1} \cdot
	B^T
	\end{equation}
	\begin{equation}
	\Sigma_{+}^{M, Y_{new}} =
	P
	^{(Y_{new}, :)}
	\end{equation}
	\begin{equation}
	C 
	\triangleq
	I_m + 
	\comb{A}{I}{}{} \cdot \Sigma_{-}^{I} \cdot (\comb{A}{I}{}{})^T
	\end{equation}
	\begin{equation}
	P
	\triangleq
	[(A_{new}^T \cdot
	C^{-1}
	\cdot A_{new})^{-1}]
	^{(:, Y_{new})}
	\end{equation}
	\begin{equation}
	F
	\triangleq
	(A_{new}^T \cdot A_{new})^{-1}
	\end{equation}
	\begin{equation}
	K
	\triangleq
	I_m -
	A_{new} \cdot
	F \cdot
	A_{new}^T
	\label{eq:KMatDef}
	\end{equation}
	\begin{equation}
	K_1
	\triangleq
	K \cdot 
	\comb{A}{I}{}{}
	\label{eq:K1MatDefF}
	\end{equation}
	\begin{equation}
	B
	\triangleq
	\Sigma_{-}^C
	\cdot
	K_1^T
	\label{eq:BMatDefF}
	\end{equation}
	\begin{equation}
	G
	\triangleq
	I_m +
	K_1 \cdot \Sigma_{-}^{I} \cdot K_1^T
	\end{equation}
	where
%$\comb{A}{I}{}{}$ consists of $A$'s columns belonging to the \involved variables $\comb{X}{I}{}{}$, and 
$A_{new}$ consists of $A$'s columns belonging to newly introduced variables $X_{new}$. Also, we use matrix slicing operator (e.g. $P^{(Y_{new}, :)}$) as it is accustomed in  Matlab syntax.
	
	Further, there are two methods to calculate $\Sigma_{+}^{(Y_{old}, Y_{new})}$ from Eq.~(\ref{eq:PostStrRRRR}):
	
	Method 1:
	\begin{equation}
	\Sigma_{+}^{(Y_{old}, Y_{new})}
	=
	\Sigma_{-}^C
	\cdot
	(\comb{A}{I}{}{})^T
	\cdot
	[
	C^{-1}
	\cdot
	\comb{A}{I}{}{} \cdot \Sigma_{-}^{I} \cdot (\comb{A}{I}{}{})^T
	- I_m]
	\cdot A_{new}
	\cdot
	P
	\end{equation}

	Method 2:
	\begin{equation}
	\Sigma_{+}^{(Y_{old}, Y_{new})}
	=
	\Sigma_{-}^C
	\cdot
	[
	K_1^T \cdot
	G^{-1} \cdot
	K_1 \cdot \Sigma_{-}^{I}
	- 
	I_k
	]
	\cdot
	(\comb{A}{I}{}{})^T
	\cdot A_{new}
	\cdot
	F^{(:, Y_{new})}
	\label{eq:PosteriorMargCross}
	\end{equation}
\end{lemma}
Empirically we  found that method 2 is the fastest option. The proof of Lemma \ref{lemma:RectFFuncLemma} is given in Appendix \ref{sec:RectFFuncLemmaProof}.

\begin{lemma}\label{lemma:SquareFFuncLemma}
	For the squared case the prior marginal covariance $\Sigma_{-}^{M,W}$ and the posterior marginal covariance $\Sigma_{+}^{M,Y}$ have the forms:
	\begin{equation}
	\Sigma_{-}^{M,W} =
	\begin{pmatrix} 
	\Sigma_{-}^{Y_{old}} & \Sigma_{-}^C\\
	(\Sigma_{-}^C)^T & \Sigma_{-}^I\\
	\end{pmatrix}
	\end{equation}
	\begin{equation}
	\Sigma_{+}^{M, Y} =
	\begin{pmatrix} 
	\Sigma_{+}^{M, Y_{old}} & \Sigma_{+}^{(Y_{old}, Y_{new})}\\
	(\Sigma_{+}^{(Y_{old}, Y_{new})})^T & \Sigma_{+}^{M, Y_{new}}\\
	\end{pmatrix}
	\end{equation}
	where we partition $Y$ variables into two subsets $Y_{old} \doteq X_{-} \cap Y$ and $Y_{new} \doteq X_{new} \cap Y$, and where $W = \{ Y_{old}, \comb{X}{I}{}{}\}$. 
	
	Using parts of $\Sigma_{-}^{M,W}$ we can calculate parts of $\Sigma_{+}^{M,Y}$ as:
	\begin{equation}
	\Sigma_{+}^{M, Y_{old}}
	=
	\Sigma_{-}^{Y_{old}}
	\end{equation}
	\begin{equation}
	\Sigma_{+}^{M, Y_{new}} =
	A_{iv}
	\cdot
	C
	\cdot
	A_{iv}^T
	\end{equation}
	\begin{equation}
	\Sigma_{+}^{(Y_{old}, Y_{new})} =
	- \Sigma_{-}^C
	\cdot
	(\comb{A}{I}{}{})^T
	\cdot
	(A_{iv})^T
	\end{equation}
	\begin{equation}
	A_{iv}
	\triangleq
	[A_{new}^{-1}]^{(Y_{new}, :)}
	\end{equation}
	\begin{equation}
	C 
	\triangleq
	I_m + 
	\comb{A}{I}{}{} \cdot \Sigma_{-}^{I} \cdot (\comb{A}{I}{}{})^{T}.
	\end{equation}
\end{lemma}

We can see that in case of a \actsqr alteration, the covariances of old variables $X_{-}$ do not change. The proof of Lemma \ref{lemma:SquareFFuncLemma} is given in Appendix \ref{sec:SquareFFuncLemmaProof}.

Note that in some applications the inner structure of Jacobian partitions $\comb{A}{I}{}{}$ and $A_{new}$ can be known a priori. In these cases such knowledge can be exploited and the runtime complexity of the above equations can be reduced even more.

\subsubsection{Incremental Covariance Update After Relinearization}
\label{sec:ReleanFFuncPresent}

Till now we have explored scenarios where new information is introduced into our estimation system in a \emph{quadratic} form via Eq.~(\ref{eq:FactorPosteriorInfoMatrixBSDM}). Such information update is appropriate for planning problems where we take linearization point of existing variables $X_{-}$ (their current mean vector) and assume to know linearization point of newly introduced variables $X_{new}$. However, during the inference process itself,  state relinearization can happen and such a \emph{quadratic} update form is not valid anymore. This is because some factors, linearized with the old linearization point, are removed from the system and their relinearized versions are then introduced. In this case the derived approach to incrementally update posterior covariances cannot be used as it is. In this section we  describe the alternative that can be applied after a relinearization event and which is more efficient than state-of-the-art approaches that calculate specific posterior covariances from posterior information matrix from scratch.

Relineariztion may happen when a significantly new information was added into the inference problem and current linearization point of state vector $X_{-}$ does not optimally \emph{explain} it anymore. In such cases, iterative optimization algorithms, such as Gauss-Newton, are responsible to update the current linearization point, i.e.~to find a more optimal linearization point that better \emph{explains} the collected so far measurement/motion/prior factors. Conventional approaches re-linearize the entire state vector when new data comes in. On the other hand, incremental optimizer ISAM2 \citep{Kaess12ijrr} tracks instead the validity of a linearization point of each state variable and re-linearizes only those variables whose change in the linearization point was above a predefined threshold. At each iteration of the nonlinear optimization and for each state variable $x_i$, ISAM2 finds $\delta_i$ and, given it is too big (norm of $\delta_i$ is bigger than the threshold), updates the current estimate of $x_i$ to  $x_{i}^{*} = x_{i}^{*} + \delta_i$. In such case, factors involving this state variable need to be relinearized. Clearly, the frequency of such a relinearization event during the inference process depends on the value of the threshold, and can be especially high during, for example, loop-closures in SLAM scenario. Still, in our simulations we have seen that even with a relatively high threshold and  small number of loop-closures, 
 relinearization of some small state subset $R \subseteq X_{-}$ happens almost every second timestep. Thus, in order to accurately track covariances in the general case, while using conventional approaches that re-linearize each time or ISAM2 which re-linearizes only when it is needed, it is very important to know how to incrementally update covariance entries also after the state was relinearized. Below we  show that information update of such a  relinearization event can be also expressed in a \emph{quadratic} form; thus, the methods from Section \ref{sec:FFuncPresent} that incrementally update specific covariance terms can be applied also here.

Denote by $F_R$ the factors that involve any of the variables in $R$. In order to update information of the estimation after relinearization, we would want to remove $F_R$'s information w.r.t. old linearization point and to add $F_R$'s information w.r.t. the new one.
It is not hard to see that posterior information matrix (after relinearization of subset $R$) can be calculated through
\begin{equation}
\Lambda_{+} = 
\Lambda_{-} - 
A_{-}^T \cdot A_{-}
+
A_{+}^T \cdot A_{+}
\label{eq:RelinearPosteriorInfoMatrixBSDM}
\end{equation}
where matrix $A_{-}$ is a noise-weighted Jacobian of factors $F_R$ w.r.t. old linearization point, and matrix $A_{+}$ is a noise-weighted Jacobian of factors $F_R$ w.r.t. new linearization point.

Next, using complex numbers the above equation becomes
\begin{equation}
\Lambda_{+} = 
\Lambda_{-} +
\begin{pmatrix} 
i A_{-}^T &
A_{+}^T
\end{pmatrix}
\cdot
\begin{pmatrix} 
i A_{-}\\
A_{+}
\end{pmatrix}
=
\Lambda_{-} +
\begin{pmatrix} 
i A_{-} \\
A_{+}
\end{pmatrix}^T
\cdot
\begin{pmatrix} 
i A_{-}\\
A_{+}
\end{pmatrix}
=
\Lambda_{-} +
B^T \cdot B
, \quad
B 
\triangleq
\begin{pmatrix} 
i A_{-}\\
A_{+}
\end{pmatrix}.
\label{eq:RelinearPosteriorInfoMatrixCmplx}
\end{equation}
Note that ${}^T$ operator is transpose and not conjugate transpose. Above we see that also here, the information update is \emph{quadratic} and the \emph{update} matrix $B$ contains terms of old and new Jacobians of factors $F_R$ that were affected by the relinearization event. Therefore, the incremental covariance update described in Section \ref{sec:FFuncPresent}  is also applicable here, making the update of specific covariances much more efficient compared to computation of the covariances from scratch (e.g.~through  Eqs.~(\ref{eq:RecursCovRec1})-(\ref{eq:RecursCovRec2})).

More specifically, the update in Eq.~(\ref{eq:RelinearPosteriorInfoMatrixCmplx}) is an instance of the \actnaug case from Section \ref{sec:FFuncPresent}. By exploiting the specifics of matrix $B$'s structure, Lemma \ref{lemma:NAugFFuncLemma} can be reduced to:
\begin{lemma}\label{lemma:RelinFFuncLemma}For the relinearization case (Eq.~(\ref{eq:RelinearPosteriorInfoMatrixCmplx})), the posterior marginal covariance $\Sigma_{+}^{M,Y}$ can be calculated as:
	\begin{equation}
	\Sigma_{+}^{M,Y}
	=
	\Sigma_{-}^{Y} -
	U
	\cdot
	U^T
	, \quad
	U
	\triangleq
	\Sigma_{-}^{C}
	\cdot
	M
	, \quad
	M \triangleq
\begin{pmatrix} 
i M_1 & M_2\\
\end{pmatrix},
	\end{equation}
\begin{equation}
M_2
\triangleq
(\comb{A}{I}{}{+})^T
\diagup
chol
\Big[
I + 
\comb{A}{I}{}{+} \cdot \Sigma_{-}^{I} \cdot (\comb{A}{I}{}{+})^T
\Big],
\end{equation}
\begin{equation}
M_1
\triangleq
\Big[
(\comb{A}{I}{}{-})^T - M_2 \cdot G
\Big]
\diagup
chol
\Big[
I
-
\comb{A}{I}{}{-} \cdot \Sigma_{-}^{I} \cdot (\comb{A}{I}{}{-})^T
+
G^T \cdot G
\Big],
\end{equation}
\begin{equation}
G
\triangleq
M_2^T
\cdot
\Sigma_{-}^{I} \cdot (\comb{A}{I}{}{-})^T
,
\end{equation}
	where $\Sigma_{-}^I$, $\Sigma_{-}^{Y}$ and $\Sigma_{-}^{C}$ are parts of the prior marginal covariance $\Sigma_{-}^{M,W}$ partitioned through $W = \{ Y, \comb{X}{I}{}{}\}$:
	\begin{equation}
	\Sigma_{-}^{M,W} =
	\begin{pmatrix} 
	\Sigma_{-}^{Y} & \Sigma_{-}^C\\
	(\Sigma_{-}^C)^T & \Sigma_{-}^I\\
	\end{pmatrix}
	\end{equation}
	and where $\comb{A}{I}{}{-}$ consists of $A_{-}$'s columns belonging to the \involved variables $\comb{X}{I}{}{}$; $\comb{A}{I}{}{+}$ contains columns of $A_{+}$ that belong to $\comb{X}{I}{}{}$; $I$ is the identity matrix of an appropriate dimension; $chol(\cdot)$ represents cholesky decomposition which returns an upper triangular matrix; "$\diagup$" is the  backslash operator from Matlab syntax ($A \diagup B = A \cdot B^{-1}$).
\end{lemma}
The proof of Lemma \ref{lemma:RelinFFuncLemma} is given in Appendix \ref{sec:RelinFFuncLemmaProof}. While it is mathematically equivalent to  Lemma \ref{lemma:NAugFFuncLemma}, empirically we found that such a formulation is faster and more numerically stable in case of  relineariztion.

\subsubsection{Incremental Conditional Covariance Update}
\label{sec:CondFFuncPresent}

Above we have seen how to update specific prior marginal covariances given that state's information update has a  \emph{quadratic} form $\Lambda_{+} = \Lambda_{-} + A^T \cdot A$ or $\Lambda_{+} = \Lambda_{+}^{Aug} + A^T \cdot A$. Similarly, we can derive such a method that incrementally updates \emph{specific} conditional covariances since, as we show below, the update of the conditional information matrix from the conditional probability density function has a similar form. 

To prove this statement, let us focus on the \actnaug case where $X_{new}$ is empty. Define a set of variables $Y$, whose posterior conditional covariance $\Sigma_{+}^{Y|F}$, conditioned on an arbitrary disjoint variable set $F$ (with $\{ Y \cup F \} = \varnothing$), needs to be updated. Next, let $U$ be the set of all state variables that are not in $F$, and note that $Y \subseteq U$. The prior information matrix $\Lambda_{-}^{U|V}$ of the prior conditional probability distribution $U|F$ is just a partition of the entire prior information matrix $\Lambda_{-}$ that belongs to columns/rows of variables in $U$. Similarly, the posterior $\Lambda_{+}^{U|V}$ is a partition of $\Lambda_{+}$. It can be easily shown that
\begin{equation}
\Lambda_{+}^{U|V} = 
\Lambda_{-}^{U|V} + (A^U)^T \cdot A^U
\label{eq:FactorPosteriorCondInfoMatrixBSDM}
\end{equation}
where $A^U$ is a partition of noise-weighted Jacobian matrix $A$ that belong to columns of variables in $U$.

 Eq.~(\ref{eq:FactorPosteriorCondInfoMatrixBSDM}) shows that the conditional probability distribution $U|F$ has a \emph{quadratic} update, similar to the marginal probability distribution of the entire state vector $X$. Also, note that the required posterior conditional matrix $\Sigma_{+}^{Y|F}$ is a  partition of the posterior conditional covariance matrix $\Sigma_{+}^{U|V} = (\Lambda_{+}^{U|V})^{-1}$.
For better intuition, it can be seen similar to the  posterior marginal matrix $\Sigma_{+}^{M,Y}$ being a partition of the  posterior marginal covariance matrix $\Sigma_{+} = (\Lambda_{+})^{-1}$ in the \actnaug case (see Lemma \ref{lemma:NAugFFuncLemma}). Thus, there exists a function $f^C(\cdot)$ that calculates $\Sigma_{+}^{Y|F}$ from $\Sigma_{-}^{W|F}$, where $\Sigma_{-}^{W|F}$ is the prior conditional covariance matrix of set $W \doteq \{ Y, \comb{X}{I}{U}{}\}$, conditioned on the set $F$;  here, $\comb{X}{I}{U}{}$ are the  \involved variables that are in $U$. Derivation of such a function $f^C(\cdot)$ is trivial, by following the steps to derive function $f(\cdot)$ in Section \ref{sec:FFuncPresent}, and is left out of this paper in order to not obscure it with additional complex notations.

A similar exposition can be also shown in the augmented case (i.e.~$X_{new}$ is not empty), where information update of conditional distribution also has the augmented \emph{quadratic} form. To summarize, the derived function $f(\cdot)$  in Sections \ref{sec:FFuncPresent} and \ref{sec:ReleanFFuncPresent} can also be used to incrementally update the specific conditional covariances by replacing the prior marginal covariance terms in it with appropriate prior conditional covariances.

\subsubsection{Application of Incremental Covariance Update to SLAM}
\label{sec:SLAMAppCov}

In order to apply our incremental update method in a SLAM setting, we model each change in the inference problem in the form of two separate changes as follows. We consider a specific scenario where at each time step, new robot pose $x_{k}$ ($k$ is index of time step) and new landmarks $L_{new}$ are introduced into the state vector $X$. Further, new factors are introduced into the inference system; these factors include one odometry factor $f_O$ between poses $x_{k-1}$ and $x_{k}$, projection and range factors $F_{new}^{L}$ between the new pose $x_{k}$ and new landmarks $L_{new}$, and finally projection and range factors $F_{old}^{L}$ between $x_{k}$ and old landmarks. Additionally, in general a subset of old factors (denoted by $F_R$) was relinearized as a result of a linearization point change of some old state variables during the inference stage. In case no linearization point change was performed, this set of factors $F_R$ is empty. Note that although we assume above only range and visual measurements, our approach would work for other sensors as well, e.g.~in a purely monocular case.

In the first modeled change, we introduce into the inference system all the new state variables ($x_{k}$ and $L_{new}$) and their constraining factors ($f_O$ and $F_{new}^{L}$), denoted by $X_{new}^{S} = \{ x_{k}, L_{new}\}$ and $F^S_{new} = \{ f_O, F_{new}^{L} \}$, respectively. It can be shown for this change that the dimension of its newly introduced state variables $X_{new}^{S}$ is equal to the dimension of newly introduced factors $F_{new}^{S}$. Thus, such change has a form of the \actsqr  case (see Table \ref{tab:UpdateCasesSum}) and the updated covariance entries due to this change can be calculated by applying Lemma \ref{lemma:SquareFFuncLemma}.  %can be applied to update covariance entries after it. 
Also note that after this change all the new state variables are properly constrained, which is essential for the information matrix to remain invertible. Denote this information matrix, i.e.~after applying the first change, by $\Lambda_{M}$:
\begin{equation}
\Lambda_{M}
=
\Lambda_{k}^{Aug} + A_S^T \cdot A_S
,
\label{eq:FirstStageSqr}
\end{equation}
where $\Lambda_{k}^{Aug}$ is the prior information matrix $\Lambda_{k-1}$ augmented with zero rows/columns for new state variables $X_{new}^{S}$ and $A_S$ is noise-weighted Jacobian of factors $F_{new}^{S}$.

The remaining parts of the original change in the inference problem are represented by the second change. The posterior information matrix can be updated due to this second change as
\begin{equation}
\Lambda_{k}
=
\Lambda_{M} + A_O^T \cdot A_O
-
A_{-}^T \cdot A_{-}
+
A_{+}^T \cdot A_{+}
,
\end{equation}
where $A_O$ is noise-weighted Jacobian of factors $F_{old}^{S}$; $A_{-}$ and $A_{+}$ are noise-weighted Jacobians of factors $F_R$ w.r.t. old and new linearization points, respectively. The above equation can be rewritten as:
\begin{equation}
\Lambda_{k} = 
\Lambda_{M} +
B^T \cdot B
, \quad
B 
\triangleq
\begin{pmatrix} 
i A_{-}\\
A_{+}\\
A_O
\end{pmatrix},
\label{eq:SecondStageRelin}
\end{equation}
and the corresponding covariance matrix can be calculated through Lemma \ref{lemma:RelinFFuncLemma}, or through Lemma \ref{lemma:NAugFFuncLemma} in case there was no relinearization at the current time step, i.e.~$B \equiv A_O$.

To summarize, any change in the inference problem of our SLAM scenario can be represented as a combination of two fundamental changes - \actsqr (Eq.~(\ref{eq:FirstStageSqr})) followed by (relinearized) \actnaug (Eq.~(\ref{eq:SecondStageRelin})); the information matrix is updated as
\begin{equation}
\Lambda_{k-1}
\Longrightarrow
\Lambda_{M}
\Longrightarrow
\Lambda_{k}
\end{equation}
where $M$ can be seen as a logical time step of middle point.

Covariances after the first change can be updated very fast through Lemma \ref{lemma:SquareFFuncLemma}, since as we saw in Section \ref{sec:FFuncPresent}, the marginal covariances of old variables do not change and only marginal covariances of new variables need to be computed in this case. To do this, we  require marginal covariances of \involved variables $\comb{X}{I}{}{}$ from $\Sigma_{k-1}$. Notice that $\comb{X}{I}{}{}$ of the first change contains only $x_{k-1}$, whose marginal covariance is available since it was already calculated in the previous time step. Thus, Lemma \ref{lemma:SquareFFuncLemma} can be easily applied and the marginal covariances of all state variables at middle point $M$ can be efficiently evaluated.

To update all marginal covariances after the second change (through Lemma \ref{lemma:NAugFFuncLemma} or Lemma \ref{lemma:RelinFFuncLemma}) we require marginal covariance of \involved variables $\comb{X}{I}{}{}$ (in factors $F_{old}^{S}$ and $F_R$ of the second change) from covariance matrix $\Sigma_{M} = \Lambda_{M}^{-1}$. Moreover, we will require cross-covariances from $\Sigma_{M}$ between variables $\comb{X}{I}{}{}$ and the rest of the variables, as can be seen from the equations of the lemmas. Thus, we require entire columns from $\Sigma_{M}$ that belong to $\comb{X}{I}{}{}$. These columns can be easily calculated at time $k-1$ (from prior covariance matrix $\Sigma_{k-1}$) and propagated to middle point $M$ by applying Lemma \ref{lemma:SquareFFuncLemma}. The specific columns (belonging to some state subset $Y$) of matrix $\Sigma_{k-1}$ can be efficiently calculated through two backsubstitution operations:
\begin{equation}
V 
\triangleq
R_{k-1}^T \diagdown I_Y
, \quad
\Sigma_{k-1}^{(:,Y)} = R_{k-1} \diagdown V
,
\label{eq:PriorCovColsRetrieve}
\end{equation}
where $I_Y$ are columns from the identity matrix $I$ that belong to variables in $Y$ and "$\diagdown$" is the Matlab's backsubstitution operator with $x = A \diagdown B$ being identical to solving linear equations $Ax = B$ for $x$.

The other alternative for this 2-stage incremental covariance update is to use  Lemma \ref{lemma:RectFFuncLemma} for the \actrect inference change as follows. The posterior information matrix $\Lambda_{k}$ can be calculated in one step as:
\begin{equation}
\Lambda_{k}
=
\Lambda_{k}^{Aug} + A_S^T \cdot A_S + A_O^T \cdot A_O
-
A_{-}^T \cdot A_{-}
+
A_{+}^T \cdot A_{+}
=
\Lambda_{k}^{Aug} +
B^T \cdot B
, \quad
B 
\triangleq
\begin{pmatrix} 
A_S\\
i A_{-}\\
A_{+}\\
A_O
\end{pmatrix}
.
\end{equation}
Such a change has a form of the \actrect  case (see Table \ref{tab:UpdateCasesSum}); therefore, the updated covariance entries and the marginal covariances of new state variables can be calculated by applying Lemma \ref{lemma:RectFFuncLemma}. Note that ${}^T$ operator within the lemma is transpose and not conjugate transpose. Similarly to the above 2-stage method, the \actrect case will also require entire columns from $\Sigma_{k-1}$ that belong to old \involved variables. This can be done here in the same way through Eq.~(\ref{eq:PriorCovColsRetrieve}).

We evaluate the above methods in our SLAM simulation in Section \ref{sec:CovResults} and show their superiority over other state-of-the-art alternatives.

% ==================
\subsection{Information-Theoretic Belief Space Planning}\label{sec:BSPApproachIntro}

In this section we  develop a new approach that, based on the derived above incremental covariance update method, efficiently solves the information-theoretic BSP problem defined in \textbf{Problem 2} from Section \ref{Sec_Notations}. Given a set of candidate actions, the proposed paradigm exploits common aspects
among different actions for efficient BSP in high-dimensional state spaces. Each
(non-myopic) action gives rise to a posterior belief that can be represented by
an appropriate factor graph. In many applications different candidate actions
will share some newly introduced factors and state variables (their factor graph
\increments). For example, two trajectory candidates that partially share their
navigation path, will introduce the same factors for this mutual trajectory part
(see Figure \ref{fig:SLAMExFig-a}). The posterior factor graphs of these
candidate actions therefore have common parts, in terms of factor and variable
nodes, and in addition all of these factor graphs \emph{start} from the belief
at the current time. 

Our proposed paradigm saves computation time by identifying the common parts in
these posterior factor graphs, and switching to a unified graphical model that we introduce, the 
\emph{factor-graph propagation} (FGP) action tree, which represents gradual
construction of posterior factor graphs from the current factor graph. For
instance, in Figures \ref{fig:SLAMExFig-b} and \ref{fig:SLAMExFig-c} two different
FGP action trees are depicted. Both lead to the same posterior beliefs of
candidate actions, yet one of them can be evaluated more efficiently, as will be
explained in Section \ref{sec:ActionTree}. Given such a graphical model,  we develop efficient method to evaluate information impact of each candidate action in unified way. As we show, this method requires specific covariance entries for the
intermediate beliefs that are represented by the tree's vertices, which we calculate by our incremental covariance recovery method (see Section \ref{sec:IncCovUpdate}) with computational complexity that does not depend on
state dimension $n$ (see Section
\ref{sec:IncCovUpdateFGPTree}). Further, we avoid posterior belief propagation and calculation of
determinants of huge matrices for each candidate action by using the
aforementioned incremental covariance update and the  \ramdl method from
\citep{Kopitkov17ijrr}. Moreover, we evaluate candidates' common
parts only once instead of considering these parts separately for each of the
candidates. 

Determining the \emph{best} topology of the FGP action tree, given the
individual factor graphs for different candidate actions, is by itself a
challenge that requires further research. In this paper we consider one specific
realization of this concept, by examining the problem of motion planning under
uncertainty and using the structure of the candidate trajectories for FGP action
tree construction (see Section \ref{sec:ActionTree}). In the results reported in Section \ref{sec:Results} we consider scenario of autonomous exploration in unknown environment where such tree topology allows us to reduce computation time twice compared to baseline approaches.

\subsubsection{\ramdl Approach}\label{sec:RAMDLSurvey}

In our recently-developed approach, \ramdl \citep{Kopitkov17ijrr},
%Our previous technique \ramdl \citep{Kopitkov17ijrr} succeeded to
%evaluate 
the information-theoretic costs (\ref{eq:EntropyPriorDM}) and
(\ref{eq:ObjFuncPosteriorDeterminantFocused}) are evaluated efficiently, without
explicit inference over posterior beliefs for different actions and without calculating
determinants of large matrices. As \ramdl is an essential part of our approach presented herein, below we provide a concise summary for the sake of completeness of the current paper. For a more detailed review of \ramdl the reader is referred to \citep{Kopitkov17ijrr}.

In \citep{Kopitkov17ijrr} we showed that the  information impact of action $a$ (Eqs.~(\ref{eq:EntropyPriorDM}) and
(\ref{eq:ObjFuncPosteriorDeterminantFocused})) is a function of prior covariances for the subset $\comb{X}{I}{}{} \subseteq X_{-}$ that contains variables \involved in new factors $F_{new}$ of $a$, and of matrix $\comb{A}{I}{}{}$ that contains non-zero columns of the
noise-weighted Jacobian matrix $A$. Given the prior covariances of $\comb{X}{I}{}{}$, such a function can be calculated very fast, with complexity independent of state dimension. Thus, in \ramdl we first calculate the required prior covariances for all candidate actions as a one-time, yet still expensive, calculation, after which we efficiently evaluate information impact of each candidate action. The main structure of the \ramdl approach is shown in Algorithm \ref{alg:RAMDLAlgo}.

%\noindent \begin{flushleft}
\begin{algorithm}[t]
	\caption{{\tt rAMDLInformationEvaluation} evaluates information impact of candidates through \ramdl approach and picks the one with the biggest impact.
	} 
	\label{alg:RAMDLAlgo} 
	\SetKwInput{Initialize}{Initialize}
	\SetKwBlock{AlgoBody}{begin:}{end}
	\SetKwInput{inputs}{Inputs}{}
	\SetKwInput{outputs}{Outputs}{}
	\inputs{
		
		$\{\comb{A}{I}{}{i}\}$ : non-zero columns of noise-weighted Jacobians of action candidates $\{a_i\}$
		
		$\{\comb{X}{I}{}{i}\}$ : variables that are \involved in new factors $F_{new}$ of
		each action $a_i$
	}
	\outputs{
		$a^*$ : optimal action }
	\BlankLine
	
	\AlgoBody{
		Calculate prior covariances of variables $X_{All} = \{ \cup \comb{X}{I}{}{i} \}$
		
		\For{$a_i$}{
			Calculate information impact (IG or posterior entropy, \unfocused or \focused), using $\comb{A}{I}{}{i}$ and the required prior covariances calculated in line 6
		}
		
		Select candidate $a^*$ with maximal information impact
		
	}

	\BlankLine
	%\rule{1\columnwidth}{1pt}		
\end{algorithm}
%\rule{1\columnwidth}{1pt}
%	\par\end{flushleft}

In particular, for the case where $X_{new}$ is
empty, the \unfocused IG from Eq.~(\ref{eq:EntropyPriorDM}) can be calculated as
\begin{equation}
J_{IG}(a) = 
\frac{1}{2} \ln \begin{vmatrix} I_m + \comb{A}{I}{}{} \cdot \Sigma_{-}^{M,\comb{X}{I}{}{}}
\cdot (\comb{A}{I}{}{})^T \end{vmatrix},
\label{eq:ObjFuncRAMDL}
\end{equation}
where $\Sigma_{-}^{M,\comb{X}{I}{}{}}$ is the prior marginal covariance of $\comb{X}{I}{}{}$
variables. 

In case $X_{new}$ is empty and we want to calculate \focused IG of \focused variables in $X^F \subseteq X_{-}$ (see Eq.~(\ref{eq:ObjFuncPosteriorDeterminantFocused}), left), it can be calculated through
\begin{equation}
J_{IG}^F(a) = 
\half \ln \begin{vmatrix} I_m + \comb{A}{I}{}{} \cdot \Sigma_{-}^{M, \comb{X}{I}{}{}} \cdot (\comb{A}{I}{}{})^T \end{vmatrix} -
\half \ln \begin{vmatrix} I_m + \comb{A}{I}{U}{} \cdot \Sigma_{-}^{\comb{X}{I}{U}{}|X^F} \cdot (\comb{A}{I}{U}{})^T \end{vmatrix},
\label{eq:FocObjFuncRAMDL}
\end{equation}
where $\comb{X}{I}{U}{} \equiv \comb{X}{I}{}{} \setminus X^F$ denotes the \involved variables that are \unfocused, $\Sigma_{-}^{\comb{X}{I}{U}{}|X^F}$ is the prior conditional covariance of $\comb{X}{I}{U}{}$ conditioned on $X^F$, and $\comb{A}{I}{U}{}$ is a partition of $\comb{A}{I}{}{}$ with columns that belong to variables in $\comb{X}{I}{U}{}$.

In order to efficiently evaluate all
candidates in the \unfocused case, \ramdl first calculates the prior marginal covariance
$\Sigma_{-}^{M,X_{All}}$ of variables $X_{All} \subseteq X_{-}$, where $X_{All}$
is the union of \involved variables $\comb{X}{I}{}{}$ of all candidate actions. Further,
evaluation of IG for each action is done by first retrieving
$\Sigma_{-}^{M,\comb{X}{I}{}{}}$ from $\Sigma_{-}^{M,X_{All}}$ and then  calculating $J_{IG}(a)$ via 
Eq.~(\ref{eq:ObjFuncRAMDL}). Overall, such a  process consists of only
one-time calculation that depends on state dimension $n$, i.e.~calculation of
$\Sigma_{-}^{M,X_{All}}$. Other cases of interest (where $X_{new}$ is non-empty or for
\focused BSP objective functions) are also addressed by \ramdl. Note that in case of \focused BSP the prior conditional covariances $\Sigma_{-}^{\comb{X}{I}{U}{}|X^F}$ are additionally required (see Eq.~(\ref{eq:FocObjFuncRAMDL})) and can also be calculated for all candidates in one-block computation.

%For other cases ($X_{new}$ is non-empty or for \focused BSP objective
%functions) please see \citep{Kopitkov17ijrr}.

Yet, the \ramdl method does not fully exploit similarities between  candidate
actions. The mutual \increment part of the actions is expressed as identical block-rows in the matrix $\comb{A}{I}{}{}$ of these actions and thus is evaluated multiple times.
In the next section we present a novel approach to perform planning under
uncertainty where mutual parts of the actions can be evaluated only once,
further decreasing the CPU demand of the overall planning task.

% ============
\subsubsection{Factor-graph Propagation (FGP) Action Tree}\label{sec:ActionTree}

%In [cite previous work] we have seen that the \unfocused IG and \focused
%entropy of newly introduced variables can be calculated efficiently by using
%different partitions of noise-weighted Jacobian matrix $A$ and prior marginal
%covariance $\Sigma_{-}^{M,I}$ of variables $X_{-}^{I}$ that were \involved in
%any of new factors $F_{new}$. For \focused IG of old \focused variables
%$X^F_{-}$ we would also require prior conditional covariance entries
%$\Sigma_{-}^{IU|F}$ of \unfocused \involved variables $X_{-}^{IU} \subseteq
%X_{-}^{I}$ conditioned on $X^F_{-}$. In our \ramdl method [cite] first we are
%calculating prior covariances required by any of the candidates as one-time
%computation. Further we share these prior covariance entries to evaluate each
%candidate action in very fast way.
%
%Concluding, it is important to understand that given covariances
%$\Sigma_{-}^{M,I}$ (and $\Sigma_{-}^{IU|F}$ if needed) of prior factor graph
%$G_{-}$ and given the Jacobian $A$ representing the candidate's \increment of
%$G_{-}$, we can efficiently calculate information impact of the action.

The FGP action tree describes the concept of belief propagation through a factor
graph representation. Each vertex in this tree (see Figures
\ref{fig:SLAMExFig-b} and \ref{fig:SLAMExFig-c}) encodes a factor graph that
represents a specific belief. For example, the root represents the prior belief
$b[X_{-}]$ and leafs represent the posterior factor graphs of different
candidate actions. Each edge $e_{v \rightarrow u}$, between vertices $v$ and
$u$, represents an action $a$ with an appropriate \increment $I(a)$, see
Eq.~(\ref{eq:IncrementDef}). Thus, the factor graph encoded by vertex $u$ is
obtained by applying the increment $I(a)$ to the factor graph that is encoded by
vertex $v$. %by which the factor graph encoded by vertex $v$ is augmented in
%order to receive factor graph $u$. 
Below we will show how such a graphical model can be used to efficiently reason,
while exploiting common parts,  about posterior beliefs of different actions.

Let us  consider a simple case as a running example, where two candidate actions
$a_1$ and $a_2$  share some of their \increments (see Figure
\ref{fig:SLAMExFig-a}). As can be seen both trajectories have a mutual part
which is colored in green. One way to evaluate the action impact for actions
$a_1$ and $a_2$ is to handle each case separately (see Figure
\ref{fig:SLAMExFig-b}). Indeed, existing approaches typically perform inference
over the posterior belief for each of the actions and then evaluate the
information-theoretic cost. However, this can be done by far more efficiently
using our recently developed \ramdl  approach \citep{Kopitkov17ijrr}, where first we perform a one-time calculation of \emph{specific}
prior covariance entries that are required by both candidates, followed by
information impact evaluation of each candidate (see Section
\ref{sec:RAMDLSurvey}). 
%them is by handling them separately (see Figure \ref{fig:SLAMExFig-b}) and
%reason about the posterior information of each action in independent way. It
%can be done efficiently by using our previous approach \ramdl
%\citep{Kopitkov17ijrr} where first we perform one-time
%calculation of \emph{specific} prior covariance entries that are required by
%both candidates, following by information impact evaluation of each candidate. 
While the one-time covariance computation depends on state dimension $n$, the
candidates evaluation does not. Yet, in such an approach, although we
significantly reduce run-time by gathering the expensive computation of prior
covariances from all candidates into a single computational block, we still
waste computational resources related to the mutual \increment, which is
calculated separately for each candidate action (e.g.~twice in the considered
example).

%would be calculated twice (in the considered example), i.e.~for each candidate
%action separately.

%we will still waste  computation resources in order to consider the mutual
%\increment part twice, for each candidate separately.

In this paper we propose another alternative. Referring to the running example,
we split each of the two actions into $a_1 = \{ a_{shr}, a_1' \}$ and $a_2 = \{
a_{shr}, a_2' \}$ and present them through a  multiple-layered FGP tree (see Figure
\ref{fig:SLAMExFig-c}), where $a_{shr}$ represents the shared part of actions'
\increments, and where $a_1'$ and $a_2'$  represent parts of the original
actions that are not shared. It is not difficult to show that IG of each
candidate $a_i$ is equal to sum of IG's of its sub-actions $a_{shr}$ and $a_i'$,
i.e.~$IG(a_i) = IG(a_{shr}) + IG(a_i')$ (see proof in Appendix \ref{sec:IGSumProof}). Thus, in this specific example in order to
select the best action it is enough to calculate IG of $a_1'$ and $a_2'$. This
IG can be efficiently calculated through the \ramdl technique, but this time we
will require \emph{specific} covariance entries of the intermediate belief
associated with $G_{+}^{a_{shr}}$ (factor graph obtained after execution of
action $a_{shr}$, see Figure
\ref{fig:SLAMExFig-c}).
For example, \unfocused IG of $a_1'$ (see Eq.~(\ref{eq:ObjFuncRAMDL})) can be calculated as
\begin{equation}
J_{IG}(a_1') = 
\frac{1}{2} \ln \begin{vmatrix} I_m + \comb{A}{I}{}{1} \cdot \Sigma_{shr}^{M,\comb{X}{I}{}{1}}
\cdot (\comb{A}{I}{}{1})^T \end{vmatrix},
\label{eq:ObjFuncRAMDLExample}
\end{equation}
where $\comb{X}{I}{}{1}$ are variables \involved in new factors of $a_1'$, $\comb{A}{I}{}{1}$ are non-zero columns from noise-weighted Jacobian of these new factors and $\Sigma_{shr}^{M,\comb{X}{I}{}{1}}$ is the marginal covariance of $\comb{X}{I}{}{1}$ from intermediate belief represented by $G_{+}^{a_{shr}}$. Assuming there is an efficient way to calculate
\emph{specific} covariance entries for each vertex within FGP tree (see Section
\ref{sec:IncCovUpdateFGPTree}), we can apply the \ramdl method, calculate the
required information impacts and make decision between $a_1$ and $a_2$ while
handling mutual \increment $a_{shr}$ only once, and not twice as would be done
by existing approaches.

%We can compute the information gain of action $a_{shr}$ that was received after
%propagating from $G_{-}$ to $G_{+}^{a_{shr}}$, by again applying the \ramdl
%technique. 
%
%However, in order to decide which of the two trajectories have bigger
%information impact, we need also to evaluate information impact of both $a_1'$
%and $a_2'$. And in order to do that by using \ramdl method, we will require
%specific covariance entries from the middle belief represented by
%$G_{+}^{a_{shr}}$. Assuming that we can calculate these entries in efficient
%way (see section Incremental Covariance Update), we can apply \ramdl method and
%calculate the information impact of each edge of the \emph{factor-graph} action
%tree (see Figure \ref{fig:TreeExample}, right) representing specific action
%with specific \increment.

%
\begin{figure}[!t]
	\centering
	
	\subfloat{\includegraphics[width=0.8\textwidth]{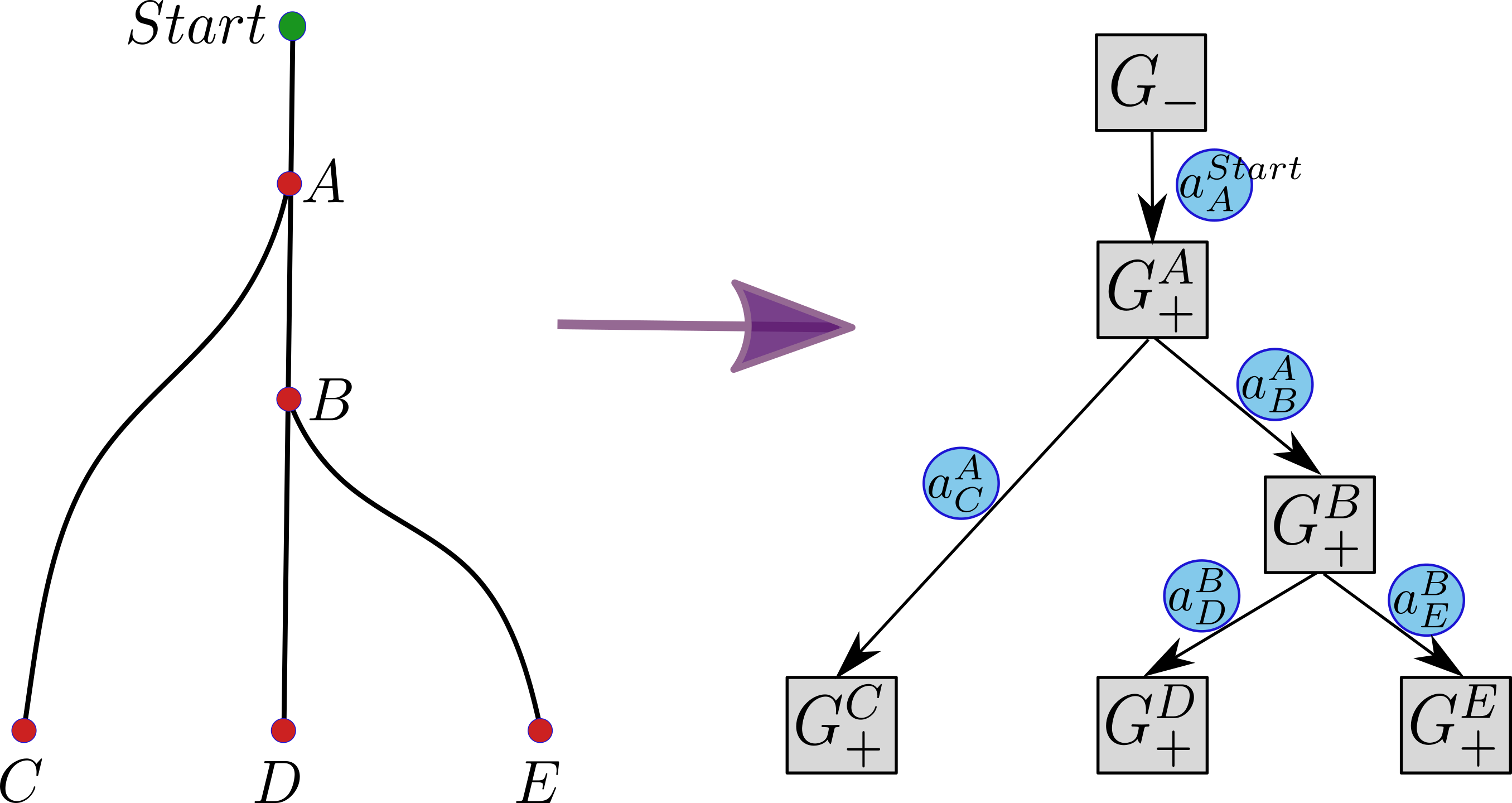}}
	\protect
	\caption{Different candidate trajectories and their FGP action tree
		representation. $A$ and $B$ are splitting waypoints; $C$, $D$ and $E$ are final
		waypoints of 3 trajectory candidates. Each waypoint has associated factor graph
		vertex within the tree. Each action $a_{y}^{x}$ is augmenting factor graph with
		factors/variables gathered by passing path $x \rightarrow y$.
	}
	\label{fig:TreeBiggerExample}
\end{figure}

The above concept applies also to more general problem settings, with numerous
candidate actions with mutual parts in their \increments. %In similar way, we
%can extend this idea to any number of candidate actions where actions'
%\increments have many mutual parts. 
An excellent example for this is belief space planning for autonomous
navigation. Here, the set of trajectory candidates can be naturally represented
as tree of possible paths, and the FGP action tree can be constructed in such a
way that each of its intermediate vertices will represent a belief at a specific
splitting waypoint of the trajectories (see Figure \ref{fig:TreeBiggerExample}).
In such a general case, in order to pick up the optimal action we will need to
calculate IG for each one of the tree's edges. This can be done again by
applying \ramdl technique but will require us to know the \emph{specific}
covariance entries for each intermediate vertex within FGP tree. An efficient
calculation of these entries is presented in Section \ref{sec:IncCovUpdateFGPTree},
while the overall algorithm to evaluate the FGP tree is summarized in Algorithm
\ref{alg:FGPTreeEval}.

%	%
%	\begin{algorithm}
%		\caption{Tree Evaluation}\label{euclid}
%		\begin{algorithmic}[1]
%			\Procedure{EvaluateFGPTree}{}
%			\State $\textit{T} \gets \text{FGP tree}$
%			\BState \emph{begin}:
%			\State For each vertex  $\textit{v}$ of $\textit{T}$, determine the set of
%			variables  $Y$ whose covariances are required by \ramdl
%			\State Calculate these covariances (see Section \ref{sec:IncCovUpdate})
%			\State Calculate IG of tree's each edge through \ramdl
%			\State Calculate IG of each candidate by summarizing IG along candidate's
%			trajectory
%			\State Select candidate with maximal IG
%			\BState \emph{end}
%			\EndProcedure
%		\end{algorithmic}
%		\label{alg:FGPTreeEval}
%	\end{algorithm}
%	%

%\noindent \begin{flushleft}
\begin{algorithm}[t]
	\caption{{\tt EvaluateFGPTree} evaluates information impact of candidates and picks the one with the biggest impact.
	} 
	\label{alg:FGPTreeEval} 
	\SetKwInput{Initialize}{Initialize}
	\SetKwBlock{AlgoBody}{begin:}{end}
	\SetKwInput{inputs}{Inputs}{}
	\SetKwInput{outputs}{Outputs}{}
	\inputs{
		$T$ : FGP tree
	}
	\outputs{
		$a^*$ : optimal action }
	\BlankLine
	
	\AlgoBody{

	    \For{$\textit{v}$: vertex of $\textit{T}$}{
	Determine the set of
	variables  $Y$ whose covariances are required by \ramdl in order to calculate IG of action between $v$'s parent and $v$ (variables \involved in factors that were introduced by augmenting $v$'s parent to acquire $v$)
}

		Calculate these covariances (see Section \ref{sec:IncCovUpdate})

		Calculate IG of tree's each edge through \ramdl (e.g. through Eq.~(\ref{eq:ObjFuncRAMDL}))
		
		Calculate IG of each candidate by summarizing IG along candidate's trajectory
		
		Select candidate $a^*$ with maximal IG
		
	}

	\BlankLine
	%\rule{1\columnwidth}{1pt}		
\end{algorithm}
%\rule{1\columnwidth}{1pt}
%	\par\end{flushleft}

Note that although in this paper we create an FGP action tree with a structure
similar to the tree of candidate navigation paths, in general, different structures can be
used. For example, if candidates share their trajectories' terminal part, this
part can be represented as first action under root $G_{-}$. As long as tree's
root represents the prior belief $b[X_{-}]$ and the tree has a vertex for
posterior belief of each candidate action, it represents the same decision
problem. An interesting question that arises is how to find the tree's structure
that provides the biggest calculation re-use between the candidates and can be
evaluated most efficiently. We will leave this question for future research.

Also note that the proposed method can be also applied to the scenario where a similar candidate trajectory is evaluated at sequential time steps. Such a candidate trajectory, taking the robot to some location, at each time step may have a different starting section due to robot's movement since the previous time step, but will have the same terminal section that brings the robot to the aforementioned location (see also \citep{Chaves16iros}). Thus, this candidate trajectory will have similar posterior factor graphs each time it is evaluated. This similarity between posterior factor graphs can be naturally represented through our FGP tree and hereof it is just another application for our BSP approach.

% ==================
\subsubsection{Incremental Covariance Update within FGP Action Tree}\label{sec:IncCovUpdateFGPTree}

\begin{figure}[!t]
	\centering

	\begin{tabular}{cccc}
		
		\subfloat{\includegraphics[width=0.45\textwidth]{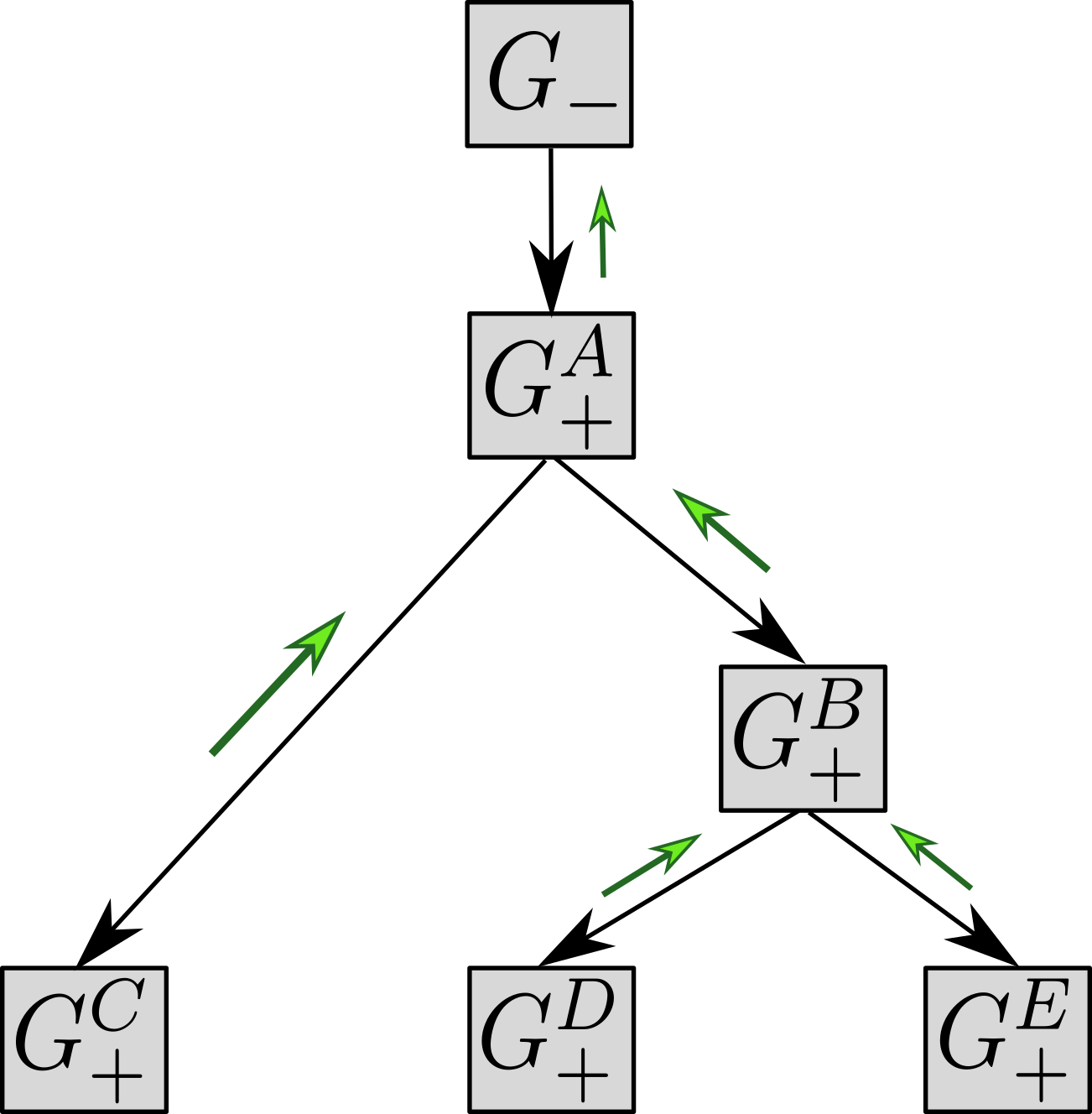}}
		& 
		\hspace{15pt}
		&
		
		\subfloat{\includegraphics[width=0.45\textwidth]{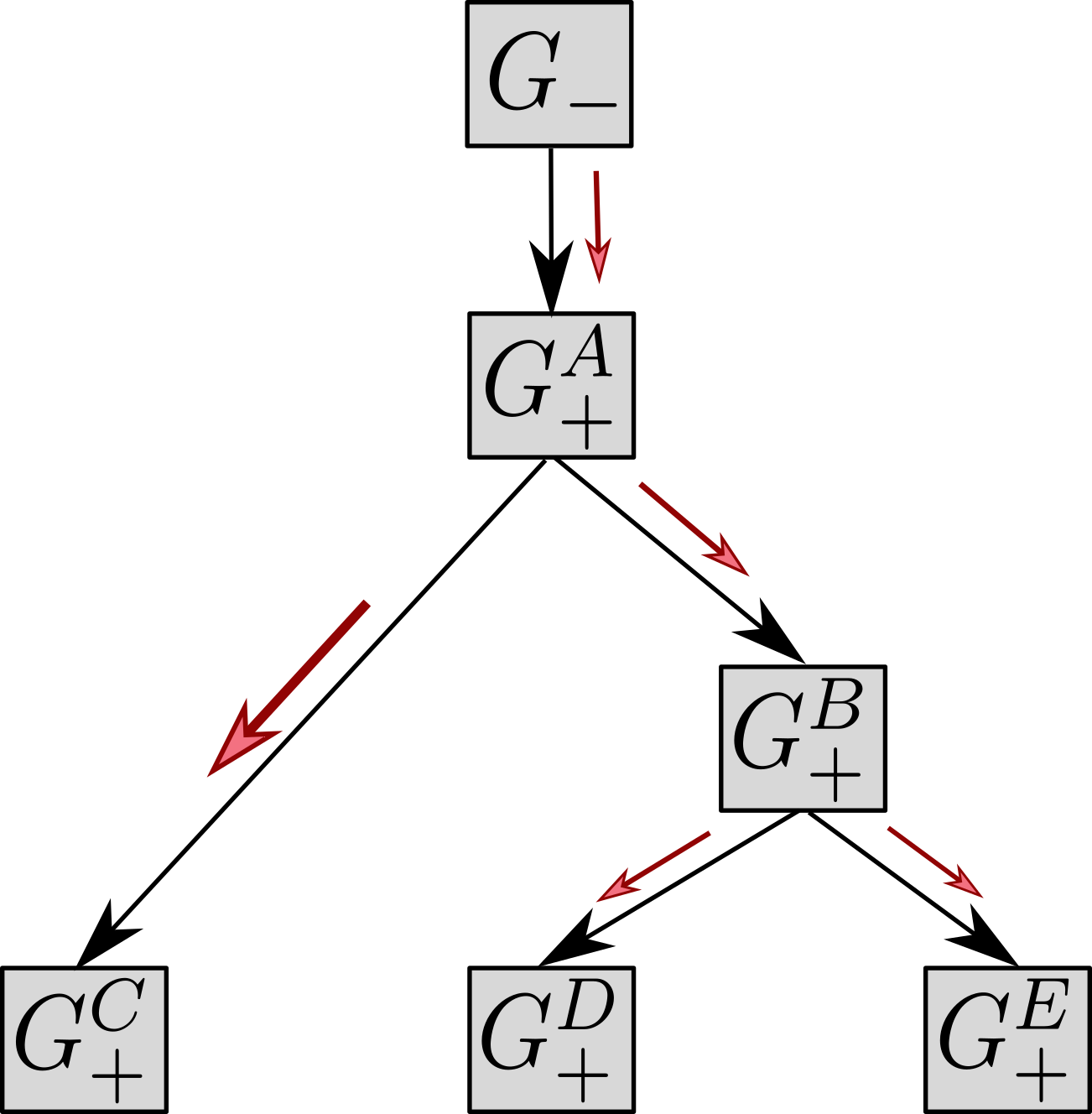}}
		
		\\
	\end{tabular}
	
	\protect
	\caption{Incremental covariance update within FGP action tree, illustrated on
		tree from Figure \ref{fig:TreeBiggerExample}. Each vertex in tree represents
		specific factor graph and state belief associated with it. Each edge in tree
		(black arrows) represents action that augments parent's factor graph in order to
		obtain the child's factor graph.
		The covariance update consists of two stages. 
		First (left drawing): from bottom to top each vertex notifies (green arrows)
		its parent what are the marginal covariance entries that it needs from parent's
		belief.
		Second (right drawing): from top to bottom each vertex calculates the required
		marginal covariance entries of its belief and notifies its children (red arrows)
		to proceed with their covariance calculations. 
		This covariance update process contains one-time calculation depending on state
		dimension $n$ - computation of required marginal covariance entries at root
		$G_{-}$.
		Rest of the calculations is incremental and does not depends on $n$, as
		described in Section \ref{sec:IncCovUpdate}. 
	}
	\label{fig:TreeIncCovUpdate}
\end{figure}
%

%\noindent \begin{flushleft}
\begin{algorithm}[t]
	\caption{{\tt CalculateCovariances} incrementally calculates specific covariances in the beliefs represented by vetrteces of FGP Action Tree.
	} 
	\label{alg:IncremenCovCalc}
	\SetKwInput{Initialize}{Initialize}
	\SetKwBlock{AlgoBody}{begin:}{end}
	\SetKwInput{inputs}{Inputs}{}
	\SetKwInput{outputs}{Outputs}{}
	\inputs{
		
		$T$ : FGP tree
		
		$\{Y_{u}\}$ : set of variables whose covariances from vertex $u$'s belief we are interested in, for each vertex $u$ in $T$
	}
	\outputs{
		$\{\Sigma_{u}^{M,Y_{u}}\}$ : the calculated covariances}
	\BlankLine
	
	\AlgoBody{
		
		\For{$\textit{u}$: vertex of $\textit{T}$, in bottom-top ordering }{
			Message $u$'s parent $v$ that we are interested in covariances in $v$'s belief for variables $\{ Y_{u}, \comb{X}{I}{}{}, Y_{ch}\}$, where $Y_{u}$ are variables required by the main algorithm \ref{alg:FGPTreeEval}, $\comb{X}{I}{}{}$ are variables \involved in factors that were introduced by augmenting $v$ to acquire $u$, and $Y_{ch}$ are variables that were required by children of $u$
		}
		
		Define set of variables $Y_{T} \doteq \{ Y_{u}, Y_{ch}\}$ for each vertex $u$ 
		
		Calculate marginal covariances of $Y_{T}$ at $T$'s root $G_{-}$, $\Sigma_{-}^{M,Y_{T}}$ from prior belief $b[X_{-}]$ (e.g. through Schur Comlement)

		\For{$\textit{u}$: vertex of $\textit{T}$, in top-bottom ordering }{
			Calculate $\Sigma_{u}^{M,Y_{T}}$ through function $f(\cdot)$ (see Section \ref{sec:FFuncPresent}), by using required covariances $\Sigma_{v}^{M,W}$ from $u$'s parent
			$v$
			
			Retrieve the required $\Sigma_{u}^{M,Y_{u}}$ from calculated $\Sigma_{u}^{M,Y_{T}}$
		}

	}

	\BlankLine
	%\rule{1\columnwidth}{1pt}		
\end{algorithm}
%\rule{1\columnwidth}{1pt}
%	\par\end{flushleft}

In order to reason about different actions inside an FGP action tree, we have to
know \emph{specific} covariance entries for each intermediate vertex in the
tree. We can calculate these entries by first propagating the beliefs through
Eq.~(\ref{eq:FactorPosteriorInfoMatrixBSDM}) followed by appropriate Schur
complement and inverse operations. However, such a procedure will depend on a
potentially huge state dimension $n$, which we would like to avoid. Here we
propose an alternative method to calculate specific covariance entries at each
one of the beliefs inside the tree which is based on our incremental covariance update technique (see Section \ref{sec:IncCovUpdate}) and does \emph{not} depend on $n$. Moreover, the proposed method can be applied to calculate both specific marginal and conditional covariance entries, where the former are required for the \unfocused information objective function (see Eq.~(\ref{eq:ObjFuncRAMDL})) and the latter are required for the \focused information objective function (see Eq.~(\ref{eq:FocObjFuncRAMDL})).

%The main idea of our approach for incremental covariance update is similar in spirit to work \citep{Ila15icra} but is more general and sophisticated. While in \citep{Ila15icra} authors handle only a simple covariance update in case where action is not augmenting state vector ($X_{new}$ of action is empty), we solve also the augmented case (action is introducing new state variables $X_{new}$). Moreover, we show additionally how to apply a similar approach in order to incrementally update conditional covariances which are of interest for \focused information objective functions (see Eq.~(\ref{eq:FocObjFuncRAMDL})).

First of all, let us focus on a specific edge $e_{v \rightarrow u}$ in the FGP tree
that represents some action $a$ with \increment $I(a) = \{ F_{new}, X_{new} \}$.
In other words, the factor graph represented by $v$ is augmented by $I(a)$ in
order to receive the factor graph that is represented by $u$. Also, let us
denote state vectors of beliefs of $v$ and of $u$ by $X_{u}$ and $X_{v}$,
respectively. Note that $X_{v} \subseteq X_{u}$, and that $X_{u}$ will sometime contain state variables which are not present yet in $X_{v}$ (the variable set $X_{new}$).

Now, consider the set of variables $Y \subseteq X_{u}$ whose marginal covariance
$\Sigma_{u}^{M,Y}$ from $u$'s belief we would like to calculate. As was shown in Section \ref{sec:IncCovUpdate},  $\Sigma_{u}^{M,Y}$ can be calculated efficiently and independently of state dimension $n$,
given that we have the marginal covariance of the set $W \doteq \{ Y_{old}, \comb{X}{I}{}{}\}$ from
$v$'s belief, where $\comb{X}{I}{}{} \subseteq X_{v}$ is the set of \involved variables in
action $a$ and $Y_{old}$ is the intersection between $Y$ and $X_{v}$. It is important to note that, in a general case, the marginal
covariance of $Y_{old}$ is modified after applying some action $a$,
i.e.~$\Sigma_{u}^{M,Y_{old}} \neq \Sigma_{v}^{M,Y_{old}}$. Similarly to Section \ref{sec:IncCovUpdate} we can separate all possible
actions in the FGP tree into different categories depending on their \increments, i.e.~\actnaug, \actrect and \actsqr. Consequently, for each action type we can use an appropriate covariance update method in order to calculate $\Sigma_{u}^{M,Y} = f(\Sigma_{v}^{M,W})$.

%%In order to simplify this paper and focus on the theoretical concept instead of math derivations, 
%For simplicity, we present the way to calculate $\Sigma_{u}^{M,Y}$ from $\Sigma_{v}^{M,W}$ in Section \ref{sec:FFuncPresent}. There, we separate all possible
%actions in the FGP tree into different categories depending on their \increments
%and derive the required function $\Sigma_{u}^{M,Y} = f(\Sigma_{v}^{M,W})$ for
%all different cases. But first, in Section \ref{sec:FGPCovMethod} we will show how can we use this derived function $f(\cdot)$ in order to incrementally calculate the required \emph{specific} covariances for
%each one of the vertices in the FGP action tree.

Next, we can use the mentioned above function $\Sigma_{u}^{M,Y} = f(\Sigma_{v}^{M,W})$ to calculate the required \emph{specific} covariances for
each one of the vertices in the tree recursively (see also Figure
\ref{fig:TreeIncCovUpdate}): First, for each vertex $u$ we define by $Y$ the variables of interest whose marginal covariances $\Sigma_{u}^{M,Y}$ at the belief associated with $u$ we would like to calculate. In our case $Y$ are the variables required by \ramdl in order to evaluate
impact of actions that are performed on $u$ (see Section \ref{sec:ActionTree}).
Next, for each leaf vertex $u$ we message its parent $v$ that we require $v$'s
marginal covariances for $\{ Y, \comb{X}{I}{}{}\}$. Then in recursive form from bottom to
top each vertex $v$ will message its parent that it requires its parent's
covariances for $\{ Y, \comb{X}{I}{}{}, Y_{ch}\}$ where $Y_{ch}$ is the set of variables that
were required by $v$'s children. Eventually, for each vertex $v$ in the tree we
will have a total set of variables $Y_{T} \doteq \{ Y, Y_{ch}\}$ whose
covariances we need to compute for this specific vertex. 

Finally, we start to propagate these covariances in top to bottom order. Using
the equations from Section \ref{sec:FFuncPresent}, for each vertex $u$ we
can calculate $\Sigma_{u}^{M,Y_{T}}$ using  $\Sigma_{v}^{M,W}$ from its parent
vertex $v$. Note that when following top to bottom order, when we get to node
$u$, its parent's covariances $\Sigma_{v}^{M,W}$ will be already computed. Also
note that the required prior covariance entries of the root $G_{-}$ should be
calculated first. This is done only once and its complexity depends on state
dimension $n$, similarly to \ramdl technique. But once calculated, the rest of
the covariance updates do not depend on $n$. 

%\VI{[Not related to this paper - an interesting question is how to exploit FGP in a consecutive  planning session (MPC). Could collaborate with Elad on this]} \DK{[I'm not familiar with what exactly Elad does but I thought his work is bayes-tree dependend. Here FGP tree is working differently from bayes tree, I think. But he can maybe join them.. I need to spend more time on my PhD.]}

To summarize, the described algorithm consists of two parts - detecting
variables set $Y_{T}$ for each vertex and propagating \emph{specific}
covariances from top to bottom. See  Algorithm \ref{alg:IncremenCovCalc} and a schematic illustration of the incremental
covariance update in  Figure \ref{fig:TreeIncCovUpdate}. The runtime complexity
of the algorithm  mainly depends on its second part, since variable detection
does not require any matrix manipulations and can be done  fast. The second part
handles each edge of the FGP tree only once, thus again allowing us to evaluate
mutual \increment of actions only once. Run-time to propagate \emph{specific}
covariances along each edge depends on a number of parameters such as dimension
of required covariances and size of action's \increment.

In a similar way we can also incrementally propagate \emph{specific} conditional
covariances along the FGP tree. Such covariances may also be required in order
to perform informative-theoretic decision making when we want to reduce
uncertainty of a subset of old variables $X^F \subseteq X_{-}$, see
\citep{Kopitkov17ijrr}.

% ===============
\section{Results}
\label{sec:Results}

We evaluate the proposed approaches for incremental covariance update and BSP in simulation
considering the problem of autonomous navigation in unknown environments. The
robot has to autonomously visit a set of predefined goals while localizing
itself and mapping the environment using its onboard sensors. In our simulation,
we currently consider a monocular camera and a range sensor. 
%In simulated environment robot is visiting set of predefined goals while also
%performing SLAM. 
The code is implemented in Matlab and uses the GTSAM library\citep{Dellaert12tr,
	Kaess12ijrr}. All scenarios were executed on a Linux machine with i7 2.40 GHz
processor and 32 Gb of memory. All compared approaches were implemented in single thread to provide better visualization of their runtime complexity.
Additionally, we provide our implementation of FGP action
tree as open-source library in 
"\url{http://goo.gl/dmNenc}".

% ===============
\subsection{Covariance Recovery}
\label{sec:CovResults}

\begin{figure}[!t]
	\centering
	
	\begin{tabular}{cccc}
		\subfloat[\label{fig:CovFigA-a}]{\includegraphics[width=0.48\textwidth]{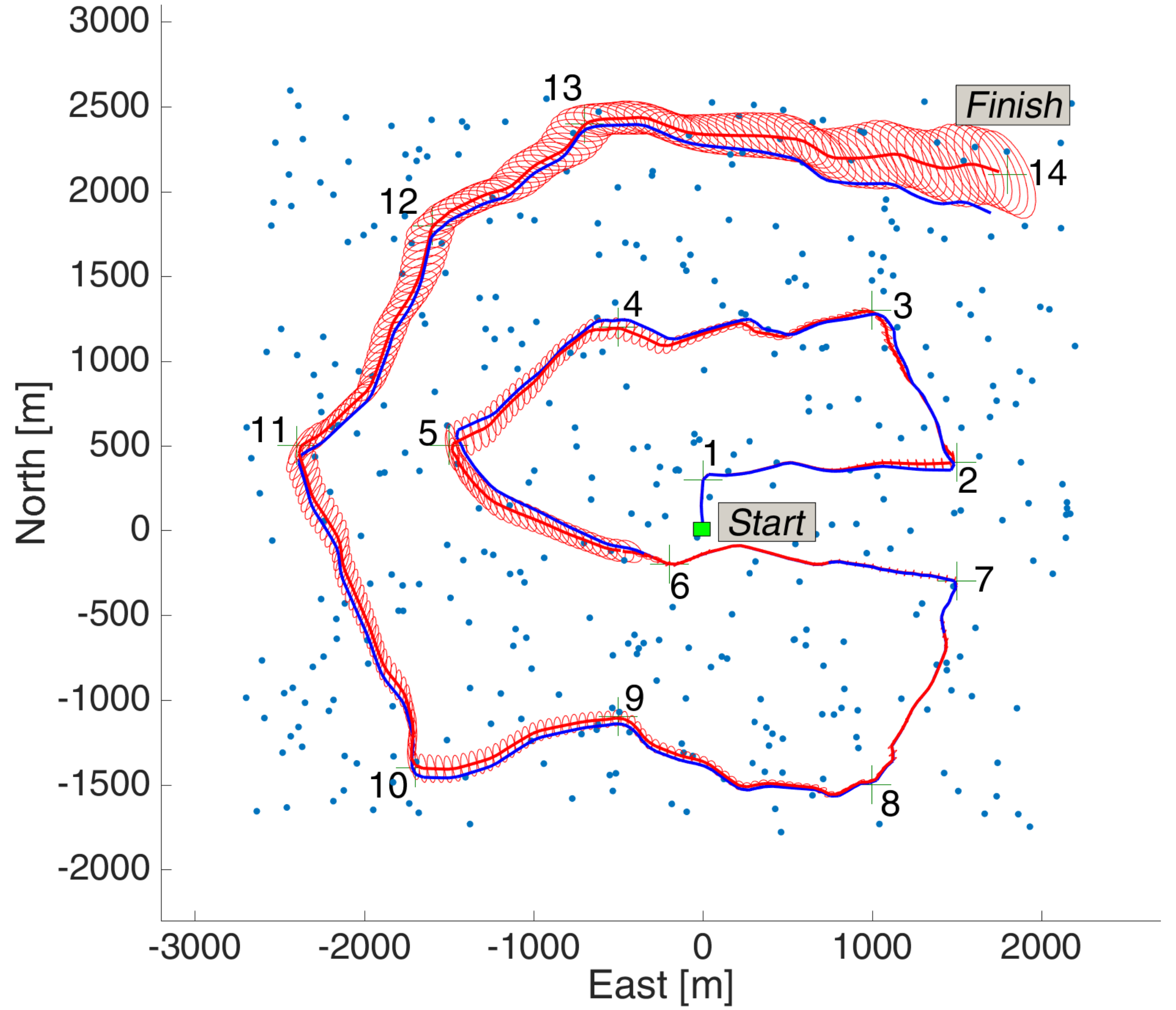}}
		&
		\subfloat[\label{fig:CovFigA-b}]{\includegraphics[width=0.48\textwidth]{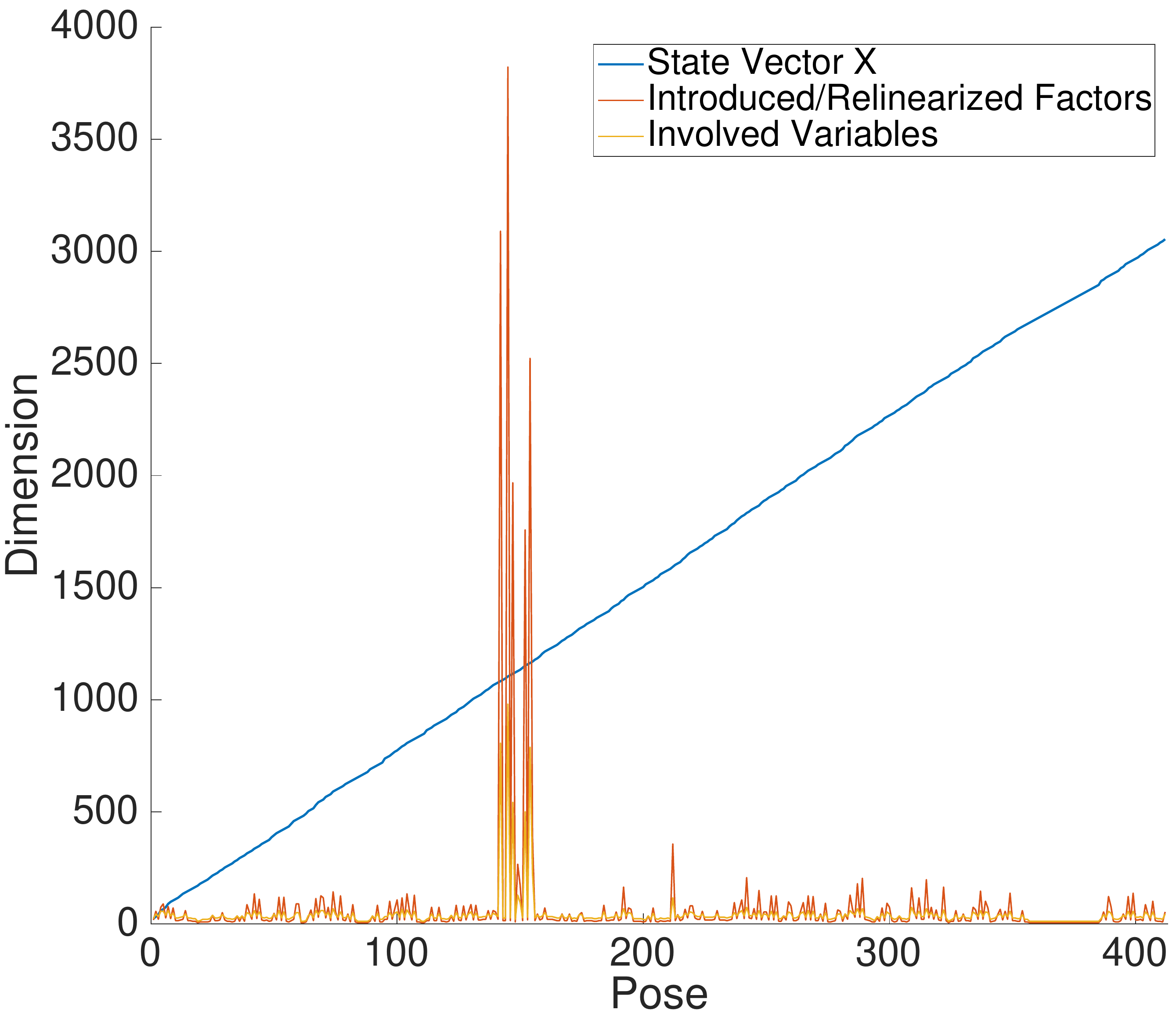}}
		\\
	\end{tabular}

	\begin{tabular}{cccc}
	\subfloat[\label{fig:CovFigA-c}]{\includegraphics[width=0.48\textwidth]{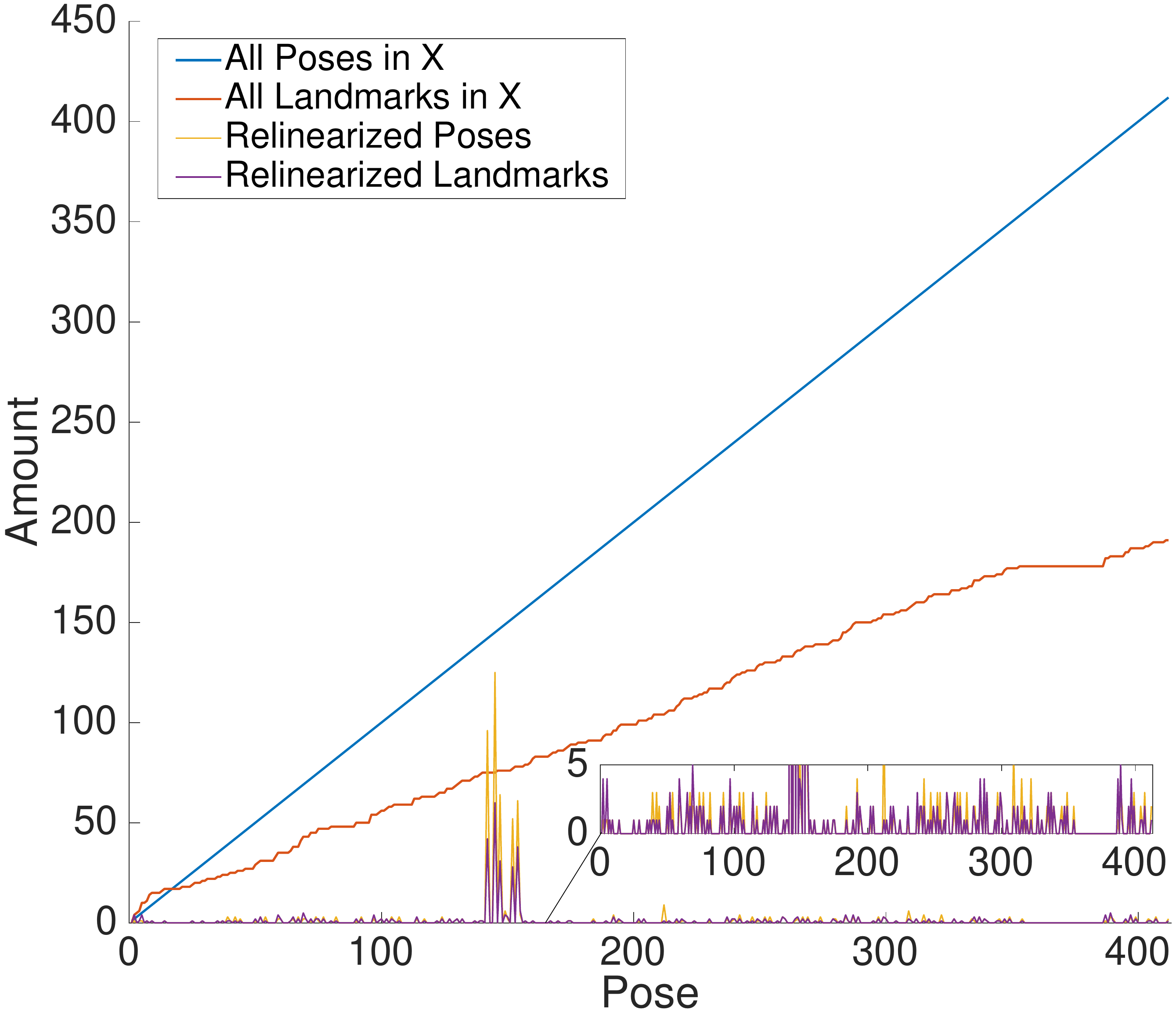}}
	&
	\subfloat[\label{fig:CovFigA-d}]{\includegraphics[width=0.48\textwidth]{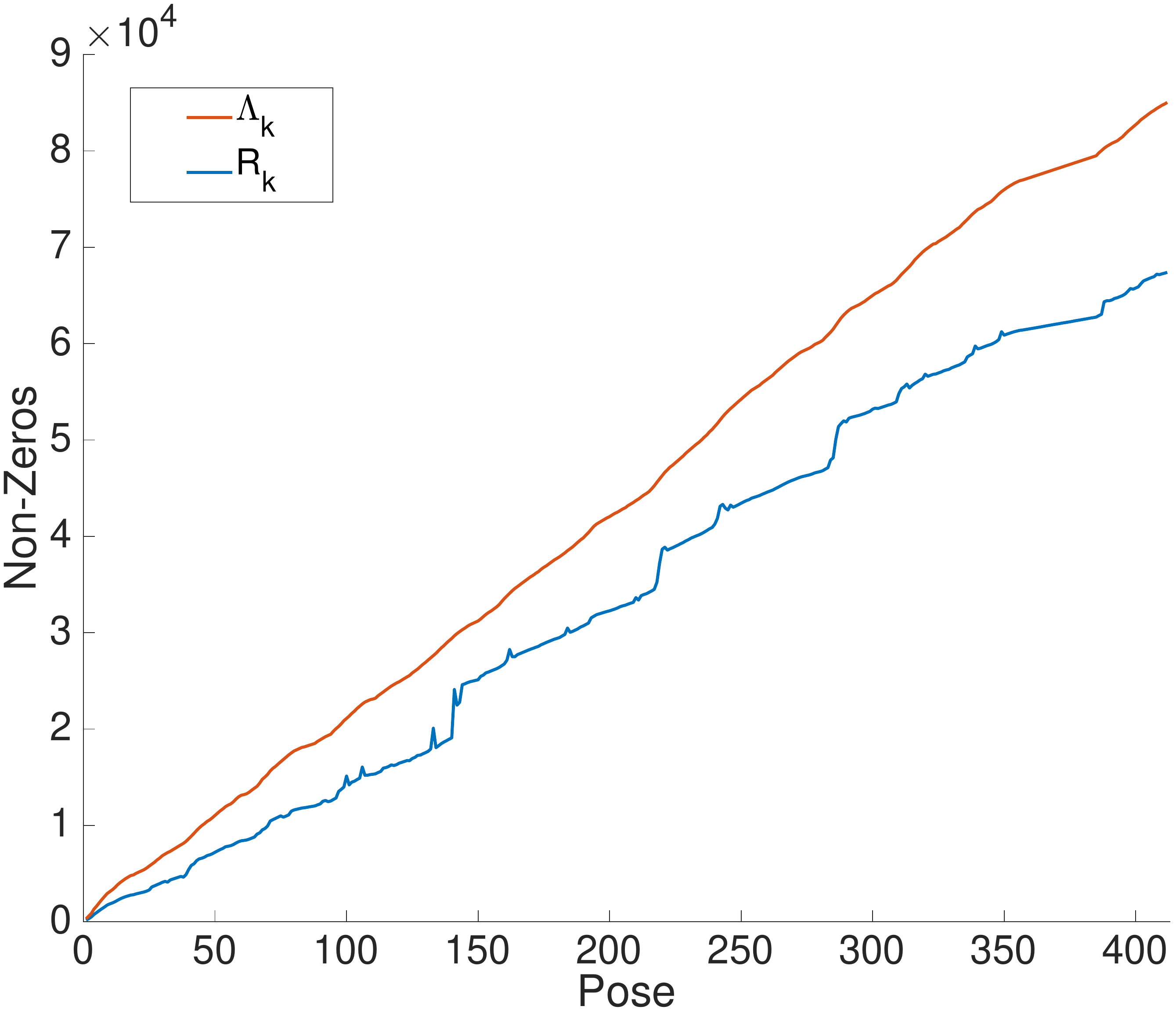}}
	\\
	\end{tabular}

	\protect
	\caption{\label{fig:CovFigA}
		Robot follows predefined path by navigating through given way-points.
		(a) Robot trajectory. Blue dots are mapped landmarks, red line with small
		ellipses is estimated trajectory with pose covariances, blue line is the real
		trajectory, pluses with numbers beside them are robot's predefined goals. Green mark is
		robot's start position;
		(b) Dimensions at each timestep of state vector $X$, of overall introduced and relinearized factors and of \involved variables in these factors;
		(c) Number of overall poses/landmarks inside state vector $X$ and number of relinearized poses/landmarks;
		(d) Number of non-zero entries inside the information matrix $\Lambda_k$ and the square-root information matrix $R_{k}$.
	}
\end{figure}

\begin{figure}[!t]
	\centering
	
	\begin{tabular}{cccc}
		\subfloat[\label{fig:CovFigB-a}]{\includegraphics[width=0.48\textwidth]{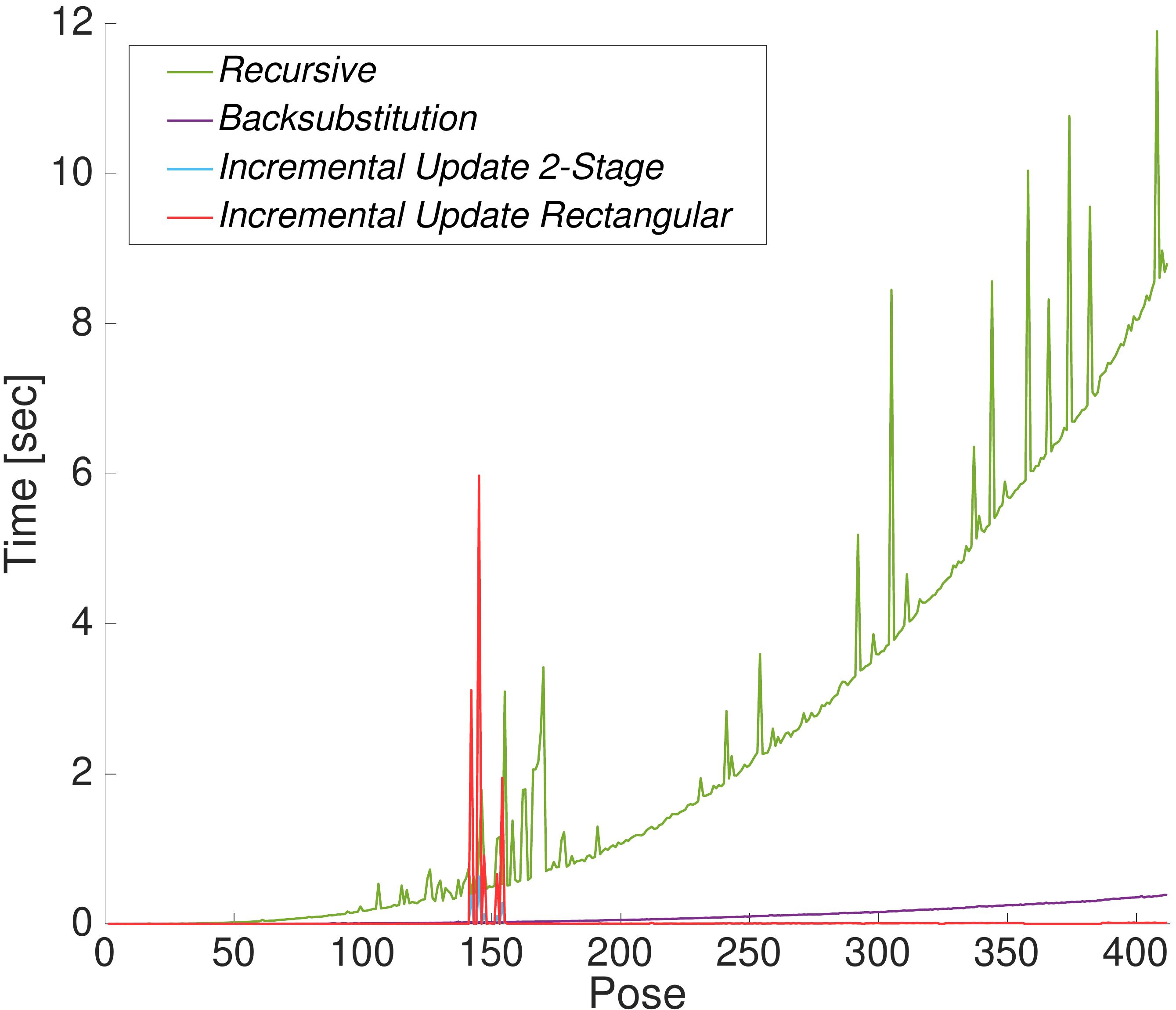}}
		&
		
		\subfloat[\label{fig:CovFigB-b}]{\includegraphics[width=0.48\textwidth]{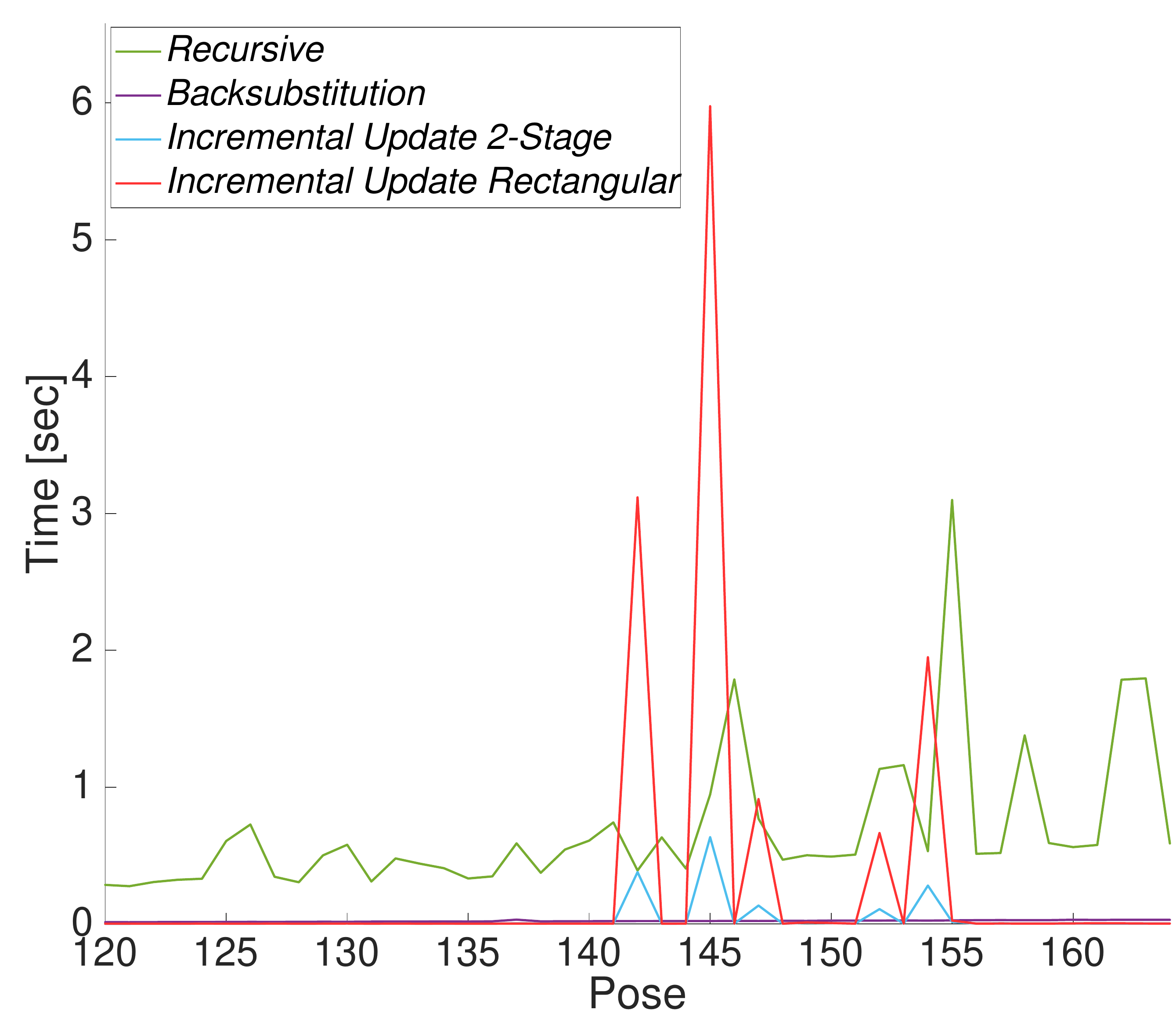}}
		\\
	\end{tabular}
	
	\begin{tabular}{cccc}
	\subfloat[\label{fig:CovFigB-c}]{\includegraphics[width=0.48\textwidth]{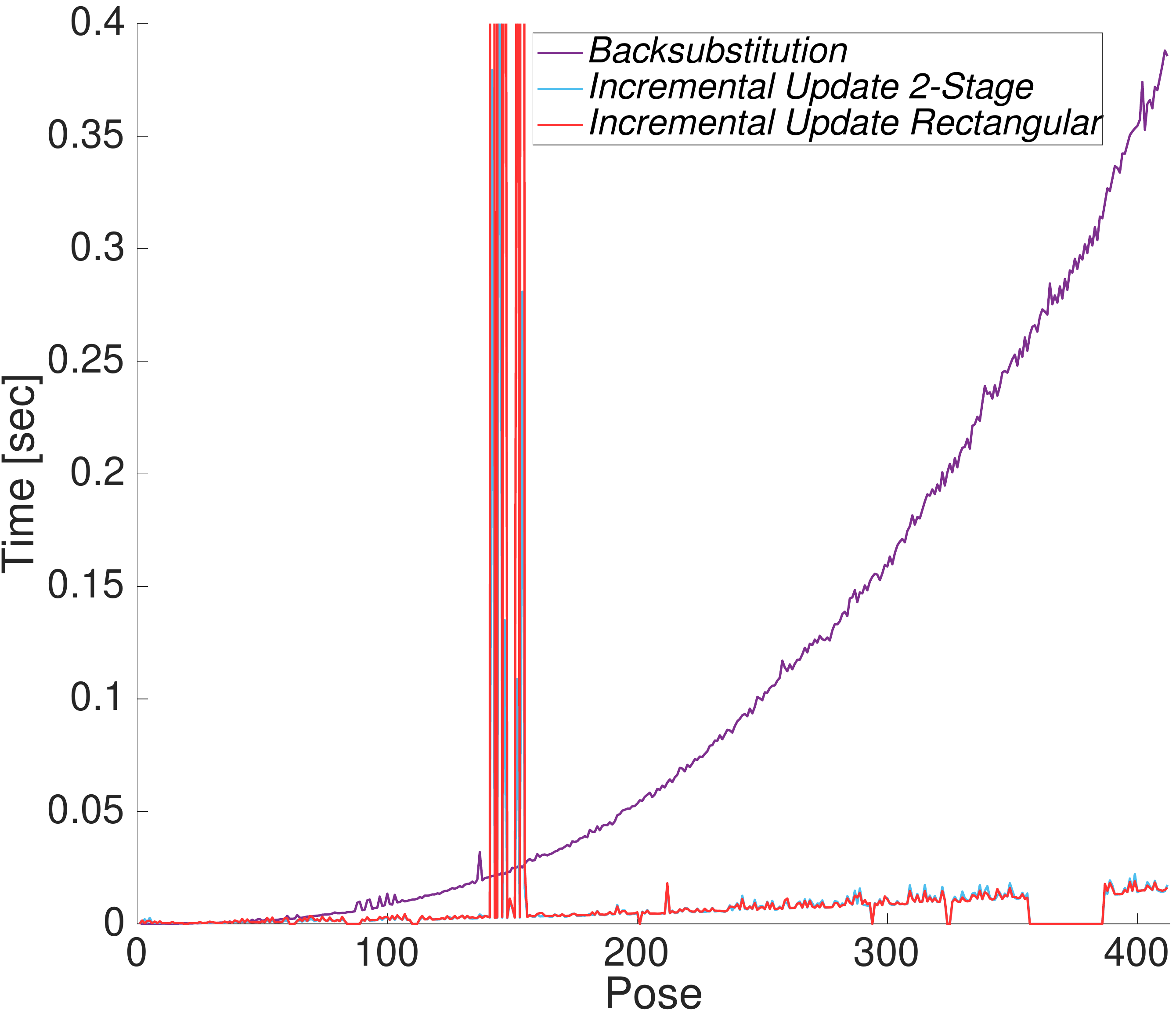}}
	&
	
	\subfloat[\label{fig:CovFigB-d}]{\includegraphics[width=0.48\textwidth]{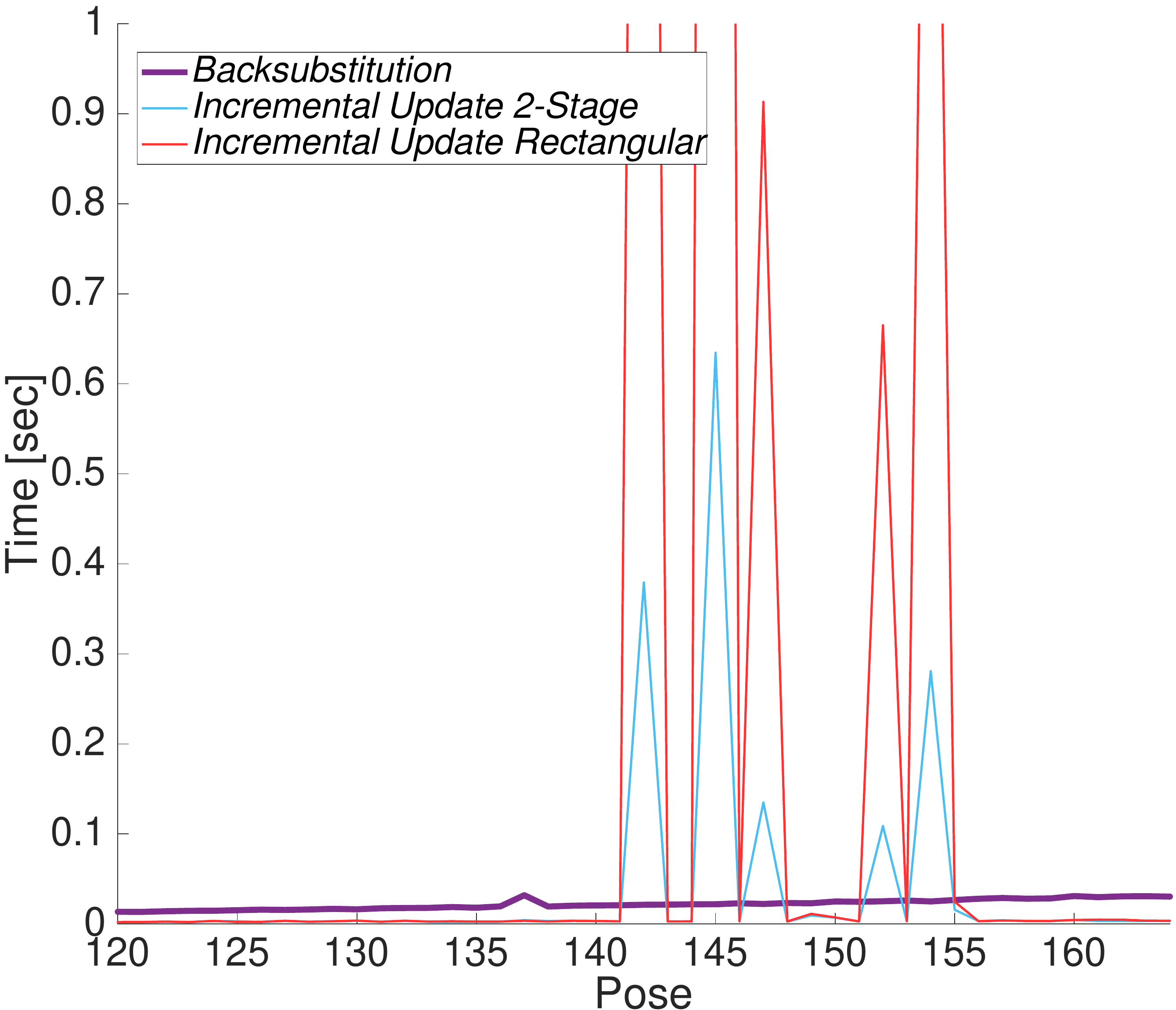}}
	\\
    \end{tabular}

	\protect
	\caption{\label{fig:CovFigB}
		Robot follows predefined path by navigating through given way-points.
		(a) Running time of marginal covariance recovery, i.e.~evaluating the marginal covariance matrix for each pose and each landmark;
		(b) Running time from (a) with zoom-in on loop-closure around pose 150;
		(c) Running time from (a) with zoom-in on three fastest approaches;
		(d) Running time from (c) with zoom-in on loop-closure around pose 150.
	}
\end{figure}

Here we consider the passive setting where at each time step the robot moves toward the next predefined goal (see Figure \ref{fig:CovFigA}), updates the inference problem with new pose/landmarks and motion/measurement factors, and calculates/updates marginal covariance of each variable inside the state vector.

We apply our incremental covariance update methods (\twostageapr and \rectapr) as it was described in Section \ref{sec:SLAMAppCov}. Their performance is compared with two baseline approaches. First, \recursive, uses a  recursive formulation (see  Eqs.~(\ref{eq:RecursCovRec1})-(\ref{eq:RecursCovRec2})) to calculate the covariance matrix $\Sigma_{k}$ ($k$ is index of time step) from a  square-root information matrix $R_{k}$. It is done for each $k$ and entire $\Sigma_{k}$ is calculated at each time step from scratch. Note that to calculate the marginal covariance of each state variable (block-diagonal of $\Sigma_{k}$) the \recursive method requires to calculate the entire covariance matrix $\Sigma_{k}$ as was explained in \textbf{Problem 1} from Section \ref{Sec_Notations}.

The second approach, \backsubstitution, calculates $\Sigma_{k}$ through the backsubstitution operation:
\begin{equation}
V \triangleq R_{k} \diagdown I
, \quad
\Sigma_{k} = V \cdot V^T
\end{equation}
where $I$ is an identity matrix of appropriate dimensions and "$\diagdown$" is the Matlab's backsubstitution operator with $x = A \diagdown B$ being identical to solving linear equations $Ax = B$ for $x$. Such backsubstitution can be done very efficiently since the matrix $R_{k}$ is upper triangular and sparse. Still, similar to \recursive, the \backsubstitution method calculates covariances from scratch for each time step and needs to calculate the entire $\Sigma_{k}$ matrix before fetching its diagonal blocks.

As can be seen in Figure \ref{fig:CovFigB}, in general both our incremental approaches have very similar runtime, and the both are significantly faster than the baseline alternatives. Towards the end of the scenario, while the fastest alternative (\backsubstitution) needs almost 400 ms to recover marginal covariance for a 3054-dimensional state vector, our incremental method does it in only 20 ms.

The only time our methods are slower than the alternatives is around pose 150, at which point a loop-closure event occurs: the robot reaches a predefined goal 6 (see Figure \ref{fig:CovFigA-a}) and observes old landmarks from the beginning of the scenario. As expected for such a relatively big loop-closure, the number of relinearized state variables and the affected factors is very large (see Figures \ref{fig:CovFigA-b}-\ref{fig:CovFigA-c}). Thus,  $m$ (overall dimension of new/relinearized factors) and $|\comb{X}{I}{}{}|$ (dimension of \involved variables) are huge and increase the runtime complexity of our incremental method. However, such results are expected; it is a known fact that incremental techniques become slower in presence of big loop-closures. For example, the incremental optimization algorithm iSAM2 \citep{Kaess12ijrr}, which calculates incrementally the MAP estimate of the state but not its covariance matrix, takes significantly more time during  loop-closure events. It is reasonable to expect a similar situation also in the context of incremental covariance recovery. Also note that during a loop-closure event, the \rectapr technique is significantly slower than the \twostageapr technique (around 6s vs 0.6s respectively). The reason is that during huge loop-closure, $m$ impacts the entire calculation of the \rectapr method (Lemma \ref{lemma:RectFFuncLemma}), while in the \twostageapr technique only the second stage is affected (Lemma \ref{lemma:NAugFFuncLemma} or Lemma \ref{lemma:RelinFFuncLemma}). Lemma \ref{lemma:RectFFuncLemma} is more computationally demanding than Lemma \ref{lemma:NAugFFuncLemma} or Lemma \ref{lemma:RelinFFuncLemma}, thus producing such a big runtime difference during a loop closure.

On the other hand, the \backsubstitution method does not depend on $m$ or $|\comb{X}{I}{}{}|$; instead its complexity mainly depends on the state dimension $n$ and the sparsity level of a matrix $R_k$. The $n$, the overall dimension of all state variables, is not affected by loop-closures. While in general the matrix $R_k$ (a factorization of information matrix $\Lambda_k$) becomes denser during the loop-closures, an appropriate variable reordering (of the entire matrix) can mitigate this effect. In our simulations we used SYMAMD ordering (symmetric approximate minimum degree permutation, \citep{Amestoy96siam}) to reorder an entire $\Lambda_k$ before producing $R_k$. As can be seen in Figure \ref{fig:CovFigA-d} (blue line), the resulting sparsity of $R_k$ grows smoothly with time, with only a minor increase during the loop-closure event (around pose 150).
Thus, we can see no peaks in calculation time plot of \backsubstitution approach around this time (see Figures \ref{fig:CovFigB-c}-\ref{fig:CovFigB-d}, purple line). In practice, when implementing our incremental approach on a real robot, to handle this loop-closure shortcoming we can check if a big loop-closure happens ($m$ or $|\comb{X}{I}{}{}|$ are bigger than current state dimension $n$) and use \backsubstitution as a fallback.

Comparing \recursive vs \backsubstitution we can see that the former is considerably slower. The first was implemented by us in C++ code, while the second is based on highly optimized Matlab implementation of backsubstitution. Apparently, our current C++ implementation of \recursive method is not properly optimized. We foresee that it can be done in much better way so that both \recursive vs \backsubstitution techniques will have very similar runtime complexity.

We note we did not compare our approach with the one from \citep{Ila15icra} since their method is limited and cannot be applied for every case of inference change, as was described already above. For example, the study \citep{Ila15icra} explicitly states that when any variable was relinearized during the change in the inference problem, the \recursive method is used to recover the marginal covariances. In Figure \ref{fig:CovFigA-c} we can see that this applies to the most of the changes in our scenario since small number of state variables is relinearized at almost any time step. Nonetheless, for cases supported by the technique from \citep{Ila15icra} we expect to see performance very similar to the one of our own approach, since runtime of both methods depends on $m$ and $|\comb{X}{I}{}{}|$.

Additionally, the incremental recovery of conditional covariance entries is essential to efficiently solve information-theoretic BSP problem which is considered in Section \ref{sec:BSPResults}; this important scenario is also not supported in \citep{Ila15icra}.

% ===============
\subsection{Belief Space Planning}\label{sec:BSPResults}

\begin{figure}[!t]
	\centering
	
	\begin{tabular}{cccc}
		\subfloat[\label{fig:AutoNavFigA-a}]{\includegraphics[width=0.48\textwidth]{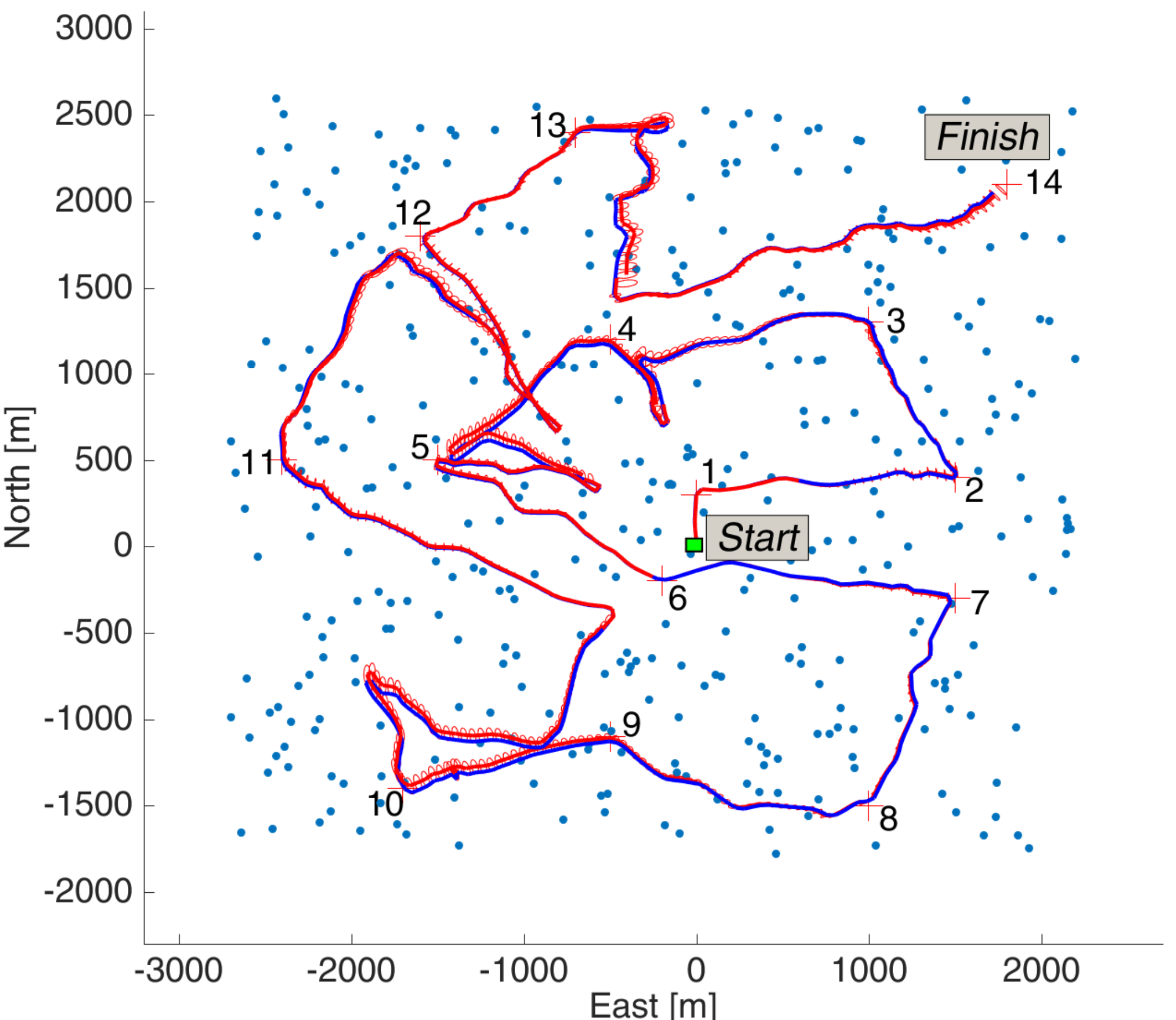}}
		&
		\subfloat[\label{fig:AutoNavFigA-b}]{\includegraphics[width=0.48\textwidth]{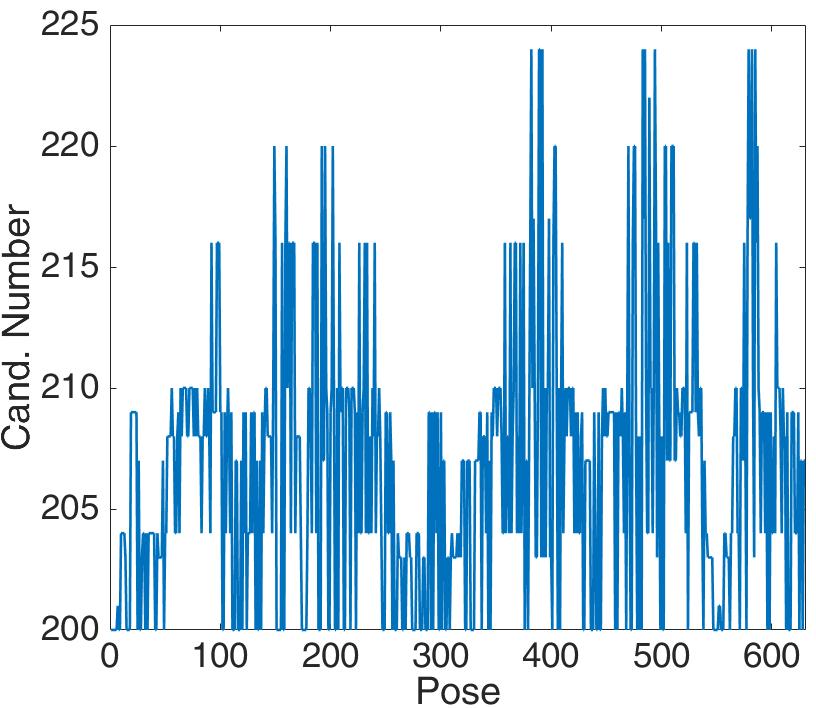}}
		\\
	\end{tabular}
	
	\begin{tabular}{cccc}
		\subfloat[\label{fig:AutoNavFigA-c}]{\includegraphics[width=0.48\textwidth]{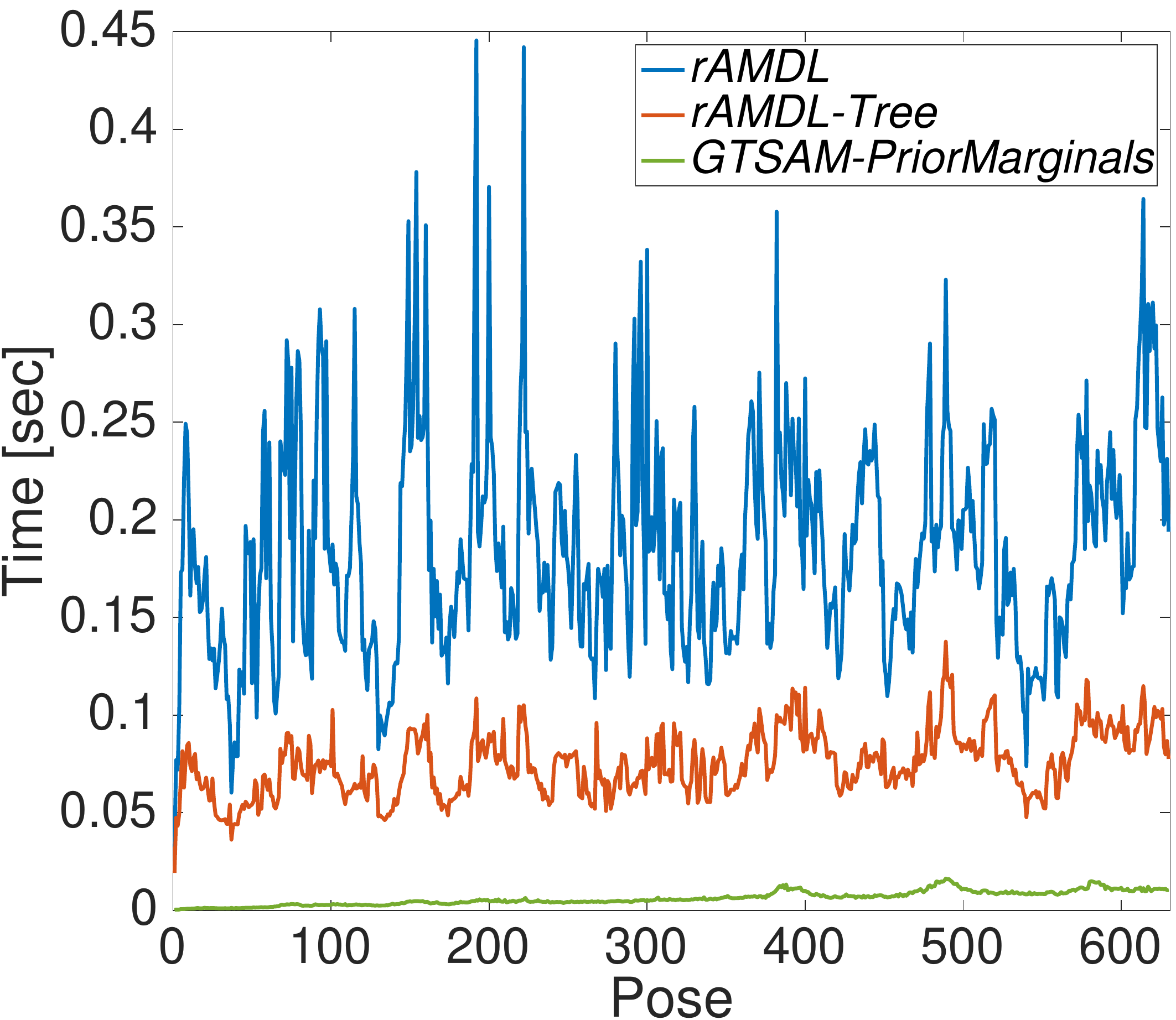}}
		&
		\subfloat[\label{fig:AutoNavFigA-d}]{\includegraphics[width=0.48\textwidth]{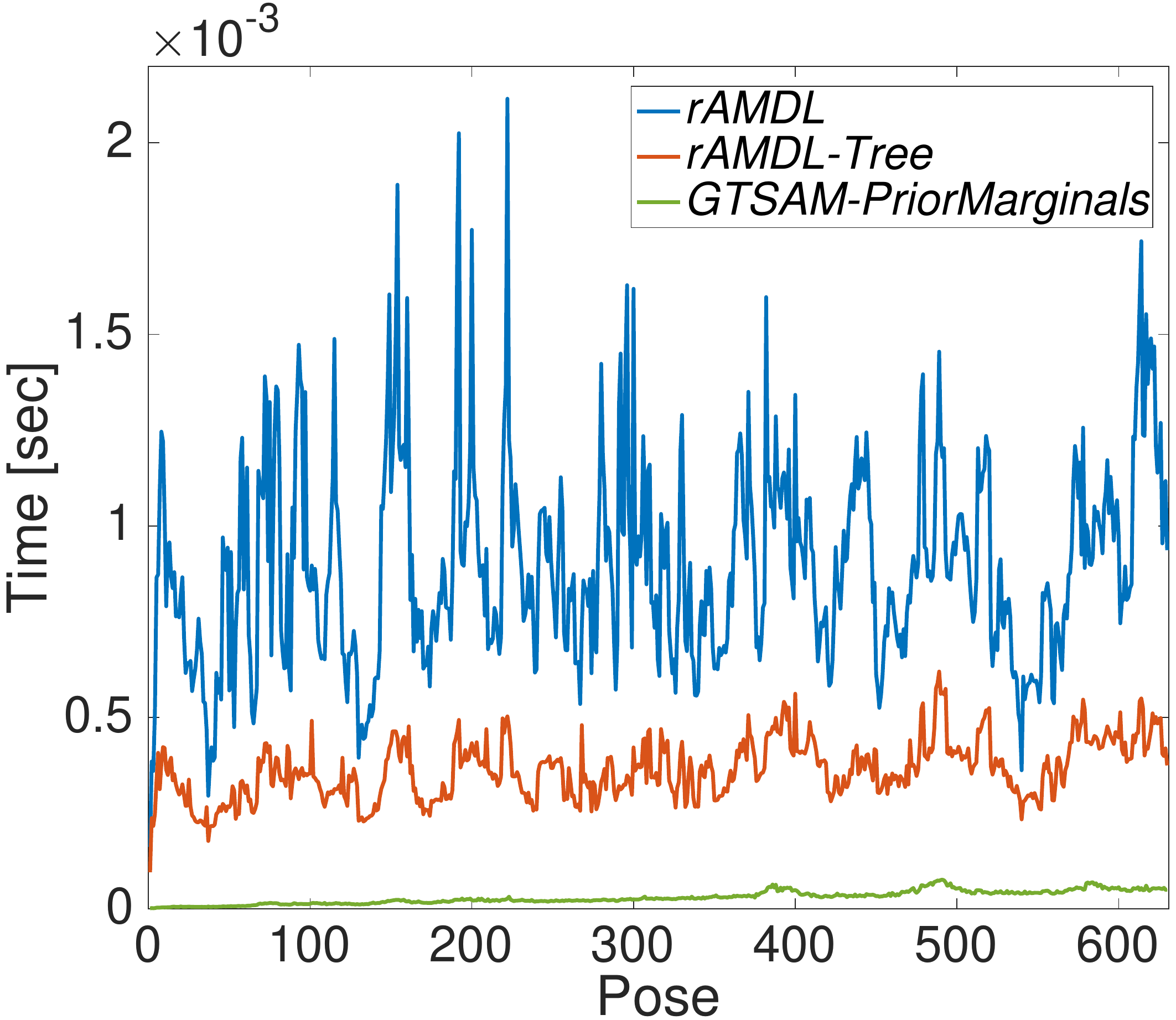}}
		\\
	\end{tabular}
	
	\protect
	\caption{\label{fig:AutoNavFigA}
		\Focused BSP scenario with \focused robot's last pose.
		(a) Final robot trajectory. Blue dots are mapped landmarks, red line with small
		ellipses is estimated trajectory with pose covariances, blue line is the real
		trajectory, red pluses with numbers beside them are robot's predefined goals. Green mark is
		robot's start position;
		(b) Number of action candidates at each time;
		(c) Running time of planning, i.e.~evaluating impact of all candidate actions,
		each representing possible trajectory;
		(d) Running time from (c) normalized by number of candidates.
	}
\end{figure}

\begin{figure}[!t]
	\centering
	
	\begin{tabular}{cccc}
		
		\subfloat[\label{fig:AutoNavFigB-a}]{\includegraphics[width=0.31\textwidth]{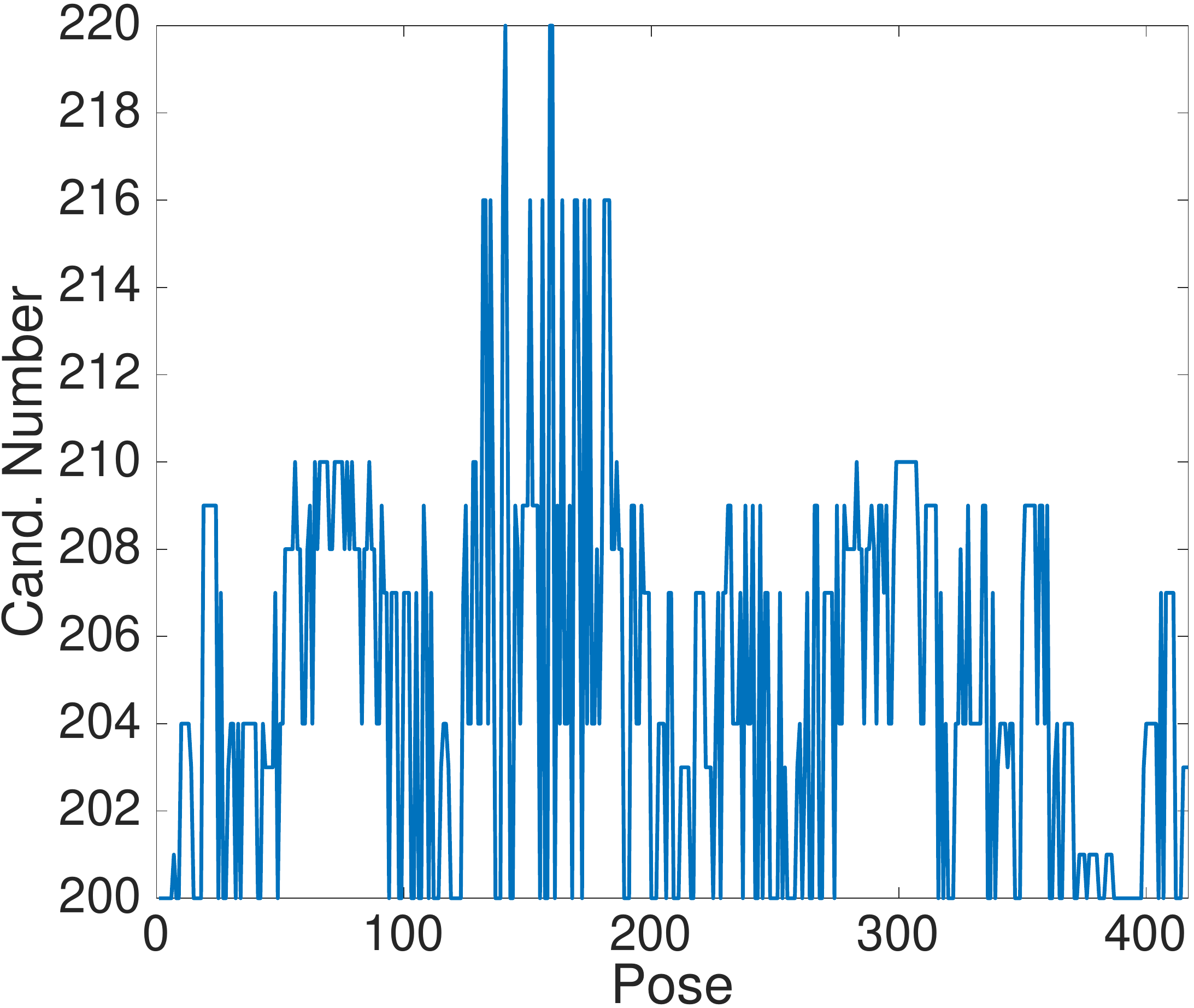}}
		&
		
		\subfloat[\label{fig:AutoNavFigB-b}]{\includegraphics[width=0.31\textwidth]{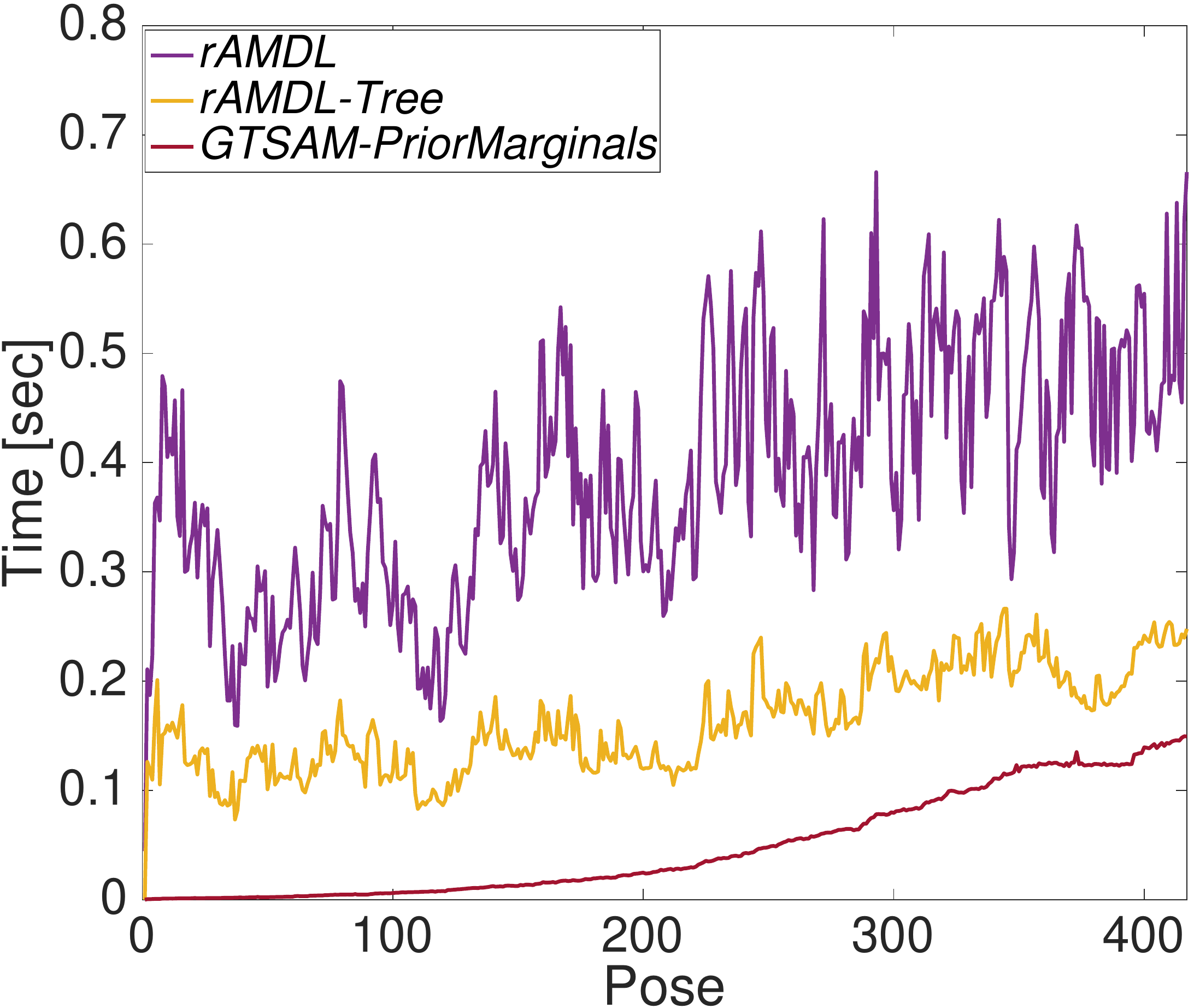}}
		&
		
		\subfloat[\label{fig:AutoNavFigB-c}]{\includegraphics[width=0.31\textwidth]{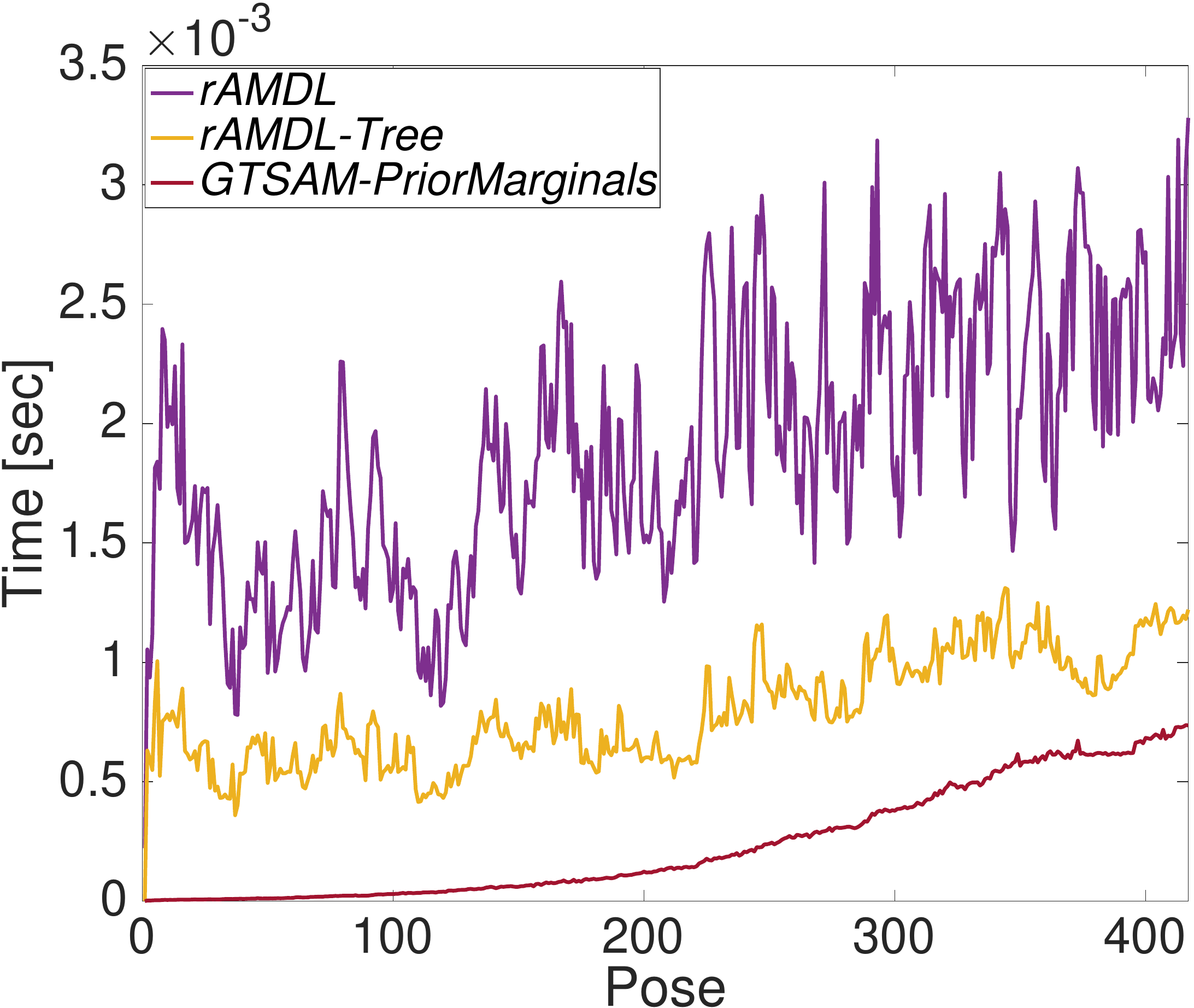}}
		\\
	\end{tabular}
	
	\protect
	\caption{\label{fig:AutoNavFigB}
		\Focused BSP scenario with \focused mapped till now landmarks.
		(a) Number of action candidates at each time;
		(b) Running time of planning, i.e.~evaluating impact of all candidate actions,
		each representing possible trajectory;
		(c) Running time from (b) normalized by number of candidates.
	}
\end{figure}

Thus far, we performed simulation of a passive SLAM problem, where robot follows a predefined trajectory. As can be seen in Figure \ref{fig:CovFigA-a}, by the end of the trajectory the covariance of robot position (red ellipse) is considerably big. Such uncertainty in robot localization may fail the navigation task and is undesirable in general. In this section we focus on an active SLAM scenario, where the robot autonomously decides whether to follow the navigation path or to perform a loop-closure and reduce state uncertainty. At each time step, the robot autonomously decides its next action according to a specified objective function that is discussed below.

We compare performance of the proposed BSP approach, that we denote as \ramdltree, with our previous method \ramdl
\citep{Kopitkov17ijrr}, which was shown to be superior in run-time
complexity to  other state-of-the-art information-based BSP methods. Note that
both \ramdl and \ramdltree, as well as other relevant state-of-the-art
alternatives, make \emph{identical} decisions, i.e.~calculate the same optimal
actions.  Thus, the only difference is the run-time complexity, reduction of
which is the main motivation behind the work presented herein.

In our simulation, at each time step we sample a set of trajectories to the
current goal $g$, and also to (clusters of) already mapped landmarks for
uncertainty reduction via loop closures (see Figure \ref{fig:AutoNavFigA-a}).
The overall number of candidate actions (number of trajectories) is around 200.
We consider the objective function 
\begin{equation}
J(a) = \alpha_1 d(x_{k+L}, g) + \alpha_2
c(a) + \alpha_3 J_{inf}(a),
\end{equation}
where $d(x_{k+L}, g)$ is the distance between
the current goal $g$ and candidate's last pose $x_{k+L}$ for a given action $a$,
$c(a)$ is the control cost,  and $J_{inf}(a)$ is an information-theoretic term.
As was mentioned before, both first and second terms can be calculated very fast
and do not require belief propagation. Thus, in the sequel we will ignore these
terms and discuss run time only for the term $J_{inf}(a)$.

In our first simulation, $J_{inf}(a)$ calculates the posterior entropy of
robot's last pose $x_{k+L}$ within the candidate trajectory. We evaluated
this term for all  action candidates independently through our previous
method, \ramdl, and through the proposed-herein approach, \ramdltree, which, using the FGP action tree, accounts for candidate actions' mutual parts and evaluates them only once. In Figures
\ref{fig:AutoNavFigA-c}-\ref{fig:AutoNavFigA-d} we can see that \ramdltree is
twice faster than \ramdl and succeeds to evaluate more than 200 actions in less
than 100ms. Also, we can see that the only algorithmic part that depends on the state
dimension $n$, i.e.~the one-time calculation of prior covariances at $G_{-}$
during the incremental covariance update termed in the figure as \emph{GTSAM-PriorMarginals}, takes a small portion of the overall
run-time (green lines in Figures
\ref{fig:AutoNavFigA-c}-\ref{fig:AutoNavFigA-d}); most of the time is consumed
by propagation of covariance entries within the FGP action tree and IG calculation for each edge in this tree.  Note that the marginal/conditional covariances required by \ramdltree are propagated from the root of FGP action tree to its leafs based on our incremental covariance update technique (see Section \ref{sec:IncCovUpdate}). Specifically, as described in Section \ref{sec:IncCovUpdateFGPTree}, covariance propagation is performed in two phases. In the first one, each tree node in bottom-to-top order determines what covariance entries are required from its belief. In the second phase the required covariances are calculated in top-to-bottom order using incremental covariance update lemmas (see also Figure \ref{fig:TreeIncCovUpdate}).

Additionally, we have performed a similar simulation considering this time
$J_{inf}(a)$ calculating the IG of the landmarks that were mapped till now. The
results are shown in Figure \ref{fig:AutoNavFigB}. Comparing the
time-performance between the first and second scenarios we can see that
"focused-landmarks" requires more time; while \ramdltree takes around 50-100ms in the first one, it requires 100-250ms in the second. This is due to the fact that calculation of the 
\focused IG contains a one-time computation that depends on the dimension of \focused
variables set $X^F_{+}$. In "focused-landmarks" scenario this is the dimension
of all landmarks mapped thus far, and it increases with time as more landmarks
are observed and introduced into the state vector. Also in this scenario we see
a similar trend where \ramdltree performs twice faster than \ramdl (100-250ms
vs 200-500ms), while determining the same optimal actions.

% ===============
\section{Conclusions}\label{sec:Conclusions}

We developed computationally efficient approaches that address incremental covariance recovery and BSP over high-dimensional state spaces. %in challenging high-dimensional settings. 
Our incremental covariance update technique allows
to efficiently update specific covariance entries (both marginal and conditional) after any change in the inference problem, including introduction of new state variables, addition of new measurement factors, and relinearization of the entire state vector or only a subset of the state variables. It can be applied whenever an efficient method is required to track covariance entries within the estimation system (e.g.~in SLAM for data association or safety), and is also an indispensable part of our BSP approach. 

Furthermore, considering the BSP problem our key observation is that in many robotics applications,  candidate actions have mutual parts where each part can be evaluated only once, independently of the number of candidate actions that share it.
For this purpose, we  presented  a novel approach to model future posterior
beliefs of different candidate actions within a single graphical model which we
called \emph{factor-graph propagation} (FGP) action tree. This tree model allows
to evaluate shared parts of different actions only once by representing belief
propagation of posterior factor graphs from current factor graph sequentially.
The FGP action tree has a consecutive hierarchic form, with intermediate vertices that represent 
beliefs after applying only part of a candidate action.	Further, we use the aforementioned incremental covariance recovery technique to efficiently calculate covariances at intermediate and final
state beliefs within the FGP action tree, doing so independently of state
dimension $n$. The calculated covariance entries let us  reason about
probabilistic properties of the beliefs and actions within the FGP action tree.
Specifically, this allowed us to efficiently calculate information impact of all
candidate actions by re-using calculation from candidates' mutual parts. Overall, our 
method involves two passes over the FGP action tree, bottom-to-top query of required
covariance entries and top-to-bottom propagation of these covariance entries.
We evaluated the proposed approach in simulation considering the problem of
autonomous navigation in unknown environments, and showed it reduces run-time
twice compared to our  previous approach \citep{Kopitkov17ijrr},
\ramdl.

There are several avenues for future research to take the proposed concept of re-using calculations between different candidate actions further. In this work we considered a specific realization of the FGP action tree, using
the structure of candidate trajectories in an autonomous navigation scenario.
However, given posterior factor graphs for different candidate actions, multiple
FGP action trees can be constructed. A key question that will be addressed as
part of future research is how to construct an FGP action tree so that most of
actions' similarity would be exploited.  Another direction for future research is to make BSP runtime complexity totally independent of state dimension.
The process of information evaluation via FGP action tree consists of only a single, one-time, calculation that depends on state dimension $n$,
i.e.~recovering the prior marginal (or conditional) covariance entries of
variables \involved in candidate actions. When there are many candidates (above
500 in our simulations), this one-time part is insignificant w.r.t. overall
process time and can be ignored. However, for a smaller number of candidate
actions this part takes considerable time (about 50\% of time in our simulation in Figure \ref{fig:AutoNavFigB}) and additional research efforts are required in order to
reduce its time complexity. Anticipating what state variables would be involved in the near-future candidate actions and incrementally tracking specific
covariance entries of these involved variables during the inference process through our incremental covariance recovery method may be an appropriate solution
and will be investigated in the future.

\section*{Acknowledgments}

This work was  supported by the Israel Science Foundation.

% ======================
%\bibliographystyle{plain}

\section{Appendix}
\label{sec:AppndSec}

\subsection{Proof of Lemma \ref{lemma:NAugFFuncLemma} - \Actnaug Case}
\label{sec:NAugFFuncLemmaProof}

The variables set $W$ in this case is $\{ Y_{old}, \comb{X}{I}{}{}\} = \{ Y, \comb{X}{I}{}{}\}$. Define prior marginal covariance matrices: $\Sigma_{-}^I \equiv \Sigma_{-}^{M,\comb{X}{I}{}{}}$, $\Sigma_{-}^{Y} \equiv \Sigma_{-}^{M, Y}$. Also denote the prior cross covariance between $Y$ and $\comb{X}{I}{}{}$ as $\Sigma_{-}^C$. Then, $\Sigma_{-}^{M,W}$ will have the following form:
\begin{equation}
\Sigma_{-}^{M,W} =
\begin{pmatrix} 
\Sigma_{-}^{Y} & \Sigma_{-}^C\\
(\Sigma_{-}^C)^T & \Sigma_{-}^I\\
\end{pmatrix}.
\end{equation}
Additionally, let us separate prior (old) state variables $X_{-}$ into \involved $\comb{X}{I}{}{}$ (in new factors $F_{new}$) and \notinvolved $\comb{X}{\neg I}{}{}$ variables. Similarly, let us partition the Jacobian matrix $A$ into:
\begin{equation}
A =
\begin{pmatrix} 
\comb{A}{\neg I}{}{} & \comb{A}{I}{}{}\\
\end{pmatrix}
=
\begin{pmatrix} 
0 & \comb{A}{I}{}{}\\
\end{pmatrix},
\end{equation}
where $\comb{A}{\neg I}{}{}$ contains noise-weighted Jacobians w.r.t. $X^{\neg I}$, and $\comb{A}{I}{}{}$ w.r.t. $\comb{X}{I}{}{}$. From its definition we can conclude that $\comb{A}{\neg I}{}{}$ contains only zeros.

Next, using the Woodbury matrix identity and information update equation $\Lambda_{+} = 
\Lambda_{-} + A^T \cdot A$, the posterior covariance matrix is:
\begin{equation}
\Sigma_{+} =
(\Lambda_{+})^{-1}
=
(\Lambda_{-} + A^T \cdot A)^{-1}
=
\Sigma_{-} -
\Sigma_{-}
\cdot
A^T
\cdot
[I_m +
A \cdot 
\Sigma_{-}
\cdot A^T ]^{-1}
\cdot
A
\cdot
\Sigma_{-}
=
\Sigma_{-} -
\Sigma_{-}
\cdot
A^T
\cdot
[I_m +
\comb{A}{I}{}{} \cdot 
\Sigma_{-}^{I}
\cdot 
(\comb{A}{I}{}{})^T
]^{-1}
\cdot
A
\cdot
\Sigma_{-}
\end{equation}
where $A \cdot 
\Sigma_{-}
\cdot A^T = \comb{A}{I}{}{} \cdot 
\Sigma_{-}^{I}
\cdot 
(\comb{A}{I}{}{})^T$ because of $A$'s sparsity structure.

Then $\Sigma_{+}$ can be calculated as:
\begin{equation}
\Sigma_{+} =
\Sigma_{-} -
\Sigma_{-}
\cdot
A^T
\cdot
C^{-1}
\cdot
A
\cdot
\Sigma_{-}
\end{equation}
\begin{equation}
C = I_m + 
\comb{A}{I}{}{} \cdot \Sigma_{-}^{I} \cdot (\comb{A}{I}{}{})^T.
\end{equation}

Further, $\Sigma_{+}^{M,Y}$ can be calculated by retrieving from $\Sigma_{+}$ rows and columns that belong to variables $Y$:
\begin{multline}
\Sigma_{+}^{M,Y}
=
\Sigma_{-}^{M,Y} -
\Sigma_{-}^{(Y,:)}
\cdot
A^T
\cdot
C^{-1}
\cdot
A
\cdot
\Sigma_{-}^{(:,Y)}
=
\Sigma_{-}^{M,Y} -
[A
\cdot
\Sigma_{-}^{(:,Y)}]^T
\cdot
C^{-1}
\cdot
[A
\cdot
\Sigma_{-}^{(:,Y)}]
=\\
=
\Sigma_{-}^{M,Y} -
[\comb{A}{I}{}{}
\cdot
(\Sigma_{-}^C)^T
]^T
\cdot
C^{-1}
\cdot
[\comb{A}{I}{}{}
\cdot
(\Sigma_{-}^C)^T]
=
\Sigma_{-}^{M,Y} -
[\Sigma_{-}^{C}
\cdot
(\comb{A}{I}{}{})^T]
\cdot
C^{-1}
\cdot
[\Sigma_{-}^{C}
\cdot
(\comb{A}{I}{}{})^T]^T
\end{multline}
where using  Matlab syntax we have $\Sigma_{-}^{(Y,:)} \doteq \Sigma_{-} (Y,:)$ and $\Sigma_{-}^{(:,Y)} \doteq  \Sigma_{-} (:,Y)$.
\hfill $\blacksquare$

Note that the columns inside information matrices do not have to be ordered in any particular way, and that the provided above proof is correct for any ordering whatsoever.

\subsection{Proof of Lemma \ref{lemma:RectFFuncLemma} - \Actrect Case}
\label{sec:RectFFuncLemmaProof}

In this case we can partition variables set $Y$ into two subsets $Y_{old} \doteq X_{-} \cap Y$ and $Y_{new} \doteq X_{new} \cap Y$, or in other words, into old and new state variables. The posterior marginal covariance matrix $\Sigma_{+}^{M, Y}$ will have then the following form:
\begin{equation}
\Sigma_{+}^{M, Y} =
\begin{pmatrix} 
\Sigma_{+}^{M, Y_{old}} & \Sigma_{+}^{(Y_{old}, Y_{new})}\\
(\Sigma_{+}^{(Y_{old}, Y_{new})})^T & \Sigma_{+}^{M, Y_{new}}\\
\end{pmatrix},
\label{eq:PosteriorMargStr}
\end{equation}
and we are looking for an efficient way to calculate matrices $\Sigma_{+}^{M, Y_{old}}$, $\Sigma_{+}^{M, Y_{new}}$ and $\Sigma_{+}^{(Y_{old}, Y_{new})}$.

The variables set $W$ in this case is $\{ Y_{old}, \comb{X}{I}{}{}\}$. Define next the prior marginal covariance matrices: $\Sigma_{-}^I \equiv \Sigma_{-}^{M,\comb{X}{I}{}{}}$, $\Sigma_{-}^{Y_{old}} \equiv \Sigma_{-}^{M, Y_{old}}$. Also denote the prior cross covariance between $Y_{old}$ and $\comb{X}{I}{}{}$ as $\Sigma_{-}^C$. Then,  $\Sigma_{-}^{M,W}$ will have the following form:
\begin{equation}
\Sigma_{-}^{M,W} =
\begin{pmatrix} 
\Sigma_{-}^{Y_{old}} & \Sigma_{-}^C\\
(\Sigma_{-}^C)^T & \Sigma_{-}^I\\
\end{pmatrix}.
\label{eq:PriorMargStr}
\end{equation}
Additionally, let us separate prior (old) state variables $X_{-}$ into \involved $\comb{X}{I}{}{}$ (in new factors $F_{new}$) and \notinvolved $\comb{X}{\neg I}{}{}$. The posterior state vector is then $X_{+} = \{ \comb{X}{I}{}{}, \comb{X}{\neg I}{}{}, X_{new} \}$. Similarly, let us partition the Jacobian matrix $A$ into:
\begin{equation}
A =
\begin{pmatrix} 
A_{old} & A_{new}\\
\end{pmatrix}
, \quad
A_{old}
=
\begin{pmatrix} 
\comb{A}{\neg I}{}{} & \comb{A}{I}{}{}\\
\end{pmatrix}
=
\begin{pmatrix} 
0 & \comb{A}{I}{}{}\\
\end{pmatrix},
\label{eq:AMatStr}
\end{equation}
where $A_{old}$ contains noise-weighted Jacobians w.r.t.~old variables $X_{-}$, $A_{new}$  w.r.t.~new variables $X_{new}$, $\comb{A}{\neg I}{}{}$ w.r.t.~$\comb{X}{\neg I}{}{}$, and $\comb{A}{I}{}{}$ w.r.t.~$\comb{X}{I}{}{}$. From its definition we can conclude that $\comb{A}{\neg I}{}{}$ contains only zeros.

Following the information update equation $\Lambda_{+} = 
\Lambda_{+}^{Aug} + A^T \cdot A$ (see also Figure \ref{fig:AugmentInfoMAtBSFig}), the posterior information matrix can be partitioned using separation $X_{+} = \{ X_{-}, X_{new} \}$ as:
\begin{equation}
\Lambda_{-} =
\begin{pmatrix} 
\Lambda_{-} + A_{old}^T \cdot A_{old}  &
A_{old}^T \cdot A_{new}\\
A_{new}^T \cdot A_{old} & 
A_{new}^T \cdot A_{new}\\
\end{pmatrix}.
\end{equation}
Now, let us partition the posterior covariance matrix $\Sigma_{+}$ in a similar way:
\begin{equation}
\Sigma_{+} =
\begin{pmatrix} 
\Sigma_{+}^{old} & \Sigma_{+}^{cross}\\
(\Sigma_{+}^{cross})^T & \Sigma_{+}^{new}\\
\end{pmatrix}.
\label{eq:PosteriorCovStr}
\end{equation}
Giving the setup till now, we will derive each of the matrices $\Sigma_{+}^{M, Y_{old}}$, $\Sigma_{+}^{M, Y_{new}}$ and $\Sigma_{+}^{(Y_{old}, Y_{new})}$ from Eq.~(\ref{eq:PosteriorMargStr}) using parts from $\Sigma_{-}^{M,W}$ defined in Eq.~(\ref{eq:PriorMargStr}).

\subsubsection*{$\boldsymbol{\Sigma_{+}^{M, Y_{new}}}$:}

By using block-wise matrix inversion (which is based on the notion of Schur Complements), $\Sigma_{+}^{new}$ is equal to:
\begin{equation}
\Sigma_{+}^{new} =
(A_{new}^T \cdot A_{new} -
A_{new}^T \cdot A_{old} \cdot
(\Lambda_{-} + A_{old}^T \cdot A_{old})^{-1} \cdot
A_{old}^T \cdot A_{new})^{-1}
=
(A_{new}^T \cdot (I_m -
A_{old} \cdot
(\Lambda_{-} + A_{old}^T \cdot A_{old})^{-1} \cdot
A_{old}^T) \cdot A_{new})^{-1}
\end{equation}
Now, let's define matrix $C$ as following:
\begin{equation}
C 
\triangleq
I_m + 
A_{old} \cdot \Sigma_{-} \cdot A_{old}^T
=
I_m + 
\comb{A}{I}{}{} \cdot \Sigma_{-}^{I} \cdot (\comb{A}{I}{}{})^T.
\label{eq:CMatDef}
\end{equation}
Through Woodbury matrix identity it can be easily shown that $C$'s inverse is:
\begin{equation}
C^{-1} =
I -
A_{old} \cdot
(\Lambda_{-} + A_{old}^T \cdot A_{old})^{-1} \cdot
A_{old}^T.
\end{equation}
Therefore, $\Sigma_{+}^{new}$ is equal to:
\begin{equation}
\Sigma_{+}^{new} =
(A_{new}^T \cdot
C^{-1}
\cdot A_{new})^{-1},
\label{eq:PosteriorNewCov}
\end{equation}
and $\Sigma_{+}^{M, Y_{new}}$ can be calculated in the following way (note that such calculation's complexity is independent of state dimension):
\begin{equation}
\Sigma_{+}^{M, Y_{new}} =
P
^{(Y_{new}, :)}
, \quad
P
\triangleq
[(A_{new}^T \cdot
C^{-1}
\cdot A_{new})^{-1}]
^{(:, Y_{new})},
\label{eq:PMatDef}
\end{equation}
where in brackets we are using Matlab syntax to index relevant rows/columns. Note that $[(A_{new}^T \cdot
C^{-1}
\cdot A_{new})^{-1}]
^{(:, Y_{new})}$ can be calculated without calculation of full $(A_{new}^T \cdot
C^{-1}
\cdot A_{new})^{-1}$, by using backslash operator in Matlab:
\begin{equation}
P =
[A_{new}^T \cdot
C^{-1}
\cdot A_{new}]
\backslash
I^{(:, Y_{new})},
\end{equation}
where $I^{(:, Y_{new})}$ are particular columns of identity matrix.

\subsubsection*{$\boldsymbol{\Sigma_{+}^{M, Y_{old}}}$:}

Using the block-wise matrix inversion again we know that  $\Sigma_{+}^{old}$ from Eq.~(\ref{eq:PosteriorCovStr}) is equal to:
\begin{multline}
\Sigma_{+}^{old} =
(\Lambda_{-} + A_{old}^T \cdot A_{old} -
A_{old}^T \cdot A_{new} \cdot
(A_{new}^T \cdot A_{new})^{-1} \cdot
A_{new}^T \cdot A_{old})^{-1}
=\\
=
(\Lambda_{-} +
A_{old}^T \cdot (I_m -
A_{new} \cdot
(A_{new}^T \cdot A_{new})^{-1} \cdot
A_{new}^T) \cdot A_{old})^{-1}
=
(\Lambda_{k} +
A_{old}^T \cdot K \cdot A_{old})^{-1}
\end{multline}
with
\begin{equation}
K
\triangleq
I_m -
A_{new} \cdot
(A_{new}^T \cdot A_{new})^{-1} \cdot
A_{new}^T
=
I_m -
A_{new} \cdot
F \cdot
A_{new}^T
, \quad
F
\triangleq
(A_{new}^T \cdot A_{new})^{-1},
\label{eq:FMatDef}
\end{equation}
where $K$ is a singular, symmetric, idempotent projection matrix, with properties $K = K^2$ and $K = K^T$.

Further, $\Sigma_{+}^{old}$ can be now rewritten as:
\begin{multline}
\Sigma_{+}^{old} =
(\Lambda_{-} +
A_{old}^T \cdot K^T \cdot K \cdot A_{old})^{-1}
=
\Lambda_{-}^{-1} -
\Lambda_{-}^{-1} \cdot
A_{old}^T \cdot K^T \cdot
(I_m +
K \cdot A_{old} \cdot
\Lambda_{-}^{-1} \cdot
A_{old}^T \cdot K^T
)^{-1} \cdot
K \cdot A_{old} \cdot
\cdot \Lambda_{-}^{-1}
=\\
=
\Sigma_{-} -
\Sigma_{-} \cdot
A_{old}^T \cdot K^T \cdot
(I_m +
K \cdot A_{old} \cdot
\Sigma_{-} \cdot
A_{old}^T \cdot K^T
)^{-1} \cdot
K \cdot A_{old} \cdot
\cdot \Sigma_{-}
=
\Sigma_{-} -
\Sigma_{-} \cdot
A_{old}^T \cdot K^T \cdot
G^{-1} \cdot
K \cdot A_{old} 
\cdot \Sigma_{-}
\end{multline}
with
\begin{equation}
G
\triangleq
I_m +
K \cdot A_{old} \cdot
\Sigma_{-} \cdot
A_{old}^T \cdot K^T
=
I_m +
K \cdot 
\comb{A}{I}{}{} \cdot \Sigma_{-}^{I} \cdot (\comb{A}{I}{}{})^T
\cdot K^T
=
I_m +
K_1 \cdot \Sigma_{-}^{I} \cdot K_1^T
\end{equation}
\begin{equation}
K_1
\triangleq
K \cdot 
\comb{A}{I}{}{},
\label{eq:K1MatDef}
\end{equation}
where $K_1$ are non-zero columns from $A_{old}$ projected outside of vector space that is spanned by columns in $A_{new}$. In other words, $K_1$ contains information from $A_{old}$ that is not contained within $A_{new}$.

Then $\Sigma_{+}^{M, Y_{old}}$ can be calculated by retrieving from $\Sigma_{+}^{old}$ rows and columns that belong to variables $Y_{old}$:
\begin{multline}
\Sigma_{+}^{M, Y_{old}}
=
\Sigma_{-}^{M,Y_{old}} -
\Sigma_{-}^{(Y_{old},:)}
\cdot
A_{old}^T \cdot K^T \cdot
G^{-1} \cdot
K \cdot A_{old}
\cdot
\Sigma_{-}^{(:,Y_{old})}
=
\Sigma_{-}^{M,Y_{old}} -
[K \cdot A_{old}
\cdot
\Sigma_{-}^{(:,Y_{old})}]^T
\cdot
G^{-1} \cdot
[K \cdot A_{old}
\cdot
\Sigma_{-}^{(:,Y_{old})}]
=\\
=
\Sigma_{-}^{M,Y_{old}} -
[K \cdot
\comb{A}{I}{}{}
\cdot
\Sigma_{-}^{(X^I,Y_{old})}]^T
\cdot
G^{-1} \cdot
[K \cdot \comb{A}{I}{}{}
\cdot
\Sigma_{-}^{(X^I,Y_{old})}]
=
\Sigma_{-}^{M,Y_{old}} -
[K_1
\cdot
(\Sigma_{-}^C)^T]^T
\cdot
G^{-1} \cdot
[K_1
\cdot
(\Sigma_{-}^C)^T]
=
\Sigma_{-}^{M,Y_{old}} -
B
\cdot
G^{-1} \cdot
B^T
\end{multline}
where
\begin{equation}
B
\triangleq
\Sigma_{-}^C
\cdot
K_1^T
=
\Sigma_{-}^{C}
\cdot
(\comb{A}{I}{}{})^T
\cdot
K.
\end{equation}

\subsubsection*{$\boldsymbol{\Sigma_{+}^{(Y_{old}, Y_{new})}}$ - Method 1:}

Using the block-wise matrix inversion again we know that  $\Sigma_{+}^{cross}$ from Eq.~(\ref{eq:PosteriorCovStr}) is equal to:
\begin{multline}
\Sigma_{+}^{cross} =
- (\Lambda_{-} + A_{old}^T \cdot A_{old})^{-1}
\cdot
A_{old}^T \cdot A_{new}
\cdot
(A_{new}^T  \cdot C^{-1} \cdot A_{new})^{-1}
=\\
=
- (\Sigma_{-} -
\Sigma_{-} \cdot
A_{old}^T \cdot
C^{-1}
\cdot
A_{old}
\cdot \Sigma_{-})
\cdot
A_{old}^T \cdot A_{new}
\cdot
(A_{new}^T  \cdot C^{-1} \cdot A_{new})^{-1}
=\\
=
- \Sigma_{-} \cdot
A_{old}^T \cdot A_{new}
\cdot
(A_{new}^T  \cdot C^{-1} \cdot A_{new})^{-1}
+
\Sigma_{-} \cdot
A_{old}^T \cdot
C^{-1}
\cdot
A_{old}
\cdot \Sigma_{-}
\cdot
A_{old}^T \cdot A_{new}
\cdot
(A_{new}^T  \cdot C^{-1} \cdot A_{new})^{-1}
=\\
=
\Sigma_{-} \cdot
A_{old}^T
\cdot
[- I_m
+
C^{-1}
\cdot
A_{old}
\cdot \Sigma_{-}
\cdot
A_{old}^T]
\cdot A_{new}
\cdot
(A_{new}^T  \cdot C^{-1} \cdot A_{new})^{-1}
=\\
=
\Sigma_{-} \cdot
A_{old}^T
\cdot
[
C^{-1}
\cdot
\comb{A}{I}{}{} \cdot \Sigma_{-}^{I} \cdot (\comb{A}{I}{}{})^T
- I_m]
\cdot A_{new}
\cdot
(A_{new}^T  \cdot C^{-1} \cdot A_{new})^{-1},
\end{multline}
where matrix $C$ is defined in Eq.~(\ref{eq:CMatDef})

Then $\Sigma_{+}^{(Y_{old}, Y_{new})}$ can be calculated by retrieving from $\Sigma_{+}^{cross}$ the entries that  correspond to $Y_{old}$ rows and $Y_{new}$ columns:
\begin{equation}
\Sigma_{+}^{(Y_{old}, Y_{new})}
=
\Sigma_{-}^{(Y_{old},:)} \cdot
A_{old}^T
\cdot
[
C^{-1}
\cdot
\comb{A}{I}{}{} \cdot \Sigma_{-}^{I} \cdot (\comb{A}{I}{}{})^T
- I_m]
\cdot A_{new}
\cdot
[(A_{new}^T  \cdot C^{-1} \cdot A_{new})^{-1}]^{(:, Y_{new})}
=
\Sigma_{-}^C
\cdot
(\comb{A}{I}{}{})^T
\cdot
[
C^{-1}
\cdot
\comb{A}{I}{}{} \cdot \Sigma_{-}^{I} \cdot (\comb{A}{I}{}{})^T
- I_m]
\cdot A_{new}
\cdot
P
\end{equation}
where matrix $P$ is defined in Eq.~(\ref{eq:PMatDef}).

\subsubsection*{$\boldsymbol{\Sigma_{+}^{(Y_{old}, Y_{new})}}$ - Method 2:}

Using another form of block-wise matrix inversion,  $\Sigma_{+}^{cross}$ from Eq.~(\ref{eq:PosteriorCovStr}) is equal to:
\begin{multline}
\Sigma_{+}^{cross}
=
- 
[\Sigma_{-} -
\Sigma_{-} \cdot
A_{old}^T \cdot K^T \cdot
G^{-1} \cdot
K \cdot A_{old} \cdot
\cdot \Sigma_{-}
]
\cdot
A_{old}^T \cdot A_{new}
\cdot
(A_{new}^T \cdot A_{new})^{-1}
=\\
=
[
\Sigma_{-}
\cdot 
A_{old}^T \cdot K^T \cdot
G^{-1} \cdot
K \cdot A_{old} \cdot
\cdot \Sigma_{-}
\cdot
A_{old}^T 
- 
\Sigma_{-}
\cdot
A_{old}^T 
]
\cdot A_{new}
\cdot
(A_{new}^T \cdot A_{new})^{-1}
=\\
=
[
\Sigma_{-}
\cdot 
A_{old}^T \cdot K^T \cdot
G^{-1} \cdot
K \cdot A_{old} \cdot
\cdot \Sigma_{-}
\cdot
A_{old}^T 
- 
\Sigma_{-}
\cdot
A_{old}^T 
]
\cdot A_{new}
\cdot
F
=\\
=
[
\Sigma_{-}
\cdot 
A_{old}^T \cdot K^T \cdot
G^{-1} \cdot
K \cdot
\comb{A}{I}{}{} \cdot \Sigma_{-}^{I} \cdot (\comb{A}{I}{}{})^T
- 
\Sigma_{-}
\cdot
A_{old}^T 
]
\cdot A_{new}
\cdot
F
\end{multline}
where matrix $F$ is defined in Eq.~(\ref{eq:FMatDef}).

Then $\Sigma_{+}^{(Y_{old}, Y_{new})}$ can be calculated by retrieving from $\Sigma_{+}^{cross}$ the entries that  correspond to $Y_{old}$ rows and $Y_{new}$ columns:
\begin{multline}
\Sigma_{+}^{(Y_{old}, Y_{new})}
=
[
\Sigma_{-}^{(Y_{old},:)}
\cdot 
A_{old}^T \cdot K^T \cdot
G^{-1} \cdot
K \cdot
\comb{A}{I}{}{} \cdot \Sigma_{-}^{I} \cdot (\comb{A}{I}{}{})^T
- 
\Sigma_{-}^{(Y_{old},:)}
\cdot
A_{old}^T 
]
\cdot A_{new}
\cdot
F^{(:, Y_{new})}
=\\
=
[
\Sigma_{-}^C
\cdot
(\comb{A}{I}{}{})^T
\cdot K^T \cdot
G^{-1} \cdot
K \cdot
\comb{A}{I}{}{} \cdot \Sigma_{-}^{I} \cdot (\comb{A}{I}{}{})^T
- 
\Sigma_{-}^C
\cdot
(\comb{A}{I}{}{})^T
]
\cdot A_{new}
\cdot
F^{(:, Y_{new})}
=\\
=
\Sigma_{-}^C
\cdot
[
(\comb{A}{I}{}{})^T
\cdot K^T \cdot
G^{-1} \cdot
K \cdot
\comb{A}{I}{}{} \cdot \Sigma_{-}^{I}
- 
I_k
]
\cdot
(\comb{A}{I}{}{})^T
\cdot A_{new}
\cdot
F^{(:, Y_{new})}
=\\
=
\Sigma_{-}^C
\cdot
[
K_1^T \cdot
G^{-1} \cdot
K_1 \cdot \Sigma_{-}^{I}
- 
I_k
]
\cdot
(\comb{A}{I}{}{})^T
\cdot A_{new}
\cdot
F^{(:, Y_{new})}
\end{multline}
where matrix $K_1$ is defined in Eq.~(\ref{eq:K1MatDef}) and identity matrix $I_k$ has dimension $|\comb{X}{I}{}{}|$.
\hfill $\blacksquare$

Note that the columns inside information matrices do not have to be ordered in any particular way, and that the provided above proof is correct for any ordering whatsoever.

\subsection{Proof of Lemma \ref{lemma:SquareFFuncLemma} - \Actsqr Case}
\label{sec:SquareFFuncLemmaProof}

The \actsqr case is a special instance of the \actrect case, and thus we will use here the same setup as for the \actrect case. In other words, we will use the partitioning that was defined in Eq.~(\ref{eq:PosteriorMargStr}), (\ref{eq:PriorMargStr}) and (\ref{eq:AMatStr}).

In \actsqr case we have that $m = |X_{new}|$ from which we can conclude that matrix $A_{new}$ from Eq.~(\ref{eq:AMatStr}) is a squared matrix. Then, matrix $K$ from Eq.~(\ref{eq:KMatDef}) is equal to zero matrix:
\begin{equation}
K
=
I_m -
A_{new} \cdot
(A_{new}^T \cdot A_{new})^{-1} \cdot
A_{new}^T
=
I_m -
A_{new} \cdot
A_{new}^{-1}
\cdot
(A_{new}^T)^{-1} \cdot
A_{new}^T
=
0.
\end{equation}
Further,  matrices $K_1$ and $B$ from Eq.~(\ref{eq:K1MatDefF}) and Eq.~(\ref{eq:BMatDefF}) contain only zeros, and $\Sigma_{+}^{M, Y_{old}}$ is equal to:
\begin{equation}
\Sigma_{+}^{M, Y_{old}}
=
\Sigma_{-}^{M,Y_{old}} -
B
\cdot
G^{-1} \cdot
B^T
=
\Sigma_{-}^{M,Y_{old}}.
\end{equation}
Next,  $\Sigma_{+}^{new}$ from Eq.~(\ref{eq:PosteriorNewCov}) can be calculated as:
\begin{equation}
\Sigma_{+}^{new} =
(A_{new}^T \cdot
C^{-1}
\cdot A_{new})^{-1}
=
A_{new}^{-1}
\cdot
C
\cdot
(A_{new}^T)^{-1}
=
A_{new}^{-1}
\cdot
C
\cdot
(A_{new}^{-1})^{T},
\end{equation}
and $\Sigma_{+}^{M, Y_{new}}$ is equal to:
\begin{equation}
\Sigma_{+}^{M, Y_{new}} =
[A_{new}^{-1}]^{(Y_{new}, :)}
\cdot
C
\cdot
[(A_{new}^{-1})^{T}]^{(:,Y_{new})}
=
[A_{new}^{-1}]^{(Y_{new}, :)}
\cdot
C
\cdot
([A_{new}^{-1}]^{(Y_{new}, :)})^T
=
A_{iv}
\cdot
C
\cdot
A_{iv}^T,
\end{equation}
where 
\begin{equation}
A_{iv}
\triangleq
[A_{new}^{-1}]^{(Y_{new}, :)},
\end{equation}
and can be efficiently calculated through Matlab backslash operator:
\begin{equation}
A_{iv} =
A_{new}
\backslash
I^{(:, Y_{new})}.
\end{equation}
Next, we can reduce Eq.~(\ref{eq:PosteriorMargCross}) to:
\begin{multline}
\Sigma_{+}^{(Y_{old}, Y_{new})}
=
\Sigma_{-}^C
\cdot
[
K_1^T \cdot
G^{-1} \cdot
K_1 \cdot \Sigma_{-}^{I}
- 
I_k
]
\cdot
(\comb{A}{I}{}{})^T
\cdot A_{new}
\cdot
F^{(:, Y_{new})}
=\\
=
- \Sigma_{-}^C
\cdot
(\comb{A}{I}{}{})^T
\cdot A_{new}
\cdot
F^{(:, Y_{new})}
=
- [\Sigma_{-}^C
\cdot
(\comb{A}{I}{}{})^T
\cdot A_{new}
\cdot
F]^{(:, Y_{new})}
=\\
=
- [\Sigma_{-}^C
\cdot
(\comb{A}{I}{}{})^T
\cdot A_{new}
\cdot
A_{new}^{-1}
\cdot
(A_{new}^T)^{-1}
]^{(:, Y_{new})}
=
- [\Sigma_{-}^C
\cdot
(\comb{A}{I}{}{})^T
\cdot
(A_{new}^T)^{-1}
]^{(:, Y_{new})}
=\\
=
- \Sigma_{-}^C
\cdot
(\comb{A}{I}{}{})^T
\cdot
[(A_{new}^T)^{-1}
]^{(:, Y_{new})}
=
- \Sigma_{-}^C
\cdot
(\comb{A}{I}{}{})^T
\cdot
([A_{new}^{-1}
]^{(Y_{new},:)})^T
=
- \Sigma_{-}^C
\cdot
(\comb{A}{I}{}{})^T
\cdot
(A_{iv})^T
\end{multline}
\hfill $\blacksquare$

Note that the columns inside information matrices do not have to be ordered in any particular way, and that the provided above proof is correct for any ordering whatsoever.

\subsection{Proof of Lemma \ref{lemma:RelinFFuncLemma} - Relinearization Case}
\label{sec:RelinFFuncLemmaProof}

As we saw in Eq.~(\ref{eq:RelinearPosteriorInfoMatrixCmplx}), the information update here has the following form:
\begin{equation}
\Lambda_{+} = 
\Lambda_{-} +
B^T \cdot B
, \quad
B 
\triangleq
\begin{pmatrix} 
i A_{-}\\
A_{+}
\end{pmatrix}.
\end{equation}
First, denote by $\comb{A}{I}{}{-}$ the non-zero columns of $A_{-}$'s and by $\comb{A}{I}{}{+}$ the non-zero columns of $A_{+}$ (note that indices of such columns are the same in both $A_{-}$ and $A_{+}$). Next, apply Lemma \ref{lemma:NAugFFuncLemma} as following:
\begin{equation}
\Sigma_{+}^{M,Y}
=
\Sigma_{-}^{Y} -
V
\cdot
C^{-1}
\cdot
V^T
, \quad
V
\triangleq
\Sigma_{-}^{C}
\cdot
(\comb{B}{I}{}{})^T
, \quad
C
\triangleq
I_m + 
\comb{B}{I}{}{} \cdot \Sigma_{-}^{I} \cdot (\comb{B}{I}{}{})^T
, \quad
\comb{B}{I}{}{} 
\triangleq
\begin{pmatrix} 
i \cdot \comb{A}{I}{}{-}\\
\comb{A}{I}{}{+}
\end{pmatrix},
\label{eq:RelinBase1}
\end{equation}
where $\Sigma_{-}^I \equiv \Sigma_{-}^{M,\comb{X}{I}{}{}}$ is the prior marginal covariance of variables $\comb{X}{I}{}{}$ \involved in the relinearized factors $F_R$; $\Sigma_{-}^{Y} \equiv \Sigma_{-}^{M, Y}$ is the prior marginal covariance of variables $Y$; $\Sigma_{-}^C$ is the cross-covariance between $Y$ and $X^I$.

Next we can see that:
\begin{equation}
C
=
I_m + 
\begin{pmatrix} 
i \cdot \comb{A}{I}{}{-}\\
\comb{A}{I}{}{+}
\end{pmatrix}
\cdot \Sigma_{-}^{I} \cdot
\begin{pmatrix} 
i \cdot (\comb{A}{I}{}{-})^T&
(\comb{A}{I}{}{+})^T
\end{pmatrix}
=
\begin{pmatrix}
I - \comb{A}{I}{}{-}
\cdot
\Sigma_{-}^{I}
\cdot
(\comb{A}{I}{}{-})^T
&
i
\cdot
\comb{A}{I}{}{-}
\cdot
\Sigma_{-}^{I}
\cdot
(\comb{A}{I}{}{+})^T
\\
i
\cdot
\comb{A}{I}{}{+}
\cdot
\Sigma_{-}^{I}
\cdot
(\comb{A}{I}{}{-})^T
&
I + \comb{A}{I}{}{+}
\cdot
\Sigma_{-}^{I}
\cdot
(\comb{A}{I}{}{+})^T
\end{pmatrix}
\triangleq
\begin{pmatrix}
C_{11} & i \cdot C_{12}\\
i \cdot C_{12}^T & C_{22}\\
\end{pmatrix},
\end{equation}
where $I$ is the identity matrix of an appropriate dimension. 
Note that while $C_{11}$ sometimes can be not positive definite (PD) matrix,  $C_{22}$ is always PD and therefore is invertible.

Next, we will use a block-wise inversion in order to calculate $C^{-1}$:
\begin{equation}
C^{-1}
\triangleq
\begin{pmatrix}
C_{11}^{inv} & C_{12}^{inv}\\
(C_{12}^{inv})^T & C_{22}^{inv}\\
\end{pmatrix},
\end{equation}
\begin{equation}
C_{11}^{inv}
\triangleq
(C_{11} + C_{12} \cdot C_{22}^{-1} \cdot C_{12}^T)^{-1},
\end{equation}
\begin{equation}
C_{12}^{inv}
\triangleq
-i
\cdot
C_{11}^{inv}
\cdot
C_{12} \cdot C_{22}^{-1},
\end{equation}
\begin{equation}
C_{22}^{inv}
\triangleq
C_{22}^{-1}
-
C_{22}^{-1}
\cdot C_{12}^T
\cdot
C_{11}^{inv}
\cdot C_{12}
\cdot C_{22}^{-1}.
\end{equation}
Using the above notations we can see that:
\begin{multline}
(\comb{B}{I}{}{})^T
\cdot
C^{-1}
\cdot
\comb{B}{I}{}{}
=
\begin{pmatrix} 
i \cdot (\comb{A}{I}{}{-})^T&
(\comb{A}{I}{}{+})^T
\end{pmatrix}
\cdot
\begin{pmatrix}
C_{11}^{inv} & C_{12}^{inv}\\
(C_{12}^{inv})^T & C_{22}^{inv}\\
\end{pmatrix}
\cdot
\begin{pmatrix} 
i \cdot \comb{A}{I}{}{-}\\
\comb{A}{I}{}{+}
\end{pmatrix}
=\\
=
(\comb{A}{I}{}{+})^T
\cdot
C_{22}^{-1}
\cdot
\comb{A}{I}{}{+}
-
(\comb{A}{I}{}{-})^T
\cdot
C_{11}^{inv}
\cdot
\comb{A}{I}{}{-}
+
(\comb{A}{I}{}{-})^T
\cdot
C_{11}^{inv}
\cdot
C_{12}
\cdot
C_{22}^{-1}
\cdot
\comb{A}{I}{}{+}
+
\\
+
(\comb{A}{I}{}{+})^T
\cdot
C_{22}^{-1}
\cdot
C_{12}^{T}
\cdot
C_{11}^{inv}
\cdot
\comb{A}{I}{}{-}
-
(\comb{A}{I}{}{+})^T
\cdot
C_{22}^{-1}
\cdot
C_{12}^{T}
\cdot
C_{11}^{inv}
\cdot
C_{12}
\cdot
C_{22}^{-1}
\cdot
\comb{A}{I}{}{+}
=\\
=
(\comb{A}{I}{}{+})^T
\cdot
C_{22}^{-1}
\cdot
\comb{A}{I}{}{+}
-
\Big[
(\comb{A}{I}{}{-})^T
-
(\comb{A}{I}{}{+})^T
\cdot
C_{22}^{-1}
\cdot
C_{12}^{T}
\Big]
\cdot
C_{11}^{inv}
\cdot
\Big[
(\comb{A}{I}{}{-})^T
-
(\comb{A}{I}{}{+})^T
\cdot
C_{22}^{-1}
\cdot
C_{12}^{T}
\Big]^T
=\\
=
(\comb{A}{I}{}{+})^T
\cdot
C_{22}^{-1}
\cdot
\comb{A}{I}{}{+}
-
\Big[
(\comb{A}{I}{}{-})^T
-
(\comb{A}{I}{}{+})^T
\cdot
C_{22}^{-1}
\cdot
C_{12}^{T}
\Big]
\cdot
\Big[
C_{11} + C_{12} \cdot C_{22}^{-1} \cdot C_{12}^T
\Big]^{-1}
\cdot
\Big[
(\comb{A}{I}{}{-})^T
-
(\comb{A}{I}{}{+})^T
\cdot
C_{22}^{-1}
\cdot
C_{12}^{T}
\Big]^T
=\\
=
(\comb{A}{I}{}{+})^T
\cdot
\Big[
I + 
\comb{A}{I}{}{+}
\cdot
\Sigma_{-}^{I}
\cdot
(\comb{A}{I}{}{+})^T
\Big]^{-1}
\cdot
\comb{A}{I}{}{+}
-\\
-
\Bigg[
(\comb{A}{I}{}{-})^T
-
(\comb{A}{I}{}{+})^T
\cdot
\Big[
I + 
\comb{A}{I}{}{+}
\cdot
\Sigma_{-}^{I}
\cdot
(\comb{A}{I}{}{+})^T
\Big]^{-1}
\cdot
\comb{A}{I}{}{+}
\cdot
\Sigma_{-}^{I}
\cdot
(\comb{A}{I}{}{-})^T
\Bigg]
\cdot
J^{-1}
\cdot
\Bigg[
(\comb{A}{I}{}{-})^T
-
(\comb{A}{I}{}{+})^T
\cdot
\Big[
I + 
\comb{A}{I}{}{+}
\cdot
\Sigma_{-}^{I}
\cdot
(\comb{A}{I}{}{+})^T
\Big]^{-1}
\cdot
\comb{A}{I}{}{+}
\cdot
\Sigma_{-}^{I}
\cdot
(\comb{A}{I}{}{-})^T
\Bigg]^T
,
\end{multline}
\begin{equation}
J
\triangleq
C_{11} + C_{12} \cdot C_{22}^{-1} \cdot C_{12}^T
=
I - \comb{A}{I}{}{-}
\cdot
\Sigma_{-}^{I}
\cdot
(\comb{A}{I}{}{-})^T
+
\comb{A}{I}{}{-}
\cdot
\Sigma_{-}^{I}
\cdot
(\comb{A}{I}{}{+})^T
\cdot
\Big[
I + 
\comb{A}{I}{}{+}
\cdot
\Sigma_{-}^{I}
\cdot
(\comb{A}{I}{}{+})^T
\Big]^{-1}
\cdot
\comb{A}{I}{}{+}
\cdot
\Sigma_{-}^{I}
\cdot
(\comb{A}{I}{}{-})^T
.
\end{equation}
Next, introduce new notations:
\begin{equation}
M_2
\triangleq
(\comb{A}{I}{}{+})^T
\diagup
chol
\Big[
I + 
\comb{A}{I}{}{+} \cdot \Sigma_{-}^{I} \cdot (\comb{A}{I}{}{+})^T
\Big],
\end{equation}
\begin{equation}
G
\triangleq
M_2^T
\cdot
\Sigma_{-}^{I} \cdot (\comb{A}{I}{}{-})^T
,
\end{equation}
where $chol(\cdot)$ represents cholesky decomposition which returns an upper triangular matrix; "$\diagup$" is the  backslash operator from Matlab syntax ($A \diagup B = A \cdot B^{-1}$).

It can be clearly seen that:
\begin{equation}
M_2 \cdot M_2^T
=
(\comb{A}{I}{}{+})^T
\cdot
\Big[
I + 
\comb{A}{I}{}{+}
\cdot
\Sigma_{-}^{I}
\cdot
(\comb{A}{I}{}{+})^T
\Big]^{-1}
\cdot
\comb{A}{I}{}{+},
\end{equation}
\begin{equation}
G^T \cdot G
=
\comb{A}{I}{}{-}
\cdot
\Sigma_{-}^{I}
\cdot
(\comb{A}{I}{}{+})^T
\cdot
\Big[
I + 
\comb{A}{I}{}{+}
\cdot
\Sigma_{-}^{I}
\cdot
(\comb{A}{I}{}{+})^T
\Big]^{-1}
\cdot
\comb{A}{I}{}{+}
\cdot
\Sigma_{-}^{I}
\cdot
(\comb{A}{I}{}{-})^T
,
\end{equation}
\begin{equation}
M_2 \cdot G
=
(\comb{A}{I}{}{+})^T
\cdot
\Big[
I + 
\comb{A}{I}{}{+}
\cdot
\Sigma_{-}^{I}
\cdot
(\comb{A}{I}{}{+})^T
\Big]^{-1}
\cdot
\comb{A}{I}{}{+}
\cdot
\Sigma_{-}^{I}
\cdot
(\comb{A}{I}{}{-})^T
.
\end{equation}
Using matrices $M_2$ and $G$, we can rewrite an expression for $(\comb{B}{I}{}{})^T
\cdot
C^{-1}
\cdot
\comb{B}{I}{}{}$ as:
\begin{equation}
(\comb{B}{I}{}{})^T
\cdot
C^{-1}
\cdot
\comb{B}{I}{}{}
=
M_2 \cdot M_2^T
-
\Big[
(\comb{A}{I}{}{-})^T
-
M_2 \cdot G
\Big]
\cdot
J^{-1}
\cdot
\Big[
(\comb{A}{I}{}{-})^T
-
M_2 \cdot G
\Big]^T,
\end{equation}
\begin{equation}
J
=
I - \comb{A}{I}{}{-}
\cdot
\Sigma_{-}^{I}
\cdot
(\comb{A}{I}{}{-})^T
+
G^T \cdot G
.
\end{equation}
Next, let us define another matrix:
\begin{equation}
M_1
\triangleq
\Big[
(\comb{A}{I}{}{-})^T - M_2 \cdot G
\Big]
\diagup
chol
\Big[
I
-
\comb{A}{I}{}{-} \cdot \Sigma_{-}^{I} \cdot (\comb{A}{I}{}{-})^T
+
G^T \cdot G
\Big]
\end{equation}
with $M_1 \cdot M_1^T$ being equal to:
\begin{equation}
M_1 \cdot M_1^T
=
\Big[
(\comb{A}{I}{}{-})^T
-
M_2 \cdot G
\Big]
\cdot
\Big[
I
-
\comb{A}{I}{}{-} \cdot \Sigma_{-}^{I} \cdot (\comb{A}{I}{}{-})^T
+
G^T \cdot G
\Big]^{-1}
\cdot
\Big[
(\comb{A}{I}{}{-})^T
-
M_2 \cdot G
\Big]^T
=
\Big[
(\comb{A}{I}{}{-})^T
-
M_2 \cdot G
\Big]
\cdot
J^{-1}
\cdot
\Big[
(\comb{A}{I}{}{-})^T
-
M_2 \cdot G
\Big]^T
.
\end{equation}
Thus, we will have:
\begin{equation}
(\comb{B}{I}{}{})^T
\cdot
C^{-1}
\cdot
\comb{B}{I}{}{}
=
M_2 \cdot M_2^T
-
M_1 \cdot M_1^T
=
M \cdot M^T
, \quad
M 
\triangleq
\begin{pmatrix} 
i M_1 & M_2\\
\end{pmatrix}.
\end{equation}
By combining the above equation with Eq.~(\ref{eq:RelinBase1}) we can conclude that:
\begin{equation}
\Sigma_{+}^{M,Y}
=
\Sigma_{-}^{Y} -
\Sigma_{-}^{C}
\cdot
M \cdot M^T
\cdot
(\Sigma_{-}^{C})^T
=
	\Sigma_{-}^{Y} -
U
\cdot
U^T
, \quad
U
\triangleq
\Sigma_{-}^{C}
\cdot
M
\end{equation}
\hfill $\blacksquare$

Note that the columns inside information matrices do not have to be ordered in any particular way, and that the provided above proof is correct for any ordering whatsoever.

\subsection{Sum of Information Gains}
\label{sec:IGSumProof}

Consider action $a$ with \increment $I(a) = \{ F_{new}, X_{new} \}$. Further, consider specific partitioning of $a$ into sub-actions $a = \{ a_1', \ldots, a_k' \}$ where each sub-action $a_i'$ has \increment $I_i(a_i') = \{ F_{i,new}, X_{i,new} \}$. The factor sets $F_{i,new}$ are disjoint, as also are the new variable sets $X_{i,new}$. Also, for proper action partitioning we will have $\cup_{i = 1}^{k} F_{i,new} = F_{new}$ and $\cup_{i = 1}^{k} X_{i,new} = X_{new}$.

Next, we will prove that information gain (IG) of $a$ is equal to sum of IGs of sub-actions $\{ a_i' \}_{i = 1}^{k}$ in \unfocused scenario. Similar proof can be shown also for \focused BSP.

The \unfocused IG of action $a$ by definition is:
\begin{equation}
J_{IG}(a) =
\mathcal{H}(b[X_{-}]) - \mathcal{H}(b[X_{+}])
\end{equation}
where $b[X_{-}]$ is a prior state belief before applying action $a$, and $b[X_{+}]$ is a posterior state belief after applying it.

Additionally, denote posterior state belief of each sub-action $a_i'$ as $b_i[X_{i,+}]$. When applying sub-actions consecutively in sequence, belief propagation will have next form:
\begin{equation}
b[X_{-}] 
\Longrightarrow
b_1[X_{1,+}]
\Longrightarrow
b_2[X_{2,+}]
\Longrightarrow
\cdots
\Longrightarrow
b_{k-1}[X_{k-1,+}]
\Longrightarrow
b[X_{+}]
\end{equation}

Then, the IG of each sub-action is equal to:
\begin{align*} 
J_{IG}(a_1') &=
\mathcal{H}(b[X_{-}]) - \mathcal{H}(b_1[X_{1,+}])
\\
J_{IG}(a_2') &=
\mathcal{H}(b_1[X_{1,+}]) - \mathcal{H}(b_2[X_{2,+}])
\\
&\cdots
\\
J_{IG}(a_{k-1}') &=
\mathcal{H}(b_{k-2}[X_{k-2,+}]) - \mathcal{H}(b_{k-1}[X_{k-1,+}])
\\
J_{IG}(a_{k}') &=
\mathcal{H}(b_{k-1}[X_{k-1,+}]) - \mathcal{H}(b[X_{+}])
\end{align*}
and sum of these IGs is equal to:
\begin{equation}
\sum_{i = 1}^{k} J_{IG}(a_i') = 
\mathcal{H}(b[X_{-}]) - \mathcal{H}(b[X_{+}])
=
J_{IG}(a)
\end{equation}
\hfill $\blacksquare$

\end{document}